\documentclass[lettersize,journal]{IEEEtran}
\usepackage{amsmath,amsfonts}
\usepackage{tabularx}
\usepackage{makecell}
\usepackage{algorithmic}
\usepackage{array}
\usepackage[caption=false,font=normalsize,labelfont=sf,textfont=sf]{subfig}
\usepackage{textcomp}
\usepackage{stfloats}
\usepackage{url}
\usepackage{verbatim}
\usepackage{graphicx}

\def\BibTeX{{\rm B\kern-.05em{\sc i\kern-.025em b}\kern-.08em
    T\kern-.1667em\lower.7ex\hbox{E}\kern-.125emX}}
\usepackage{balance}
\usepackage{graphicx}
\usepackage[percent]{overpic}
\usepackage[dvipsnames]{xcolor}
\usepackage{tikz}
\usepackage{cite}

\usepackage{graphicx}
\usepackage{amsmath}
\usepackage{xcolor}
\usepackage{tikz}

\usetikzlibrary{
    arrows.meta,
    positioning,
    shapes.geometric,
    calc
}

\begin{document}
\title{K-TRAIL: Simulator-Guided Generative Design of EM/RF Circuits }
\author{Piyush Saha, Evan Newell, Hanna O'Leary, Arun Natarajan, and Alireza Aghasi%
\thanks{This work
	is supported in part by the Center for Design of Analog-Digital Integrated
	Circuits (CDADIC).} 
\thanks{P. Saha, E. Newell, H. O'Leary, and A. Aghasi are with the School of Electrical Engineering and Computer Science, Oregon State University, Corvallis, OR 97331 USA (e-mail: sahapi@oregonstate.edu; newellev@oregonstate.edu; olearyha@oregonstate.edu; alireza.aghasi@oregonstate.edu).}%
\thanks{A. Natarajan is with the Department of Electrical and Computer Engineering, Yale University, New Haven, CT 06511 USA (e-mail: arun.natarajan@yale.edu).}%
}

\markboth{Preprint}{}


\maketitle

\begin{abstract}
Inverse design of RF and electromagnetic (EM) circuits is challenging because the relationship between circuit layout and electrical response is non-unique, and full-wave simulation is computationally expensive. This paper presents K-TRAIL, a simulator-guided generative framework for automated EM/RF circuit synthesis. K-TRAIL combines diffusion-based layout generation with derivative-free ensemble Kalman guidance, allowing feedback from a black-box EM simulator to refine candidate layouts during generation without requiring adjoint sensitivities or differentiable solver models. The framework supports both synthesis from prescribed S-parameter responses and synthesis directly from RF performance constraints. Experiments on multi-layer RFIC structures show that simulator-guided generation improves agreement with target responses and can identify structurally distinct layouts that satisfy circuit-level design requirements. The proposed approach provides a practical path toward generative, verification-aware RF circuit design while retaining the flexibility to explore diverse layout topologies.
\end{abstract}

\section{Introduction}
\IEEEPARstart{I}nverse EM design offers an alternative approach to the design of passive structures, where, instead of traditional optimization of a small number of parameters around a specific structure,  the synthesis engine searches over a much larger class of admissible layouts for the desired electromagnetic response. Recent work has shown some success with deep-learning-based forward models that predict S-parameters of arbitrary or pixelated on-chip structures, enabling optimization over layouts that are irregular, nonintuitive, and difficult to obtain using classical design templates \cite {karahan2023deep_inverse_mmwave,chu2026review,sengupta2026ai}. Such methods have already demonstrated promising results in mm-wave passives and active–passive co-designed circuits, including broadband power amplifiers and other RFIC building blocks\cite{liu2022deep,zhou202525,wang2025inverse}. More broadly, recent AI-assisted RFIC design frameworks point toward specification-to-layout automation to expand the reachable design space\cite{chu2026review,sengupta2026ai}.

    \begin{figure}[t]
    \centering
    \setlength{\fboxsep}{0pt}%
    \colorbox{white}{%
        \begin{overpic}[
    width=.485\textwidth,
    percent,
    tics=5
]{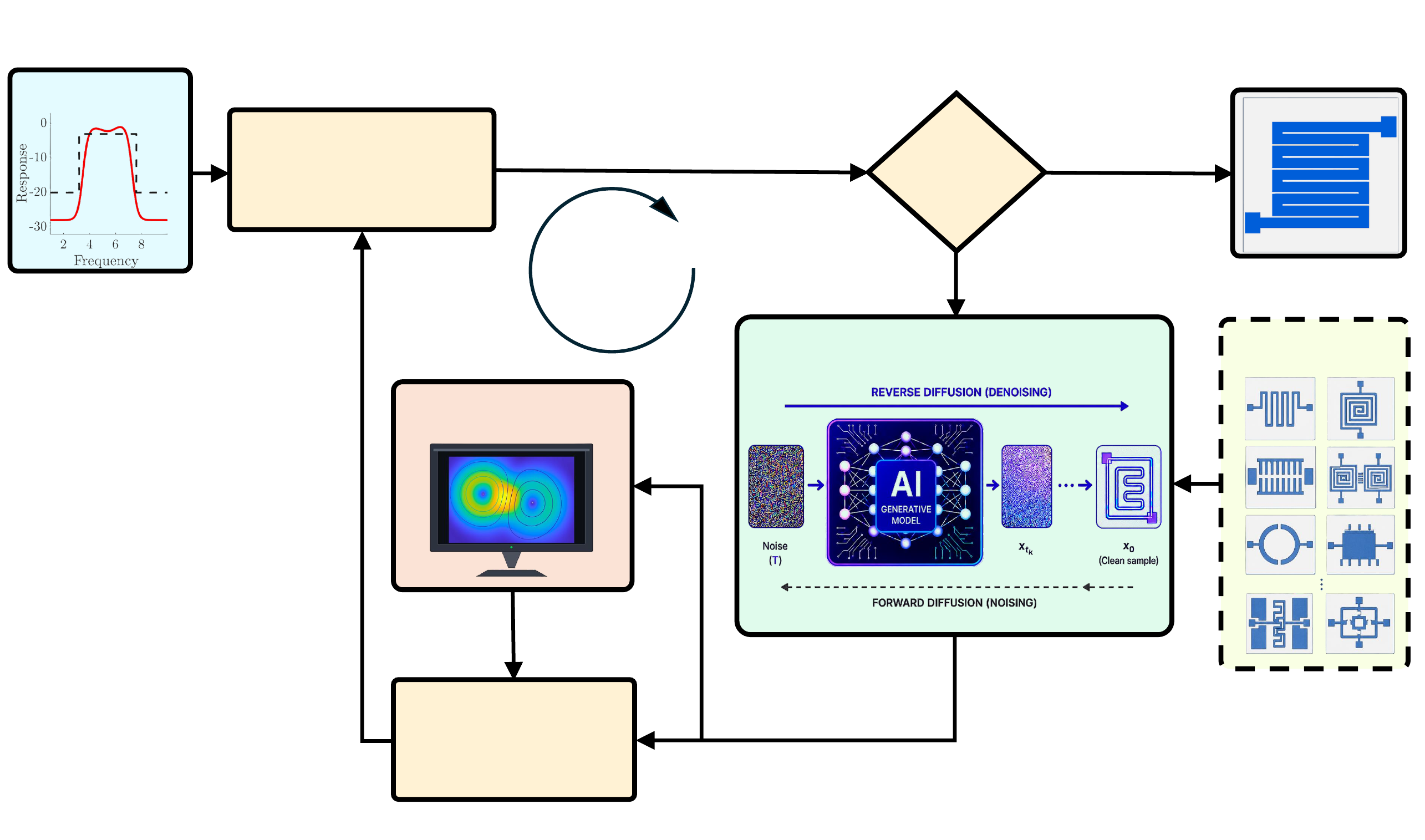}
\put(7.43,52.3115){\makebox(0,0)[c]{\scalebox{.85}{\rotatebox{0}{$y$}}}}
\put(25,48.5){\makebox(0,0)[c]{\scalebox{.7}{\rotatebox{0}{Evaluate}}}}
\put(25,45){\makebox(0,0)[c]{\scalebox{.85}{\rotatebox{0}{$y-\mathcal{E}(x_{t})$}}}}
\put(67,47){\makebox(0,0)[c]{\scalebox{.75}{\rotatebox{0}{stop?}}}}
\put(77,48){\makebox(0,0)[c]{\scalebox{.7}{\rotatebox{0}{yes}}}}
\put(92.865,54.5){\makebox(0,0)[c]{\scalebox{.85}{\rotatebox{0}{$x$}}}}
\put(92.5,34.4){\makebox(0,0)[c]{\scalebox{.7}{\rotatebox{0}{$x$ samples}}}}
\put(44.2,42){\makebox(0,0)[c]{\scalebox{.57}{\rotatebox{0}{Optimization/}}}}
\put(44.0,39.6){\makebox(0,0)[c]{\scalebox{.57}{\rotatebox{0}{Sampling}}}}
\put(44,37.2){\makebox(0,0)[c]{\scalebox{.57}{\rotatebox{0}{Loop}}}}
\put(36,30){\makebox(0,0)[c]{\scalebox{.7}{\rotatebox{0}{EM Simulator}}}}
\put(36,9){\makebox(0,0)[c]{\scalebox{.85}{\rotatebox{0}{Kalman }}}}
\put(36,5.8){\makebox(0,0)[c]{\scalebox{.8}{\rotatebox{0}{Correction }}}}
\put(67,34.5){\makebox(0,0)[c]{\scalebox{.7}{\rotatebox{0}{Trained Diffusion Model}}}}
\put(30,14.35){\makebox(0,0)[c]{\scalebox{.85}{\rotatebox{0}{ $\mathcal{E}(x) $}}}}

        \end{overpic}%
    }
    \caption{The proposed Kalman-correction based trajectory refinement for conditional diffusion-based inverse layout generation (\emph{K-TRAIL}) framework supports RF layout generation based on exact S-parameter targets as well as RF constraints}
    \label{fig:ktrail}
\end{figure}
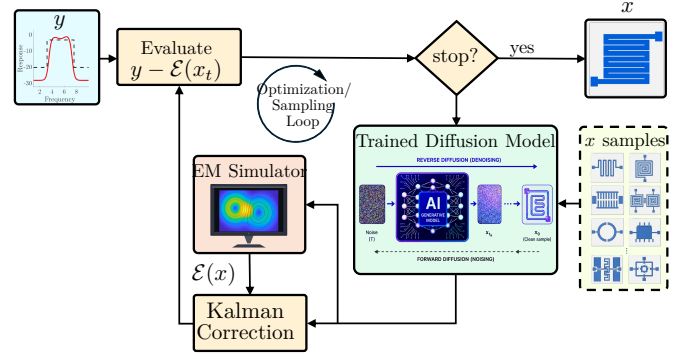

Despite this progress, inverse design of RF and EM structures remains fundamentally ill-posed since a desired scattering response does not uniquely determine a layout. This one-to-many mapping makes direct regression from specification to layout fragile, particularly for arbitrary or pixelated geometries with a larger number of layers, motivating the extensive use of EM solvers during design \cite{chenna2025algorithmic}. Secondly, although a learned forward surrogate model can reduce the cost of candidate evaluation, its errors can mislead an optimizer, leading to structures that fail under full-wave EM verification. This issue is especially acute for high-Q, broadband, or strongly coupled structures, where small local geometry perturbations can produce large changes in the scattering response. Finally, inverse design must proceed from RF constraints or incomplete S-parameter inputs (magnitude only, for example) rather than from exact S-parameter targets.

Generative models provide a natural way to address the non-uniqueness of inverse EM synthesis. Instead of predicting a single layout, a conditional generative model can learn a distribution of plausible structures conditioned on the desired RF response. Diffusion models are particularly attractive for this problem because pixelated metal/dielectric layouts can be represented as images, while the target S-parameters or RF specifications can be used as conditioning information\cite{guo2025dall,zhou2026ai}. After training, such models can rapidly generate diverse candidate layouts. However, a generated layout may resemble examples from the training set while failing to satisfy the target response after EM simulation. Therefore, generative RF design does not guarantee one-shot layout generation and must be embedded in a verification-aware loop that retains the EM simulator as the final guarantee of correctness.

In this work, we introduce a generative EM/RF synthesis framework using Kalman-correction-based trajectory refinement for inverse layout generation (K-TRAIL) that addresses this nonlinear inverse problem by combining conditional diffusion-based layout generation with Kalman correction using an EM simulator in the loop (Fig. \ref{fig:ktrail}). A candidate layout is mapped by a forward EM operator to its simulated response, and the design objective is to find layouts whose simulated responses match a target vector or satisfy a set of RF constraints. Rather than relying on a unique inverse map, the conditional diffusion model produces an ensemble of plausible layouts. The EM simulator then evaluates these candidates, and the results guide the reverse-diffusion trajectory via a derivative-free ensemble Kalman correction. This method incorporates physical feedback without requiring adjoint fields for gradient computation \cite{aghasi2011sensitivity, polydorides2012high}, differentiable EM solvers, or access to simulator internals.

The resulting \emph{K-TRAIL} framework preserves the diversity and prior-learning capability of diffusion models while explicitly enforcing constraint satisfaction and accurate layout synthesis. Importantly, it supports both exact S-parameter targets and constraint-based RF specifications, enabling practical application in typical design flows. It provides a direct tradeoff between computational cost and physical fidelity through the number of simulator-guided correction steps. By embedding the Kalman-correction and EM simulator inside the generative loop, we bridge the gap between data-driven inverse generation and verification-ready RF design automation.

The remainder of the paper is organized as follows. In Section~\ref{sec:bg}, we detail the challenges associated with inverse design and review prior approaches. Section~\ref{sec:exact_sparameter_synthesis} presents the proposed K-TRAIL framework for inverse design with exact S-parameter targets. Section~\ref{sec: FullInvDesExps} evaluates the proposed approach for both complex-valued and magnitude-only S-parameter targets. Section~\ref{sec:sparameter_conditions} extends K-TRAIL to inverse design based on RF performance constraints, and Section~\ref{sec:Block-level Inverse Design} demonstrates its application to block-level RF circuit design.

 \begin{figure}[t]
 \hspace{-.2cm}
\begin{overpic}[
    width=.53\textwidth,
    percent,
    tics=5
]{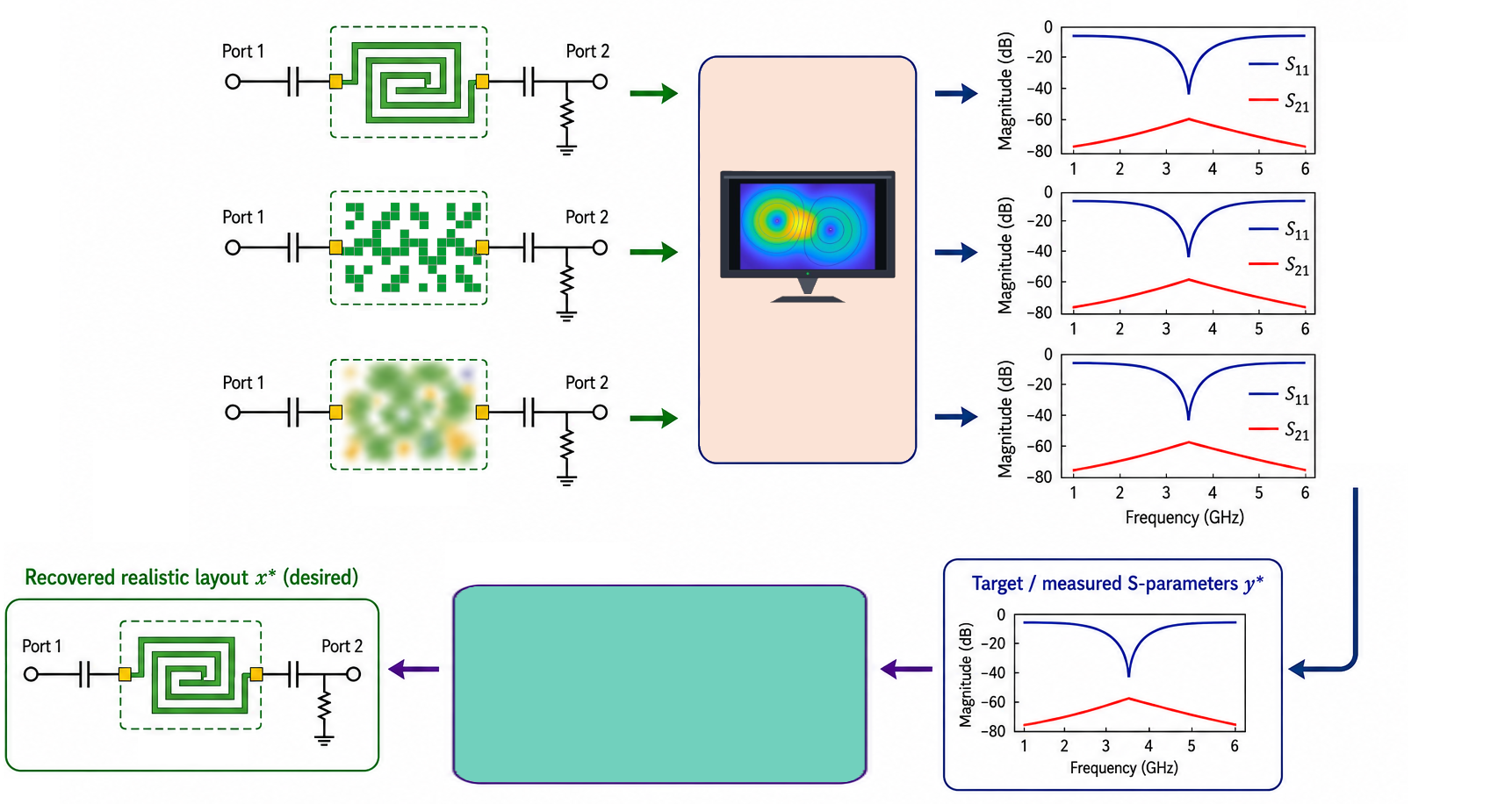}

\put(45,12){\makebox(0,0)[c]{\scalebox{.7}{\rotatebox{0}{Inverse Design}}}}
\put(43.5,9.3){\makebox(0,0)[c]{\scalebox{.57}{\rotatebox{0}{Find Desirable \& Realizable $x$}}}}
\put(43.5,7){\makebox(0,0)[c]{\scalebox{.57}{\rotatebox{0}{Such That: }}}}
\put(43.5,4){\makebox(0,0)[c]{\scalebox{.65}{\rotatebox{0}{$y^*\approx \mathcal{E}(x)$}}}}
\put(53,30){\makebox(0,0)[c]{\scalebox{.8}{\rotatebox{0}{EM}}}}
\put(53,26){\makebox(0,0)[c]{\scalebox{.8}{\rotatebox{0}{Simulator}}}}
\put(7,49){\makebox(0,0)[c]{\scalebox{.55}{\rotatebox{0}{{\color{OliveGreen}Desirable} \&}}}}
\put(7,46.5){\makebox(0,0)[c]{\scalebox{.55}{\rotatebox{0}{Physically {\color{OliveGreen}Realizable}}}}}
\put(7,38){\makebox(0,0)[c]{\scalebox{.55}{\rotatebox{0}{{\color{RedOrange}Undesirable} \&}}}}
\put(7,35.5){\makebox(0,0)[c]{\scalebox{.55}{\rotatebox{0}{Physically {\color{OliveGreen}Realizable}}}}}
\put(7,27){\makebox(0,0)[c]{\scalebox{.55}{\rotatebox{0}{{\color{RedOrange}Undesirable} \&}}}}
\put(7,24.5){\makebox(0,0)[c]{\scalebox{.55}{\rotatebox{0}{Physically {{\color{RedOrange}Unrealizable}}}}}}
\put(27,53){\makebox(0,0)[c]{\scalebox{.75}{\rotatebox{0}{$x_1 = x^*$}}}}
\put(27,42){\makebox(0,0)[c]{\scalebox{.75}{\rotatebox{0}{$x_2$}}}}
\put(27,31){\makebox(0,0)[c]{\scalebox{.75}{\rotatebox{0}{$x_3$}}}}
\put(53,45){\makebox(0,0)[c]{\scalebox{.75}{\rotatebox{0}{$\mathcal{E}(x)$}}}}
\put(63,49){\makebox(0,0)[c]{\scalebox{.75}{\rotatebox{0}{$y_1$}}}}
\put(63,38.5){\makebox(0,0)[c]{\scalebox{.75}{\rotatebox{0}{$y_2$}}}}
\put(63,28){\makebox(0,0)[c]{\scalebox{.75}{\rotatebox{0}{$y_3$}}}}
\put(91,36){\makebox(0,0)[c]{\scalebox{.7}{\rotatebox{90}{$y_3\approx y_2\approx y_1=y^*$}}}}
\end{overpic}
\caption{Multiple layouts with varying degrees of physical realizability and design quality can yield nearly identical scattering responses; inverse design seeks to recover an interpretable and physically realizable layout from a prescribed response.}
\label{fig:IP}
\end{figure}
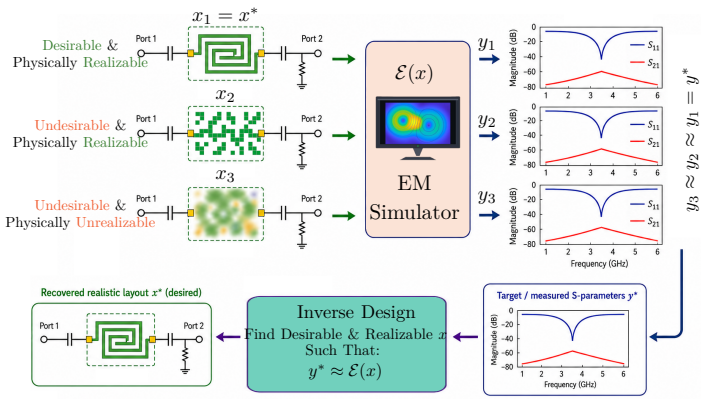

\section{Background on Inverse Generation of EM Designs}\label{sec:bg}
\subsection{Inverse-Problem View of Scattering-Response Synthesis}

Our objective is to automate the synthesis of integrated electromagnetic (EM) and RF structures that realize desired port responses. Let \(x\) denote a candidate structure, where \(x\) may represent a parameterized circuit layout, a pixelated metal/dielectric pattern, a topology, or a fabrication-constrained geometry. Let \(y\in\mathbb{C}^{m}\) denote the desired full or partial scattering response sampled over selected frequencies and port pairs. For example, \(y\) may contain samples of \(S_{11}(f)\), \(S_{21}(f)\), or the full multiport scattering matrix over a prescribed band. More generally, the specification may be expressed through functions of the scattering parameters, such as insertion loss, return loss, gain, isolation, matching bandwidth, or passband/stopband constraints, as discussed in Section~\ref{sec:sparameter_conditions}.

We denote by $\mathcal{E}:\mathcal{X}\rightarrow \mathbb{C}^{m}$ the forward EM operator that maps a structure \(x\) to its simulated scattering response. In practice, \(\mathcal{E}\) may represent a full-wave EM simulator, a circuit--EM co-simulation engine, or a surrogate model \cite{karahan2023deep_inverse_mmwave}. A standard target-matching formulation of the synthesis problem is
\begin{equation}
\min_{x} ~\Phi(x) :=
\frac{1}{2}\left\|y-\mathcal{E}(x)\right\|_2^2 .
\label{eq:inverse_synthesis_exact}
\end{equation}
Because the inverse problem is generally ill-posed, \eqref{eq:inverse_synthesis_exact} need not have a unique minimizer; instead, many, and in some cases infinitely many, distinct structures may attain the same minimum. These solutions form a \emph{solution set}, which under suitable regularity conditions may exhibit a lower-dimensional manifold structure within the design space. This places EM/RF synthesis within the broader class of \emph{nonlinear inverse problems}, where an input geometry or material distribution is inferred from indirect observations of its physical response \cite{engl1996regularization,colton2013inverse,molesky2018inverse}.

Despite its simple form, \eqref{eq:inverse_synthesis_exact} is challenging in several ways. First, this problem is ill-posed, meaning that the inverse map from S-parameters to layout is generally nonunique. Over a finite frequency band and a finite set of ports, many distinct geometries can produce nearly identical scattering responses. Thus, the practical goal is not to recover a unique structure, but to find one or more feasible layouts that satisfy the desired RF specifications. This nonuniqueness becomes especially pronounced for arbitrary or pixelated layouts, where many topologically different structures may realize similar matching, passband, or stopband behavior. Fig.~\ref{fig:IP} illustrates the many-to-one nature of \(\mathcal{E}\): among the multiple layouts that may satisfy the inverse problem, we seek solutions that are both physically realizable and interpretable.

Second, evaluating \(\mathcal{E}(x)\) can be computationally expensive. Full-wave EM analysis requires solving Maxwell's equations, or an equivalent partial differential or integral equation formulation, together with port excitations, material models, boundary conditions, and meshing rules. Therefore, each evaluation of \(\Phi(x)\) may require one or more costly frequency-domain simulations. This makes exhaustive or direct optimization over high-dimensional layout spaces impractical, particularly when \(x\) contains many geometric, material, or binary design variables.

Finally, gradient-based optimization requires sensitivity information. Even when \(\Phi(x)\) can be evaluated by an EM simulator, its gradient with respect to the layout depends on the Fr\'{e}chet derivative \(D\mathcal{E}(x)\), which describes how perturbations in geometry, topology, material distribution, or layout variables affect the simulated S-parameters. Adjoint-field methods can compute such sensitivities efficiently for many PDE-constrained EM design problems \cite{aghasi2011sensitivity,polydorides2012high, lalaukeraly2013adjoint,hughes2018adjoint}; however, they generally require access to internal field solutions and solver operators. This is often unavailable in black-box simulation workflows, motivating derivative-free, surrogate-assisted, and generative approaches for EM/RF synthesis.

\subsection{Prior Inverse Design Approaches}  
\begin{figure*}[t]
\begin{overpic}[
    width=\textwidth,
    percent,
    tics=5
]{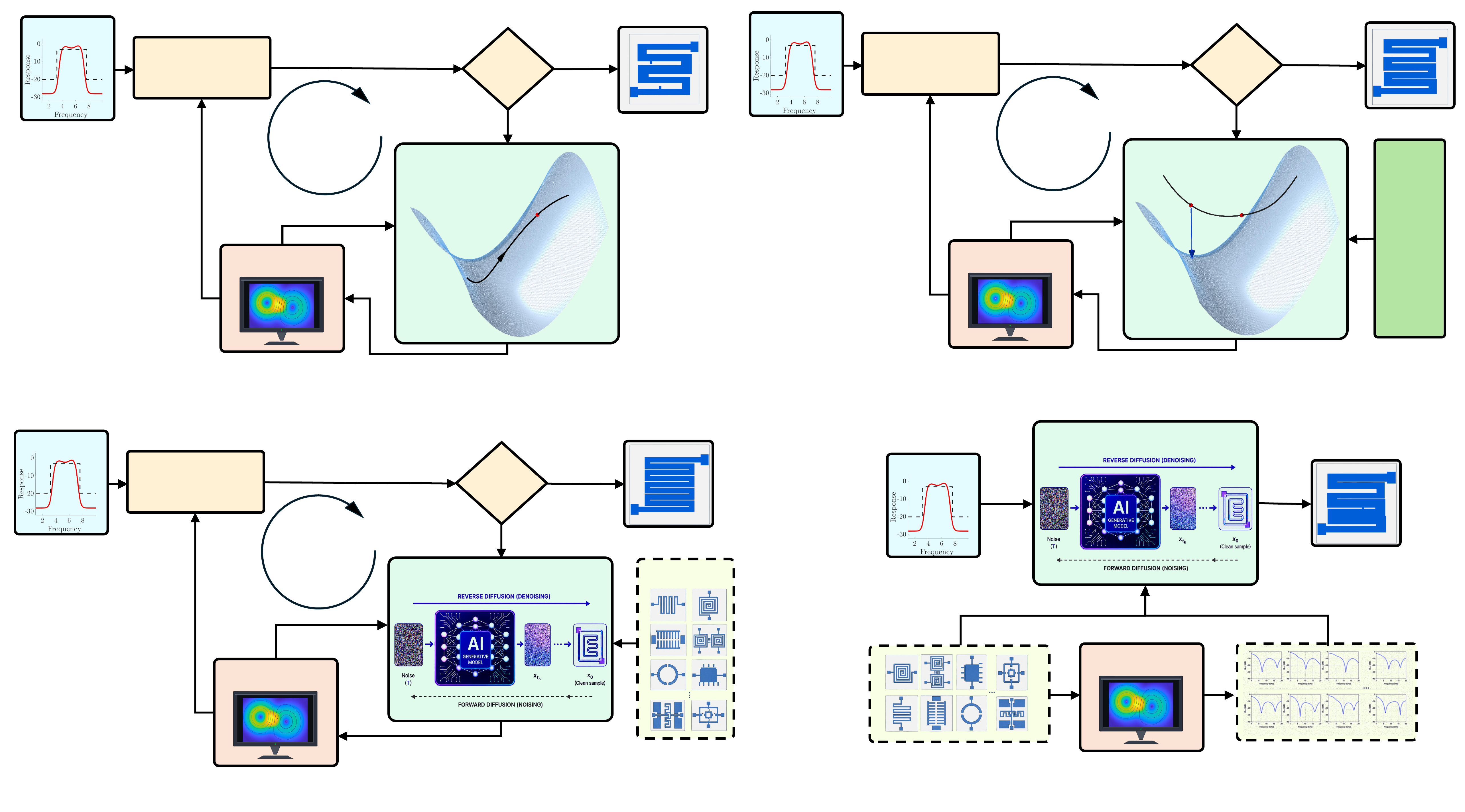}

\put(4.86,53.53){\makebox(0,0)[c]{\scalebox{.85}{\rotatebox{0}{$y$}}}}
\put(13.7,51.9){\makebox(0,0)[c]{\scalebox{.7}{\rotatebox{0}{Evaluate}}}}
\put(13.7,50){\makebox(0,0)[c]{\scalebox{.85}{\rotatebox{0}{$y-\mathcal{E}(x_{\theta_t})$}}}}
\put(34.92,50.96){\makebox(0,0)[c]{\scalebox{.75}{\rotatebox{0}{stop?}}}}
\put(39.5,51.5){\makebox(0,0)[c]{\scalebox{.7}{\rotatebox{0}{yes}}}}
\put(45.1,54.38){\makebox(0,0)[c]{\scalebox{.85}{\rotatebox{0}{$x_\theta $}}}}
\put(22.31,46.79){\makebox(0,0)[c]{\scalebox{.7}{\rotatebox{0}{Optimization}}}}
\put(22.31,44.79){\makebox(0,0)[c]{\scalebox{.7}{\rotatebox{0}{Loop}}}}
\put(19.3,37.68){\makebox(0,0)[c]{\scalebox{.7}{\rotatebox{0}{EM Simulator}}}}
\put(20,41.16){\makebox(0,0)[c]{\scalebox{.85}{\rotatebox{0}{$\mathcal{D}_\theta\mathcal{E}(x_\theta)$}}}}
\put(31.59,44){\makebox(0,0)[c]{\scalebox{.8}{\rotatebox{0}{update $\theta$}}}}
\put(32,42){\makebox(0,0)[c]{\scalebox{.85}{\rotatebox{0}{ $x_{\theta_{t-1}}\to x_{\theta_t} $}}}}
\put(11,35){\makebox(0,0)[c]{\scalebox{.85}{\rotatebox{0}{ $\mathcal{E}(x_\theta) $}}}}
\put(37,37){\makebox(0,0)[c]{\scalebox{.7}{\rotatebox{50}{ $\Phi$ Solution Set}}}}

\put(54.86,53.8){\makebox(0,0)[c]{\scalebox{.85}{\rotatebox{0}{$y$}}}}
\put(63.7,52.3){\makebox(0,0)[c]{\scalebox{.7}{\rotatebox{0}{Evaluate}}}}
\put(63.7,50.5){\makebox(0,0)[c]{\scalebox{.85}{\rotatebox{0}{$y-\mathcal{E}(x_{t})$}}}}
\put(84.5,51.1){\makebox(0,0)[c]{\scalebox{.75}{\rotatebox{0}{stop?}}}}
\put(89.5,51.8){\makebox(0,0)[c]{\scalebox{.7}{\rotatebox{0}{yes}}}}
\put(96.5,54.58){\makebox(0,0)[c]{\scalebox{.85}{\rotatebox{0}{$x $}}}}
\put(72,47){\makebox(0,0)[c]{\scalebox{.7}{\rotatebox{0}{Optimization}}}}
\put(72,45){\makebox(0,0)[c]{\scalebox{.7}{\rotatebox{0}{Loop}}}}
\put(69,37.68){\makebox(0,0)[c]{\scalebox{.7}{\rotatebox{0}{EM Simulator}}}}
\put(70,41.16){\makebox(0,0)[c]{\scalebox{.85}{\rotatebox{0}{$\mathcal{D}\mathcal{E}(x)$}}}}
\put(81.59,44){\makebox(0,0)[c]{\scalebox{.8}{\rotatebox{0}{update $x$}}}}
\put(82,42){\makebox(0,0)[c]{\scalebox{.85}{\rotatebox{0}{ $x_{t-1}\to x_{t} $}}}}
\put(61,35){\makebox(0,0)[c]{\scalebox{.85}{\rotatebox{0}{ $\mathcal{E}(x) $}}}}
\put(96.,43.5){\makebox(0,0)[c]{\scalebox{.85}{\rotatebox{0}{ $R(x) $}}}}
\put(95,38){\makebox(0,0)[c]{\scalebox{.75}{\rotatebox{90}{ structural}}}}
\put(97,38){\makebox(0,0)[c]{\scalebox{.75}{\rotatebox{90}{ regularization}}}}
\put(87,37){\makebox(0,0)[c]{\scalebox{.7}{\rotatebox{50}{ $\Phi$ Solution Set}}}}

\put(4.6,25.03){\makebox(0,0)[c]{\scalebox{.85}{\rotatebox{0}{$y$}}}}
\put(13.7,23.5){\makebox(0,0)[c]{\scalebox{.7}{\rotatebox{0}{Evaluate}}}}
\put(13.7,21.7){\makebox(0,0)[c]{\scalebox{.85}{\rotatebox{0}{$y-\mathcal{E}(x_{t})$}}}}
\put(34.25,22.46){\makebox(0,0)[c]{\scalebox{.75}{\rotatebox{0}{stop?}}}}
\put(39.,23.3){\makebox(0,0)[c]{\scalebox{.7}{\rotatebox{0}{yes}}}}
\put(45.3,26.2){\makebox(0,0)[c]{\scalebox{.85}{\rotatebox{0}{$x$}}}}
\put(46.7,16){\makebox(0,0)[c]{\scalebox{.7}{\rotatebox{0}{$x$ samples}}}}
\put(22.45,19.5){\makebox(0,0)[c]{\scalebox{.7}{\rotatebox{0}{Optimization/}}}}
\put(22.45,17.8){\makebox(0,0)[c]{\scalebox{.7}{\rotatebox{0}{Sampling}}}}
\put(22.31,16){\makebox(0,0)[c]{\scalebox{.7}{\rotatebox{0}{Loop}}}}
\put(18.8,9.3){\makebox(0,0)[c]{\scalebox{.7}{\rotatebox{0}{EM Simulator}}}}
\put(17,14){\makebox(0,0)[c]{\scalebox{.85}{\rotatebox{0}{$\mathcal{D}\mathcal{E}(x)$}}}}
\put(34,16.3){\makebox(0,0)[c]{\scalebox{.7}{\rotatebox{0}{Trained Diffusion Model}}}}
\put(30,2){\makebox(0,0)[c]{\scalebox{.7}{\rotatebox{0}{$p(x\mid y)\propto p(y\mid x)p(x)$}}}}
\put(11,7){\makebox(0,0)[c]{\scalebox{.85}{\rotatebox{0}{ $\mathcal{E}(x) $}}}}

\put(63.67,23.58){\makebox(0,0)[c]{\scalebox{.85}{\rotatebox{0}{$y$}}}}
\put(78.2,25.64){\makebox(0,0)[c]{\scalebox{.65}{\rotatebox{0}{Conditional Diffusion Model}}}}
\put(76,27.64){\makebox(0,0)[c]{\scalebox{.7}{\rotatebox{0}{$p(x\mid y)$}}}}
\put(92.92,25.19){\makebox(0,0)[c]{\scalebox{.85}{\rotatebox{0}{$x$}}}}
\put(61.8,12.46){\makebox(0,0)[c]{\scalebox{.7}{\rotatebox{0}{$x$ samples}}}}
\put(77.89,10.62){\makebox(0,0)[c]{\scalebox{.7}{\rotatebox{0}{EM Simulator}}}}
\put(82,14.5){\makebox(0,0)[c]{\scalebox{.7}{\rotatebox{0}{$(x_i,\mathcal{E}(x_i))$}}}}
\put(94.8,12.46){\makebox(0,0)[c]{\scalebox{.7}{\rotatebox{0}{$\mathcal{E}(x)$ samples}}}}

\put(20,29){\makebox(0,0)[c]{\scalebox{.85}{\rotatebox{0}{(a)}}}}

\put(80,29){\makebox(0,0)[c]{\scalebox{.85}{\rotatebox{0}{(b)}}}}

\put(20,0){\makebox(0,0)[c]{\scalebox{.85}{\rotatebox{0}{(c)}}}}

\put(80,0){\makebox(0,0)[c]{\scalebox{.85}{\rotatebox{0}{(d)}}}}
\end{overpic}
\caption{Inverse EM/RF design approaches: (a) template-based optimization reduces the search dimension while keeping the outcomes within the solution set of $\Phi$, but restricts the admissible layout space \cite{bandler1969optimization,bandler1994space,koziel2006space}; (b) free-form optimization expands the design space through structural regularization, which keeps the solutions close to the solution set of $\Phi$ \cite{engl1996regularization,stuart2010bayesian}; (c) diffusion-posterior methods combine a learned prior with simulator-based response guidance during sampling \cite{chung2023dps}; and (d) direct conditional generative models enable rapid inverse generation, with EM verification performed after sampling \cite{guo2025dallem,dreossi2026grayscale}.}

\label{fig:Methods}
\end{figure*}
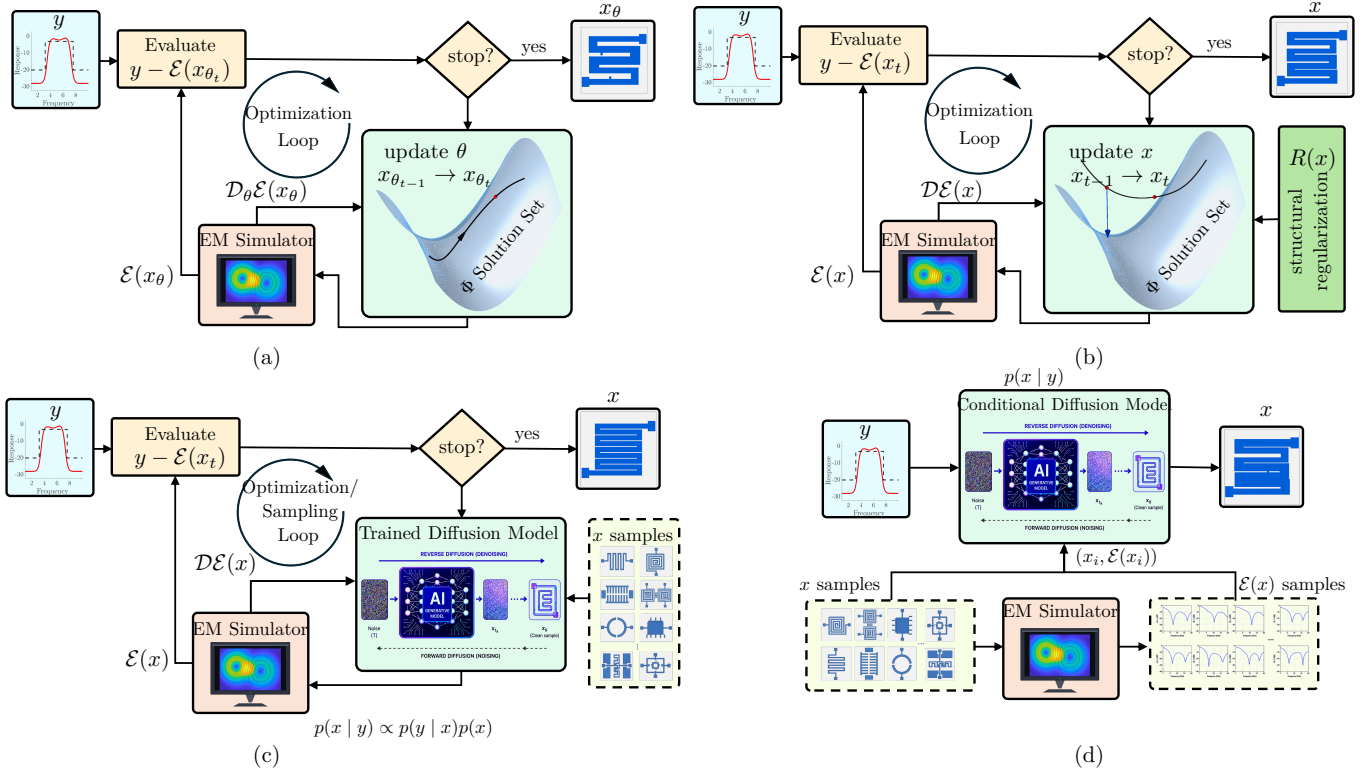

In classical RF/microwave design, the ill-posedness is mitigated by injecting prior information into the design space. This is often done by restricting \(x\) to a structured template \(x(\theta)\), where a circuit topology or EM layout is fixed, and only a limited set of physical parameters \(\theta\) is optimized. This strategy has a long history in microwave computer-aided design (CAD) and circuit optimization \cite{bandler1969optimization,bandler1988circuit}, including minimax formulations, direct EM-based optimization, and space-mapping methods that reduce the number of expensive full-wave simulations by combining accurate EM models with cheaper coarse models \cite{bandler1993minimax,bandler1994space,bandler1995electromagnetic,bakr2000space,koziel2006space}. Template-based and constrained optimization approaches are attractive because they incorporate designer knowledge, enforce manufacturable structures, and reduce the search dimension (Fig. \ref{fig:Methods}(a)). However, they may miss unconventional layouts and can require many simulator evaluations, careful initialization, or access to sensitivities. A complementary approach is to add explicit regularization and minimize 
$\Phi(x)+\lambda R(x)$,
where \(R(x)\) encodes smoothness, sparsity, connectivity, manufacturability, or similarity to admissible layouts \cite{engl1996regularization,stuart2010bayesian}. While regularization improves stability and helps select among many feasible solutions, its performance depends strongly on the choice of \(R\) and the parameter \(\lambda\), which may be difficult to tune for complex RF specifications (Fig. \ref{fig:Methods}(b)).

A probabilistic way to address ill-posedness is to model not a single inverse layout, but a distribution of plausible layouts conditioned on the desired response. In this view, \(p(x\mid y)\propto p(y\mid x)p(x)\), where \(p(x)\) encodes prior information about feasible EM/RF structures and \(p(y\mid x)\) quantifies consistency with the target scattering response. For example, under an additive Gaussian discrepancy model with variance \(\sigma^2\), the likelihood is related to \eqref{eq:inverse_synthesis_exact} as \(p(y\mid x)\propto \exp\!\left(-\Phi(x)/\sigma^2\right)\).
This formulation separates the structural prior from the EM-consistency term. Diffusion-posterior methods use a learned diffusion prior for \(p(x)\) and steer the sampling process using the likelihood or measurement-consistency term \cite{stuart2010bayesian,chung2023dps}. As shown in Fig. \ref{fig:Methods}(c), the EM simulator, or a physics-based forward model, can remain in the loop, improving consistency with the target response. However, this guidance generally requires repeated evaluations of \(\mathcal{E}\) and \(D\mathcal{E}(x)\), or a differentiable surrogate for them. For full-wave or commercial black-box EM simulators, this can be computationally expensive or inaccessible as stated earlier.

\begin{figure}[t]
    \centering
    \setlength{\fboxsep}{0pt}%
    \colorbox{white}{%
        \begin{overpic}[width=\linewidth]{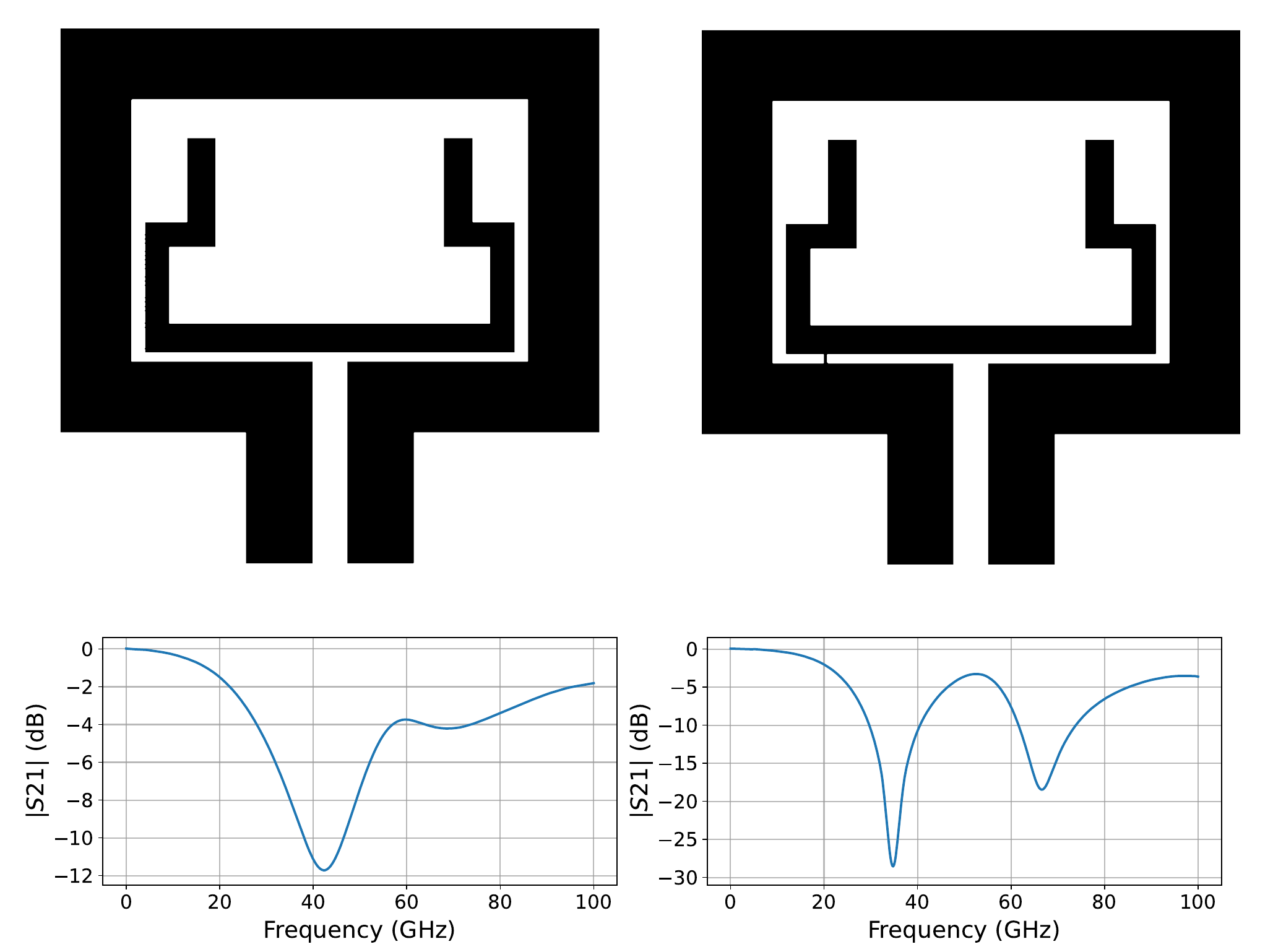}
            \put(50,50){%
                \makebox(30,-8){%
                    \tikz\draw[red, dashed, line width=1.0pt]
                        (0,0) circle [radius=0.4cm];
                }%
            }
        \end{overpic}%
    }
    \caption{Sensitivity of electromagnetic response to a small geometric perturbation. For the interleaved layout shown on the left, adding a single metal pixel, indicated by the dashed circle in the right panel, substantially changes the $S_{21}$ response. This observation underscores a fundamental difficulty in learning EM simulator outputs with standard neural-network surrogates, which often act primarily as interpolative models over the training distribution and may therefore fail to reliably capture sharp response variations caused by small layout modifications.}
    \label{fig:pixel_sensitivity}
\end{figure}

An alternative is to learn the inverse distribution directly, for example through a parametric model \(p_\theta(x\mid y)\) or a neural inverse map from scattering responses to design variables, as shown in Fig. \ref{fig:Methods}(d). Earlier microwave inverse-modeling works used neural networks, multivalued inverse models, invertible networks, or tandem architectures to address the nonuniqueness of circuit and antenna synthesis \cite{kabir2008neural_inverse,zhang2018multivalued,yu2020invertible,gupta2023tandem}. More recent generative approaches extend this idea to arbitrary or pixelated EM layouts, including conditional diffusion models for S-parameter-driven synthesis \cite{karahan2023deep_inverse_mmwave,karahan2024generalized_inverse,guo2025dallem,dreossi2026grayscale}. These methods can generate diverse candidates rapidly after training because they move the EM simulator outside the inner sampling loop. However, the learned conditional distribution is limited by the training data and may interpolate/extrapolate poorly to specifications or geometries outside the observed design distribution. Moreover, using a learned forward model \(F_\phi(x)\approx \mathcal{E}(x)\) as an EM emulator can be fragile in highly sensitive pixelated layouts, since standard neural-network surrogates often behave as data interpolators and may not reliably capture sharp changes in \(S\)-parameters caused by small local geometry changes, as demonstrated in Fig.~\ref{fig:pixel_sensitivity}. Consequently, generated candidates typically still require full-wave EM verification, filtering, or post-generation refinement.

In this paper, we introduce K-TRAIL, an automated EM/RF design framework that addresses the ill-posedness of scattering-response synthesis by exploiting a conditional diffusion model while keeping the EM simulator in the design loop (Fig. \ref{fig:ktrail}). Rather than seeking a unique inverse map from \(y\) to \(x\), the proposed approach uses a conditional generative model \(p_\theta(x\!\mid\! y)\) to produce plausible candidate layouts, while the forward simulator \(\mathcal{E}\) is used to evaluate and guide their scattering responses. This formulation is well suited to the case where \(y\) consists of prescribed scattering parameters, or their magnitudes, over a sampled frequency band, as detailed in Section~\ref{sec:exact_sparameter_synthesis}. We further consider a more general design setting in Section~\ref{sec:sparameter_conditions}, where the desired response is not specified through exact S-parameter values, but through RF design conditions (quantities such as gain, matching, isolation, bandwidth, or passband/stopband requirements). These conditions impose constraints on functions of the scattering parameters, and the objective is to find a layout that satisfies them. We show that this constrained synthesis problem can also be formulated as an optimization problem analogous to an inverse problem and addressed using K-TRAIL.

\section{Proposed Inverse Design for Exact S-parameter Targets}
\label{sec:exact_sparameter_synthesis}
We first consider the case in which the desired response is specified by exact scattering values, or by their magnitudes, over a sampled frequency band. In this setting, \(y\) denotes the target response vector supplied to the synthesis algorithm. When the full complex response is used, \(y\) may be formed by concatenating the real and imaginary parts of the relevant S-parameters, whereas when only magnitudes are prescribed, \(y\) contains samples such as \(|S_{11}(f)|\) and \(|S_{21}(f)|\). For notational simplicity, we write \(\mathcal{E}(x)\) for the simulator output expressed in the same form as \(y\).

The first component of the proposed approach is a conditional diffusion model trained from paired examples \(\{(x_0^{(j)},y^{(j)})\}_{j=1}^{n}\), where \(x_0^{(j)}\) is a feasible EM/RF layout and \(y^{(j)}=\mathcal{E}(x_0^{(j)})\) is its simulated scattering response. Similar conditional generative formulations have recently been explored for S-parameter-driven EM synthesis \cite{guo2025dallem,dreossi2026grayscale}, building on the broader diffusion-model framework \cite{ho2020denoising,song2021ddim}. Conceptually, a diffusion model learns to generate a layout by starting from noise and progressively removing that noise while being conditioned on the desired response \(y\). During training, noisy versions \(x_t\) of each clean layout \(x_0\) are generated according to
$$
p(x_t\mid x_0)
=
\mathcal{N}\left(
\sqrt{\bar{\alpha}_t}x_0,
(1-\bar{\alpha}_t)I
\right),
\qquad t=1,\ldots,T,
$$
where \(\bar{\alpha}_t\) specifies the amount of noise at diffusion step \(t\). A conditional denoising network then learns to estimate the corresponding clean layout from \(x_t\), the diffusion time \(t\), and the target response \(y\) as
$
\hat{x}_0(t)=D_\theta(x_t,t,y).
$

After training, generation begins from a random noisy state and repeatedly applies this denoising process, producing samples from an approximation of \(p_\theta(x\!\mid\! y)\). Because the inverse problem is ill-posed, different initial noise realizations can produce different layouts for the same target response. However, conditioning on \(y\) alone does not guarantee that a generated layout satisfies \(\mathcal{E}(x)\approx y\), particularly for pixelated EM structures in which small geometric changes can produce large changes in S-parameters.

We therefore introduce simulator feedback directly into the denoising process using a derivative-free ensemble Kalman guidance. During generation, \(N\) diffusion trajectories, or candidate layouts, are evolved in parallel. At selected reverse-diffusion times \(t\in\mathcal{S}\subseteq\{T,\ldots,1\}\), each trajectory first produces a predicted clean layout
$$
\hat{x}_0^{(i)}(t)=D_\theta(x_t^{(i)},t,y),
\qquad i=1,\ldots,N.
$$
Rather than simulating the noisy state \(x_t^{(i)}\), which does not generally represent a physical layout, the EM simulator evaluates the corresponding clean-layout estimate:
$$
z^{(i)}(t)=\mathcal{E}\!\left(\hat{x}_0^{(i)}(t)\right).
$$
Any required decoding, thresholding, meshing, or layout projection is included in this simulator-evaluation pipeline.

The resulting ensemble provides paired samples of layout and response variations. Their means are
$$
\bar{x}(t)=\frac{1}{N}\sum_{i=1}^{N}\hat{x}_0^{(i)}(t),
\qquad
\bar{z}(t)=\frac{1}{N}\sum_{i=1}^{N}z^{(i)}(t),
$$
and the empirical layout--response and response--response covariances are
\begin{align*}
C_{xz}(t)
&=
\frac{1}{N-1}\sum_{i=1}^{N}
\left(\hat{x}_0^{(i)}(t)-\bar{x}(t)\right)
\left(z^{(i)}(t)-\bar{z}(t)\right)^\top,\\
C_{zz}(t)
&=
\frac{1}{N-1}\sum_{i=1}^{N}
\left(z^{(i)}(t)-\bar{z}(t)\right)
\left(z^{(i)}(t)-\bar{z}(t)\right)^\top.
\end{align*}
Here, \(C_{xz}\) captures how variations in the generated layouts correlate with variations in their simulated responses. Thus, it provides an empirical estimate of which layout changes are likely to move the S-parameters in a desired direction, without differentiating the EM simulator.

Following ensemble Kalman methods for inverse problems \cite{evensen2003ensemble,iglesias2013ensemble} and recent ensemble Kalman diffusion guidance \cite{zheng2024enkg}, we define
$$
K_t
=
C_{xz}(t)
\left(C_{zz}(t)+\sigma_y^2 I\right)^{-1},
$$
where \(\sigma_y^2\) controls the assumed response tolerance and regularizes the update. If the S-parameter responses are normalized or whitened, we use the simplified choice \(\sigma_y^2=1\). Each predicted layout is then shifted according to its simulated response error:
$$
\tilde{x}_0^{(i)}(t)
=
\hat{x}_0^{(i)}(t)
+
\rho_t K_t\left(y-z^{(i)}(t)\right),
$$
where \(\rho_t\in[0,1]\) controls the strength of the simulator-based correction. The corrected estimate \(\tilde{x}_0^{(i)}(t)\) then replaces the diffusion model's original clean-layout prediction \(\hat{x}_0^{(i)}(t)\) when advancing to the next denoising step. Thus, the reverse diffusion process proceeds in the usual manner, but is periodically redirected using EM-simulator feedback:
$$
x_{t-1}^{(i)}
=
\mathrm{SamplerStep}
\left(
x_t^{(i)},\tilde{x}_0^{(i)}(t),t,y
\right).
$$
Here, \(\mathrm{SamplerStep}\) denotes the standard reverse-diffusion update from noise level \(t\) to \(t-1\); in our implementation, this update follows a DDIM sampler \cite{song2021ddim}.

The Kalman correction uses the current ensemble of simulated designs to estimate locally how layout variations affect the scattering response, without requiring \(D\mathcal{E}(x)\), adjoint fields, or a differentiable EM solver \cite{chung2023dps}. Basically, the simulator is used only as a black-box evaluator. Because full-wave simulations are expensive and early denoised states may not yet represent meaningful layouts, guidance is applied only at a sparse set of intermediate-to-late diffusion steps. The resulting \(N|\mathcal{S}|\) simulator evaluations provide a direct trade-off between computational cost and response consistency. After sampling, the final ensemble is simulated once more, and a design is selected either by minimum mismatch \(\|y-\mathcal{E}(x_0^{(i)})\|_2\) or by satisfying a prescribed response tolerance.

In the next section, we detail inverse generation using the K-TRAIL approach for inverse design of blocks for specific S-parameter targets. Later, in Section \ref{sec:Block-level Inverse Design}, we use the K-TRAIL approach to inverse-design an amplifier, where both input and output blocks are generated concurrently in order to satisfy amplifier-level constraints such as gain and input match.

\section{Inverse Design Experiments Using Proposed K-TRAIL Approach}\label{sec: FullInvDesExps}

\begin{figure}[t]
    \centering
    \setlength{\fboxsep}{0pt}%
    \colorbox{white}{%
        \begin{overpic}[width=\linewidth]{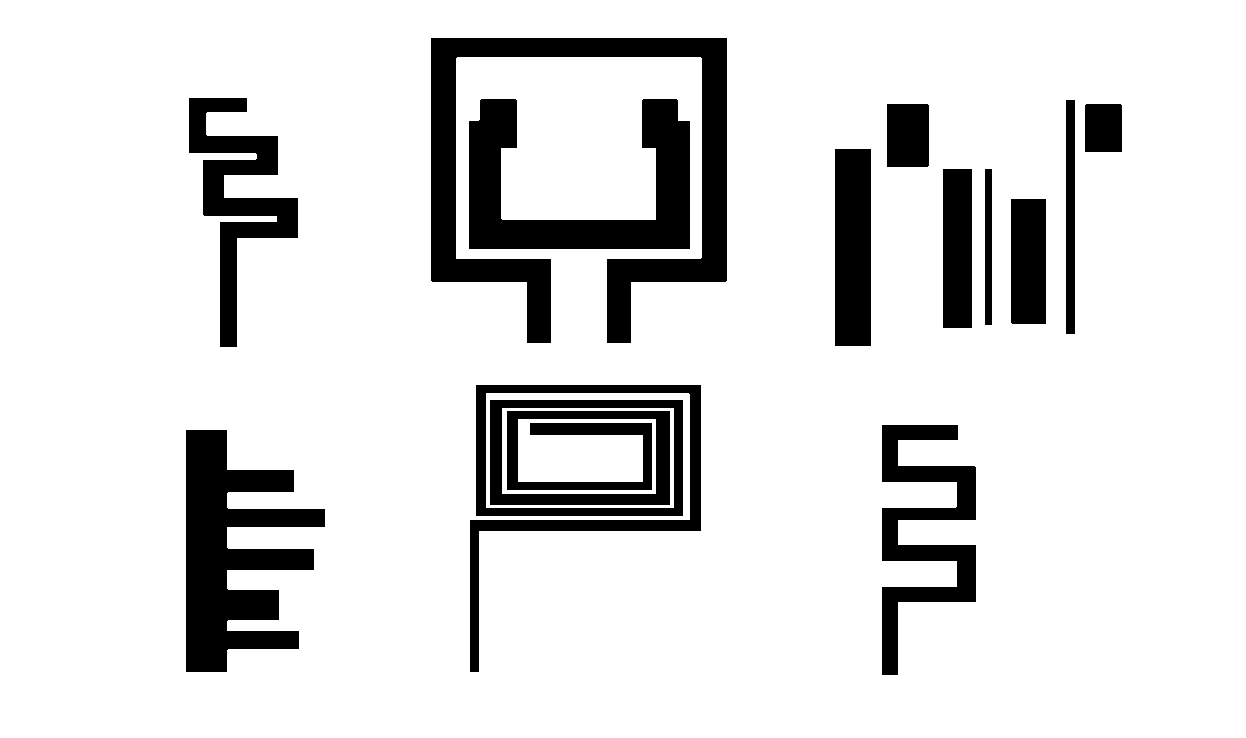}
        \end{overpic}%
    }\vspace{-.5cm}
    \caption{Layout templates used to train the diffusion model. From left to right and top to bottom: meandered line, interleaved, coupled line, stub tuner, spiral inductor, and uniform meandered line. A common 500 × 500 
    $\mu\mathrm{m}^2$ design region and fixed port
configurations are used for all topologies.}
    \label{fig:templates}
\end{figure}

In the following, we demonstrate the K-TRAIL approach for inverse design when the target includes full complex-valued S-parameters, and the practical design case where only target S-parameter magnitudes are available. Details of the data-factory setup, EM simulations performed using EMX \cite{CadenceEMX}, dataset construction, and training and validation procedures are provided in the Appendix. The training library contains six RF layout topologies shown in Fig.~\ref{fig:templates}: uniform meandered lines, nonuniform meandered lines, stub tuners, interleaved transformers, spiral inductors, and coupled-line segments. Following model training (detailed in Appendix), the conditional diffusion model is used to generate designs conditioned on target S-parameters, both with and without ensemble Kalman guidance.

\begin{figure*}[!htbp]
    \centering

    \includegraphics[
        height=.17\linewidth,
        trim={0.25cm 0.10cm 0.15cm 0.10cm},
        clip
    ]{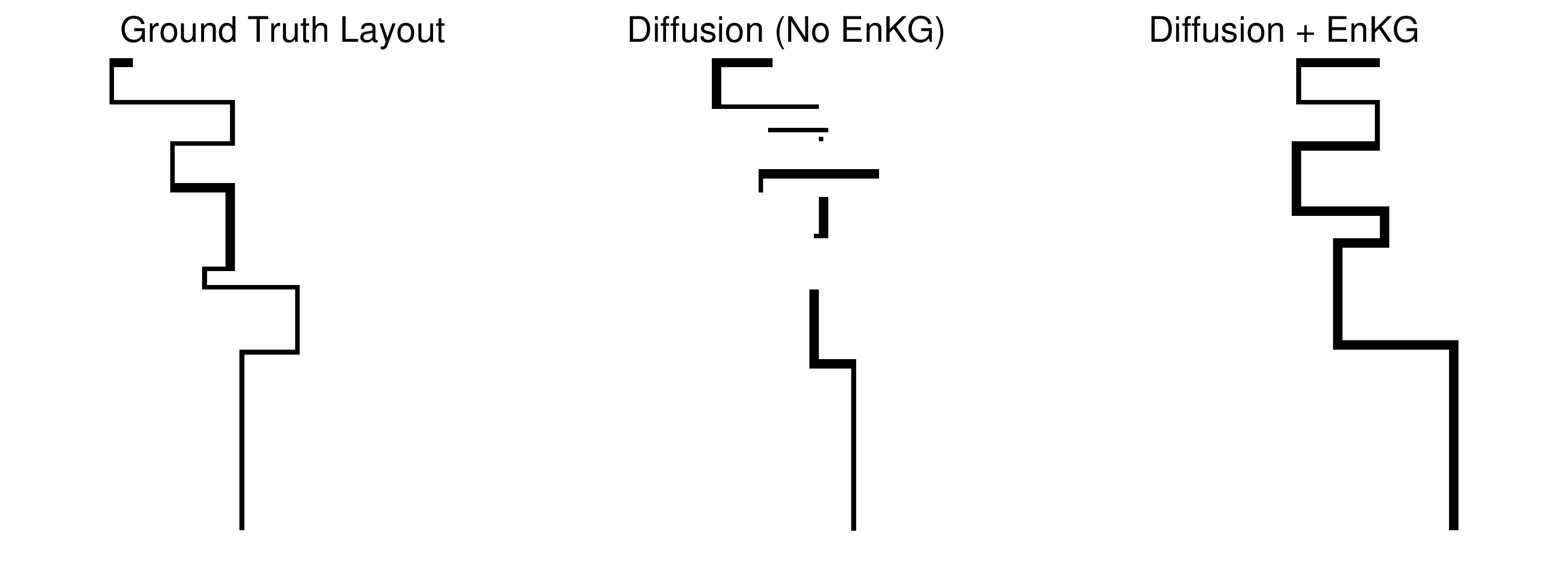}%
    \includegraphics[height=.16\linewidth]
    {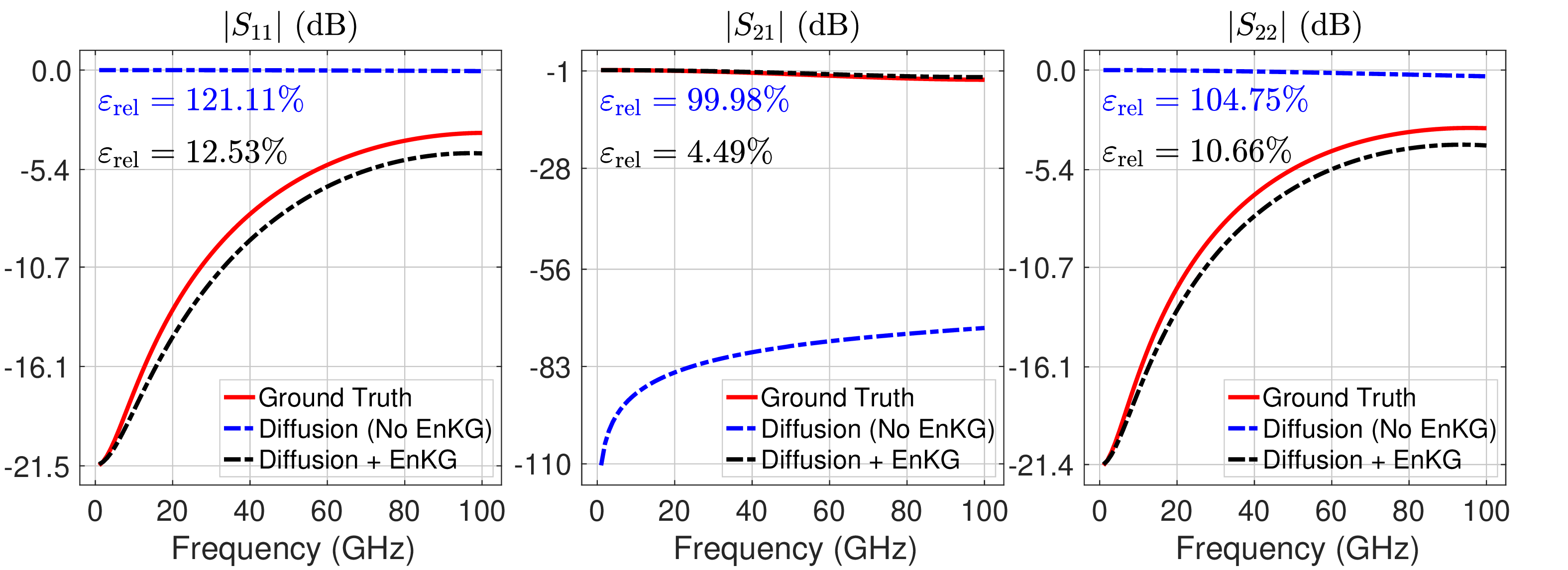}\par\vspace{0.15cm}

    \includegraphics[
        height=.17\linewidth,
        trim={0.25cm 0.10cm 0.15cm 0.10cm},
        clip
    ]{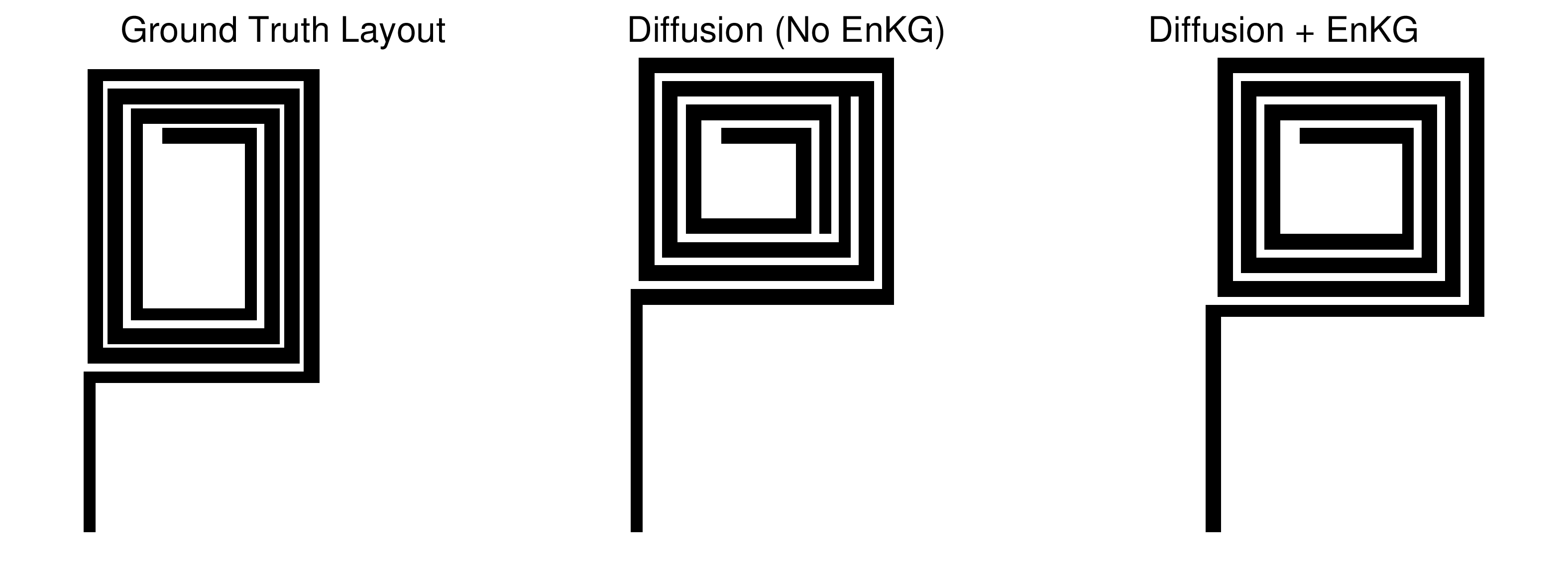}%
    \includegraphics[height=.17\linewidth]
    {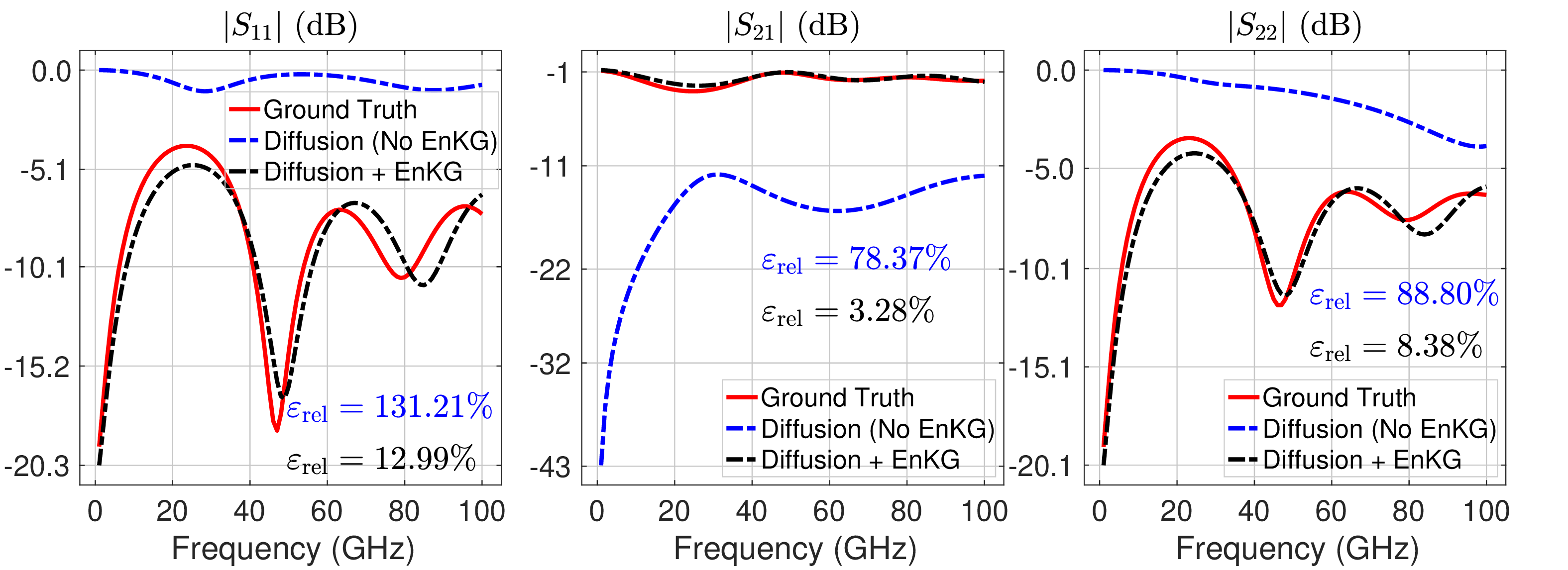}\par\vspace{0.15cm}

    \includegraphics[
        height=.17\linewidth,
        trim={0.25cm 0.10cm 0.15cm 0.10cm},
        clip
    ]{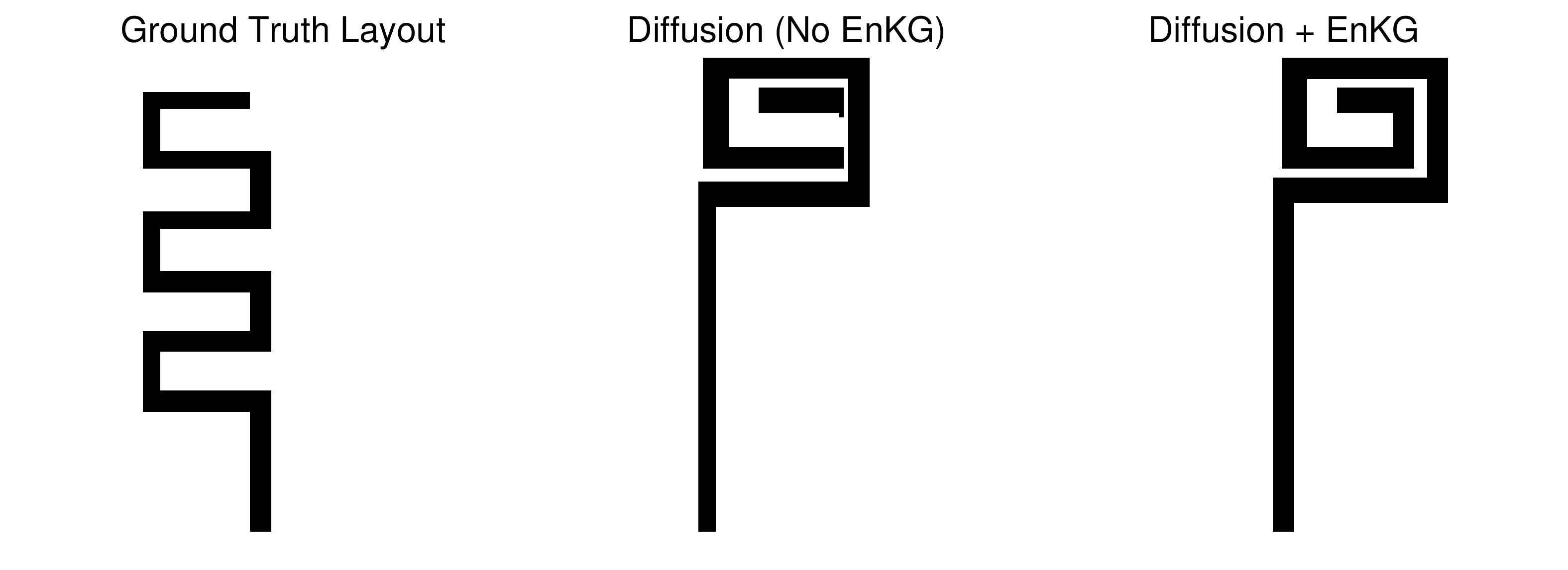}%
    \includegraphics[height=.17\linewidth]
    {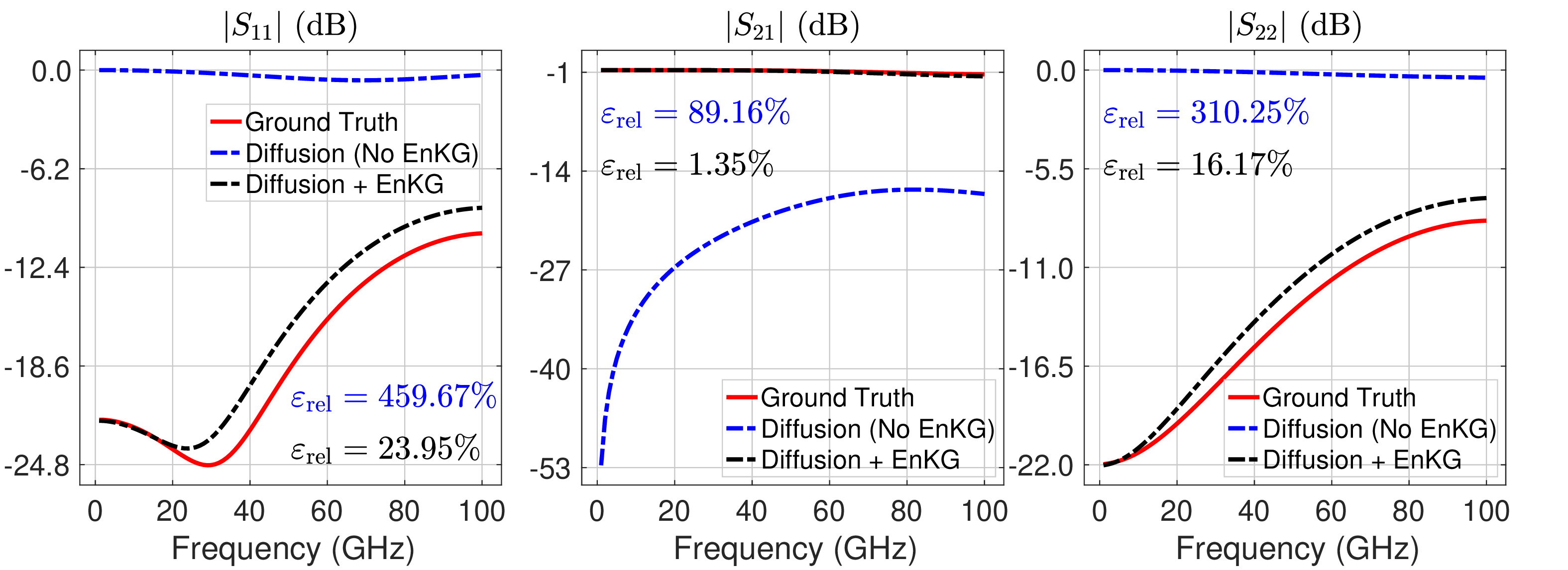}\par\vspace{0.15cm}

    \includegraphics[
        height=.17\linewidth,
        trim={0.25cm 0.10cm 0.15cm 0.10cm},
        clip
    ]{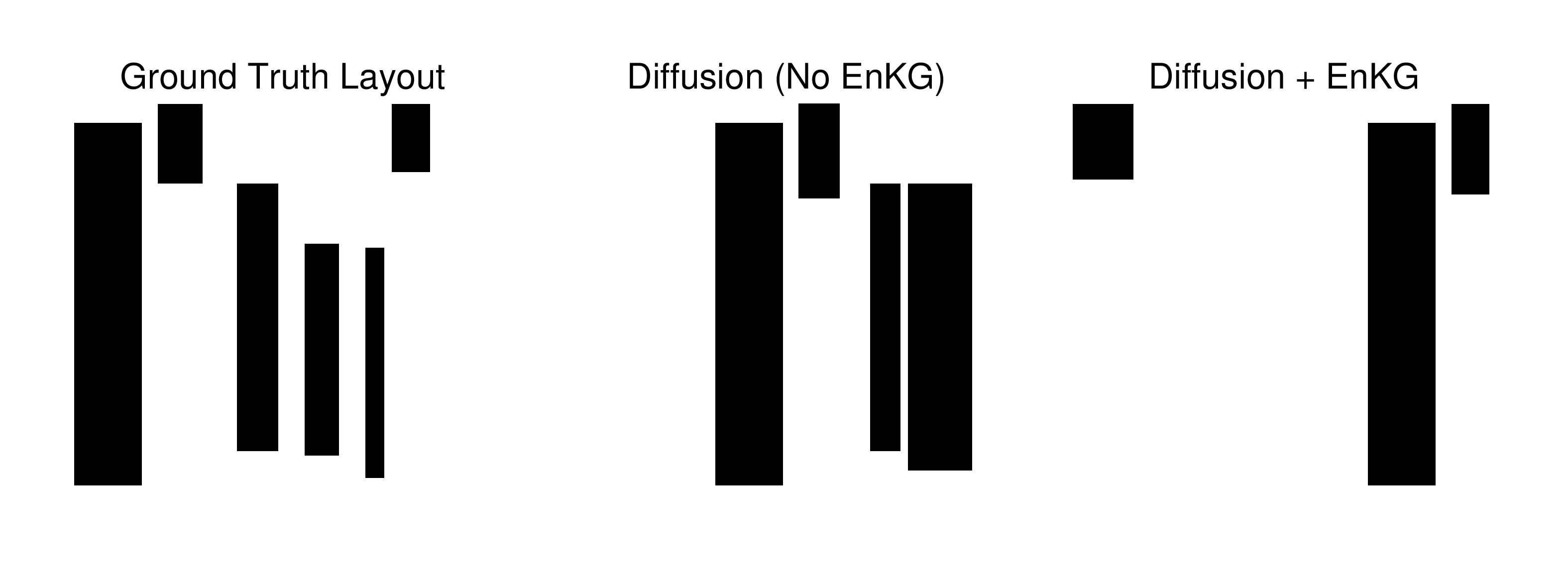}%
    \includegraphics[height=.17\linewidth]
    {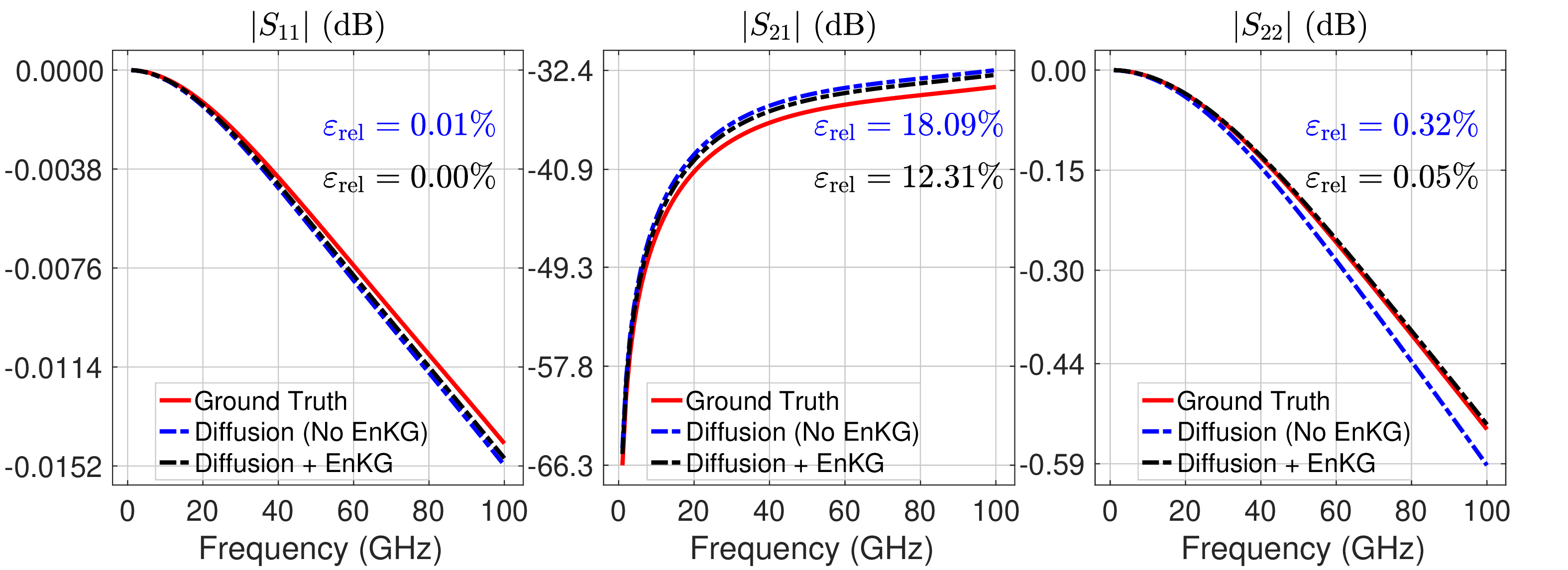}

        \includegraphics[
        height=.17\linewidth,
        trim={0.25cm 0.10cm 0.15cm 0.10cm},
        clip
    ]{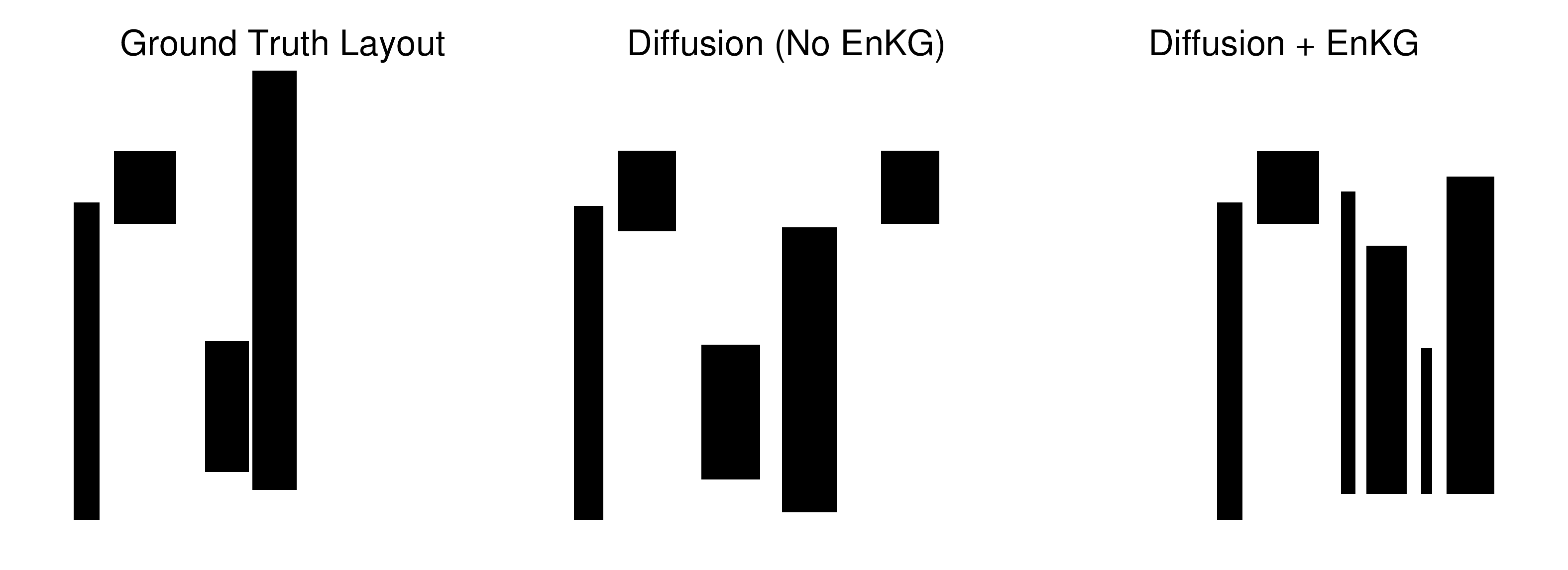}%
    \includegraphics[height=.17\linewidth]
    {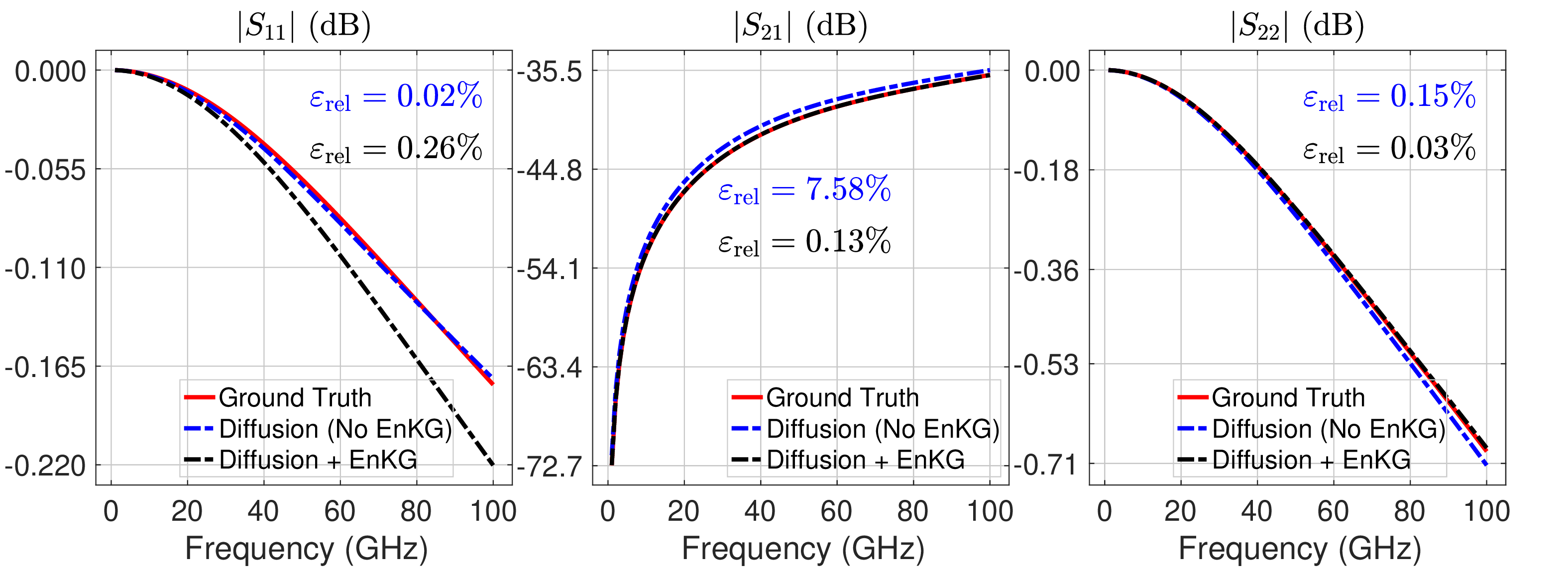}

            \includegraphics[
        height=.17\linewidth,
        trim={0.25cm 0.10cm 0.15cm 0.10cm},
        clip
    ]{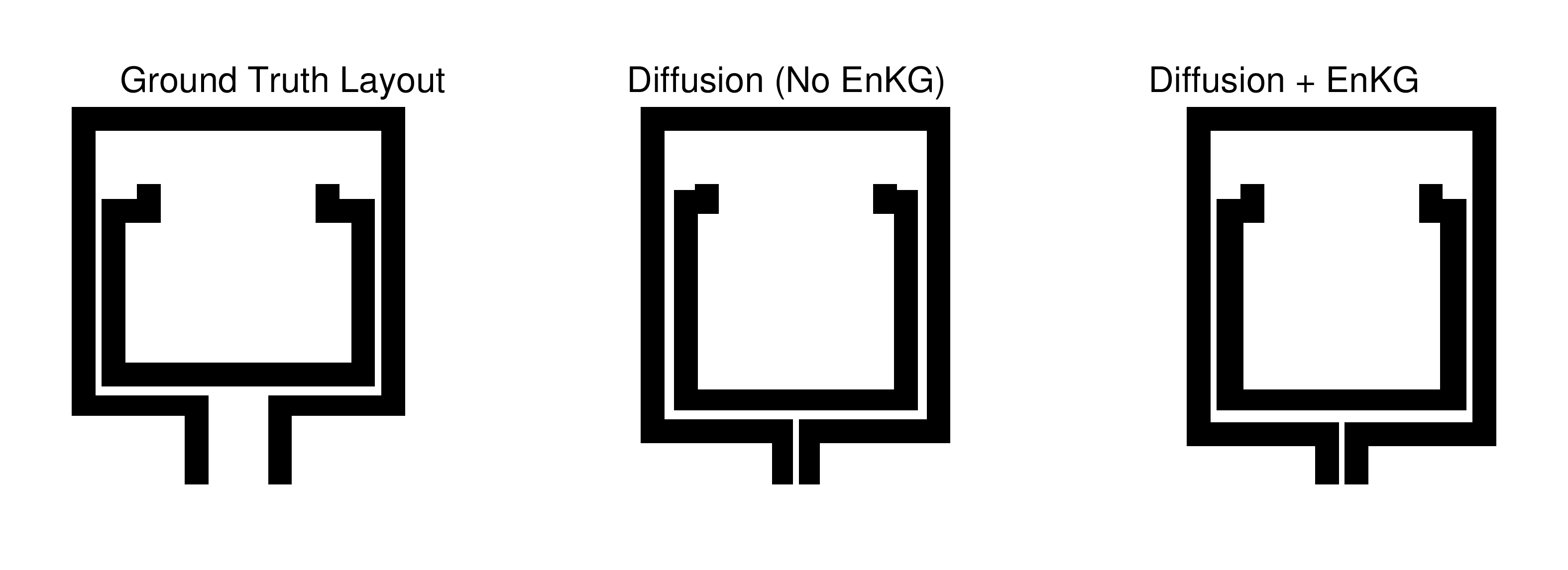}%
    \includegraphics[height=.17\linewidth]
    {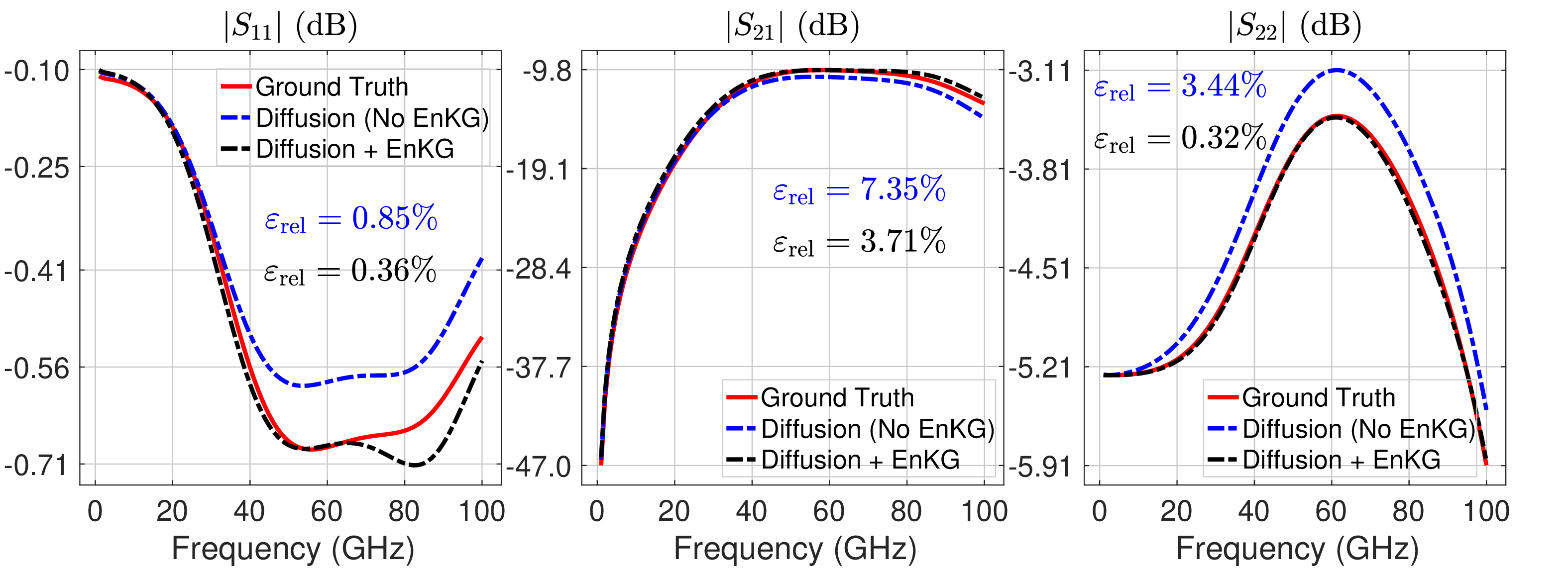}

            \includegraphics[
        height=.17\linewidth,
        trim={0.25cm 0.10cm 0.15cm 0.10cm},
        clip
    ]{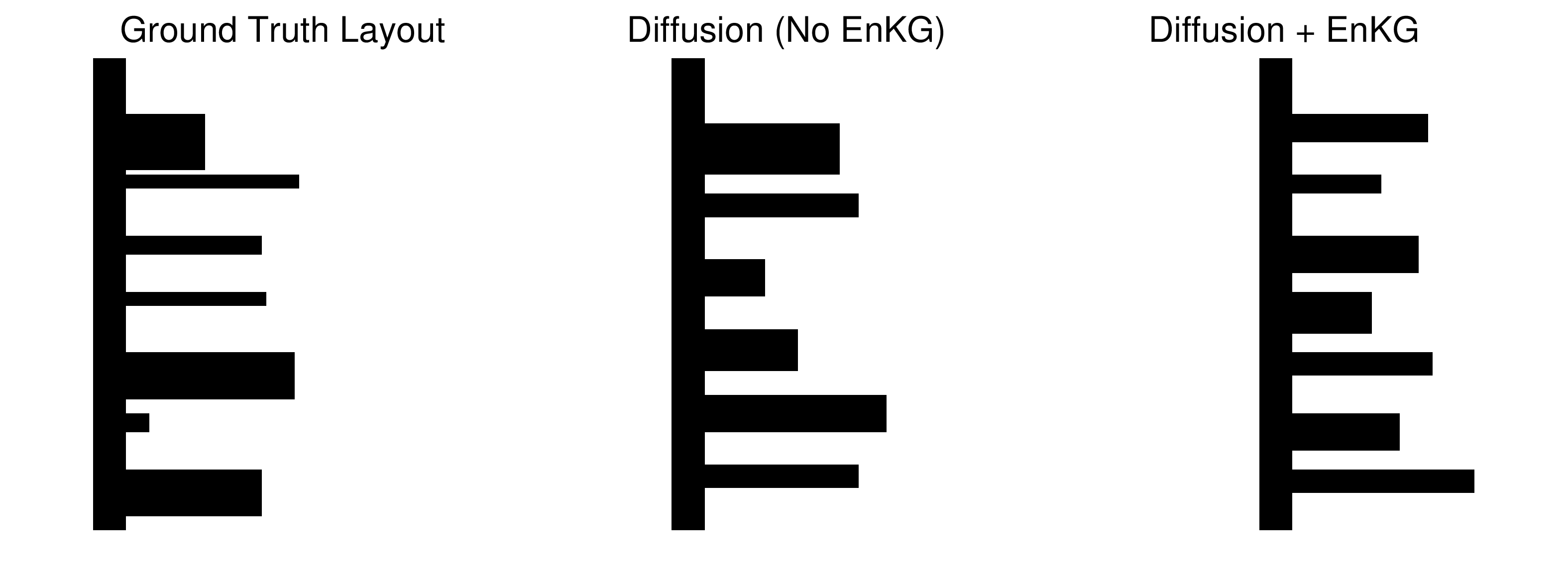}%
    \includegraphics[height=.17\linewidth]
    {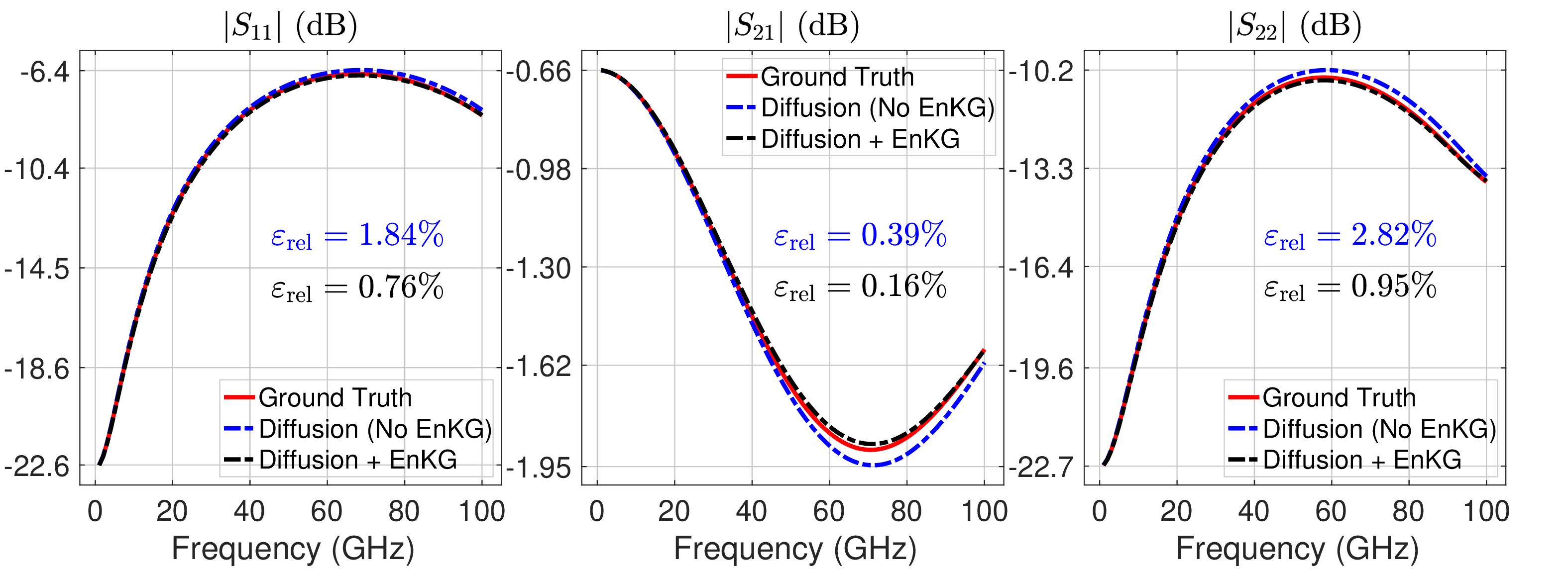}

    \caption{Representative simulator-guided synthesis results where $y$ consists of full complex S-parameter values. For each experiment, the layout comparison shows the ground-truth design, the unguided diffusion sample, and the diffusion sample corrected using ensemble Kalman guidance. The corresponding S-parameter responses show that simulator feedback substantially improves agreement with the target response relative to unguided diffusion.}
    \label{fig:full-exp}
\end{figure*}

\begin{figure*}[!htbp]
    \centering

    \includegraphics[
        height=.17\linewidth,
        trim={0.25cm 0.10cm 0.15cm 0.10cm},
        clip
    ]{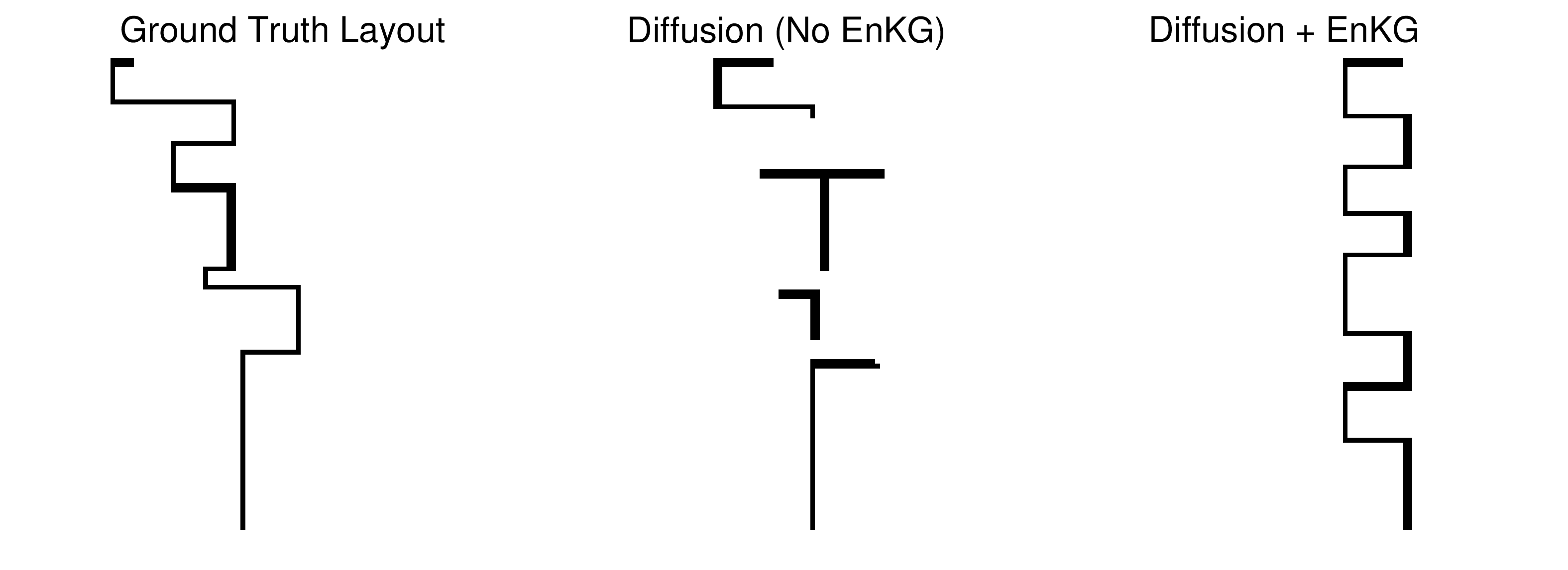}%
    \includegraphics[height=.16\linewidth]
    {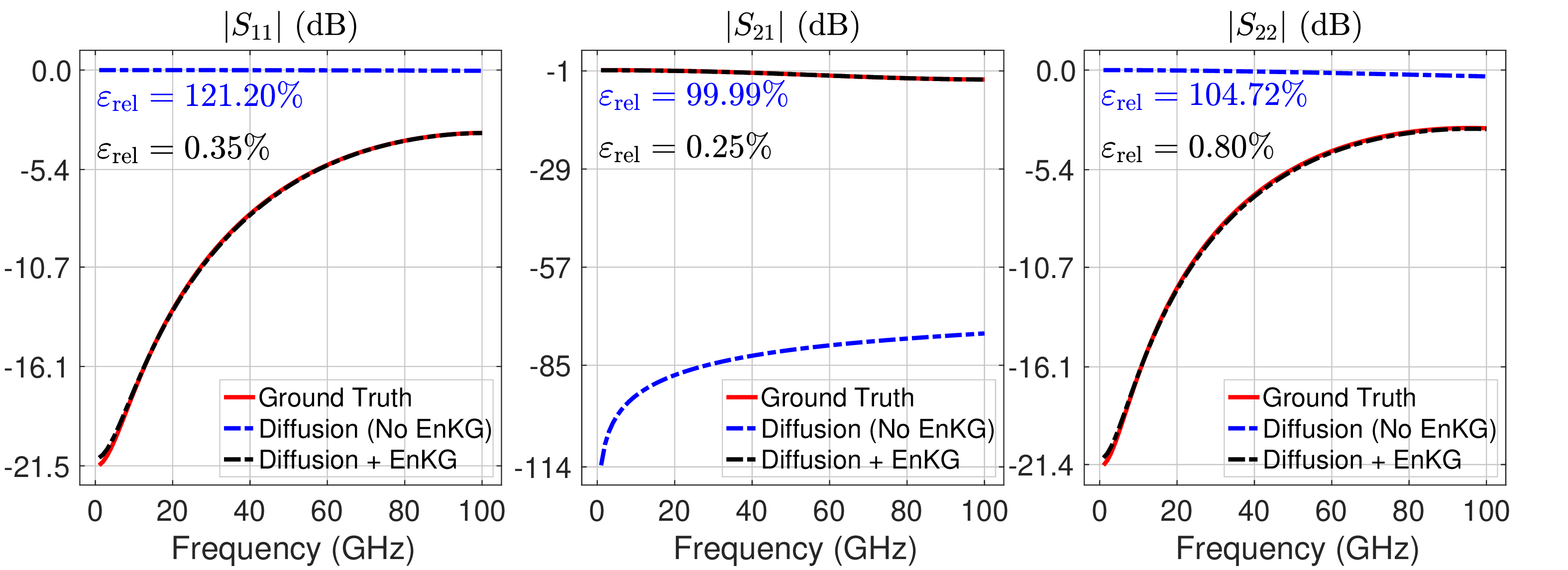}\par\vspace{0.15cm}

    \includegraphics[
        height=.17\linewidth,
        trim={0.25cm 0.10cm 0.15cm 0.10cm},
        clip
    ]{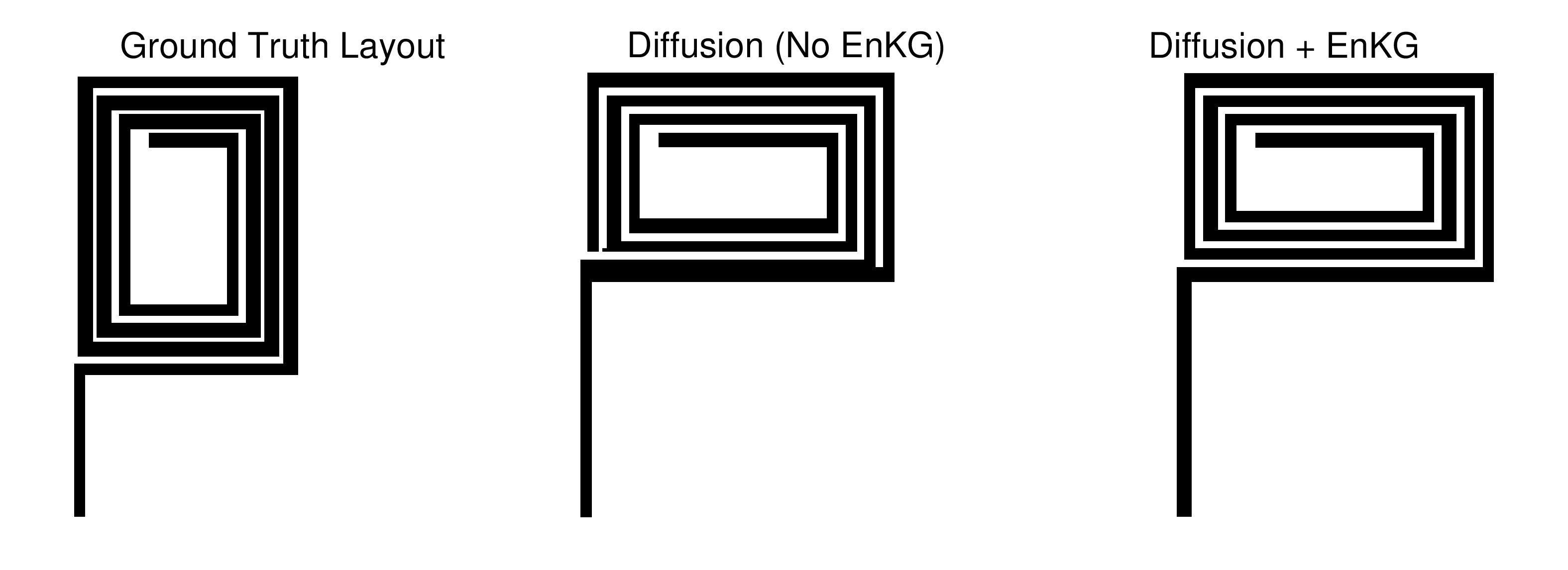}%
    \includegraphics[height=.17\linewidth]
    {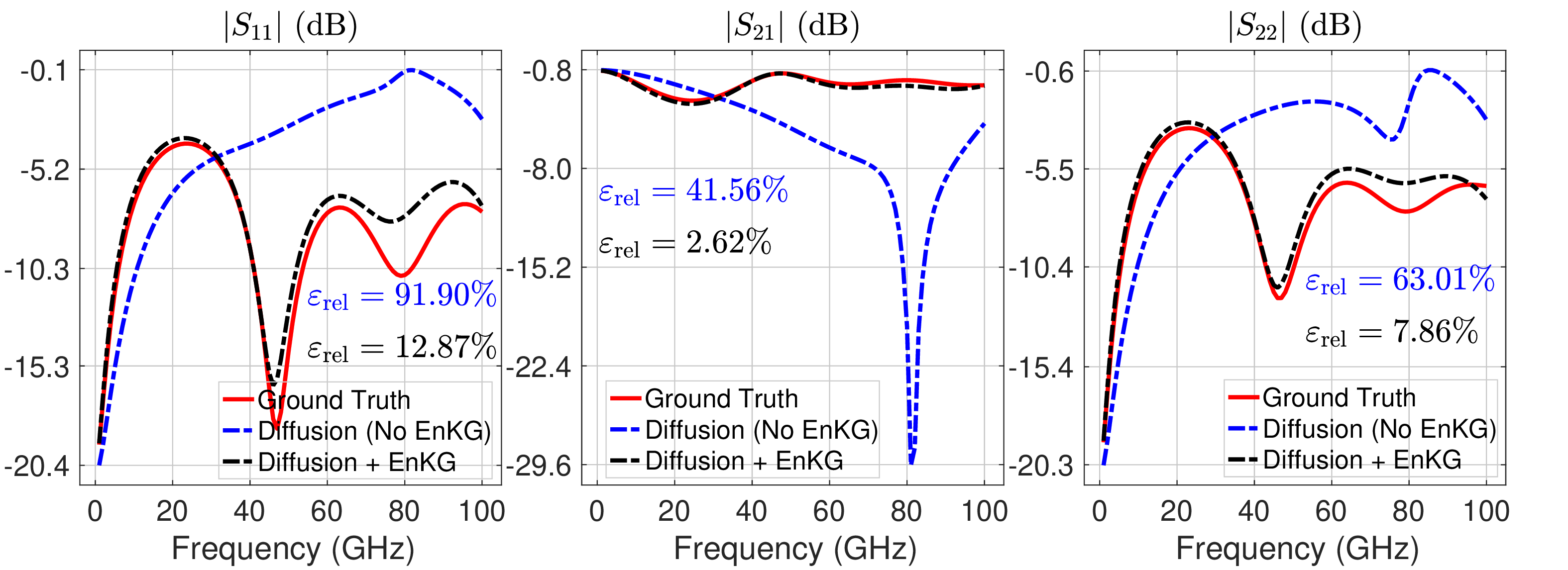}\par\vspace{0.15cm}

    \includegraphics[
        height=.17\linewidth,
        trim={0.25cm 0.10cm 0.15cm 0.10cm},
        clip
    ]{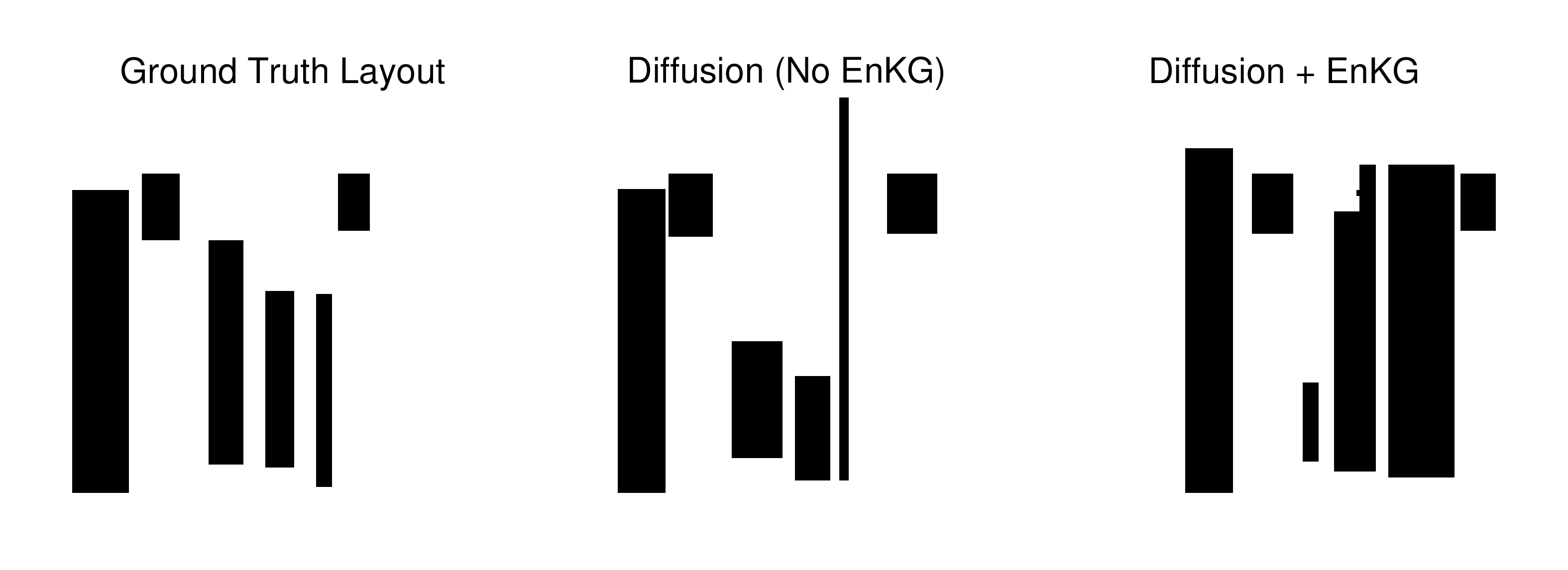}%
    \includegraphics[height=.17\linewidth]
    {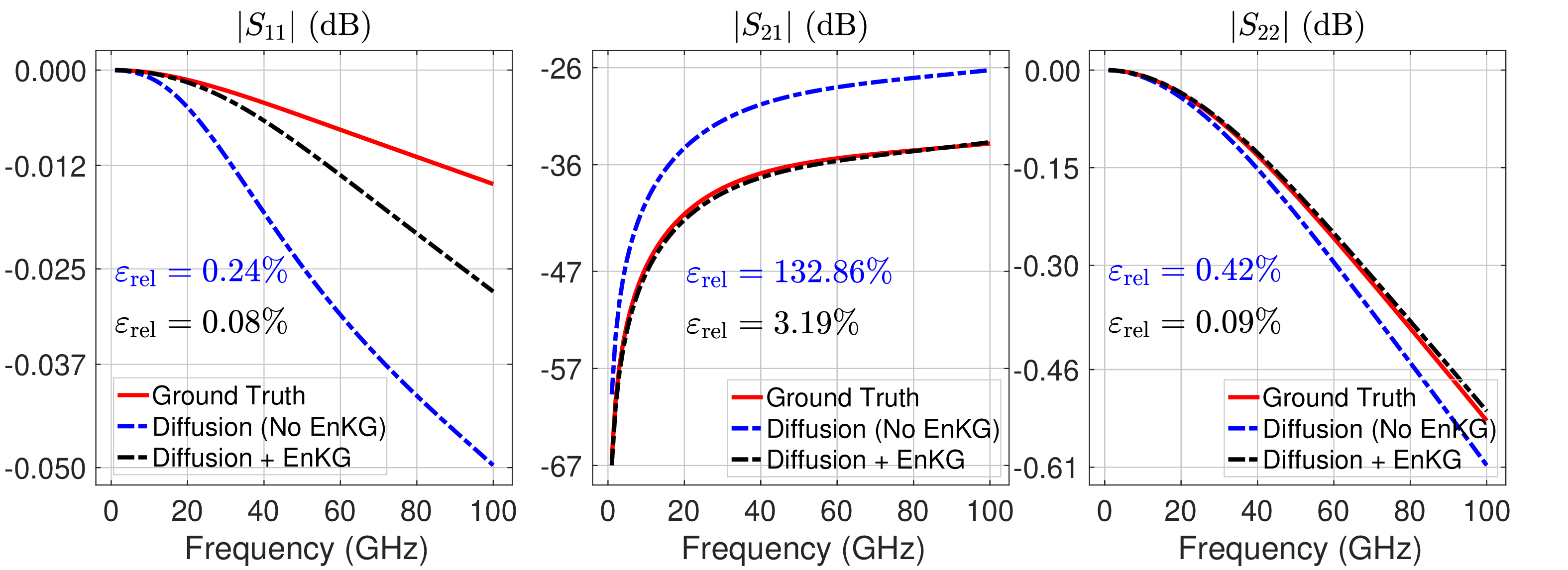}

        \includegraphics[
        height=.17\linewidth,
        trim={0.25cm 0.10cm 0.15cm 0.10cm},
        clip
    ]{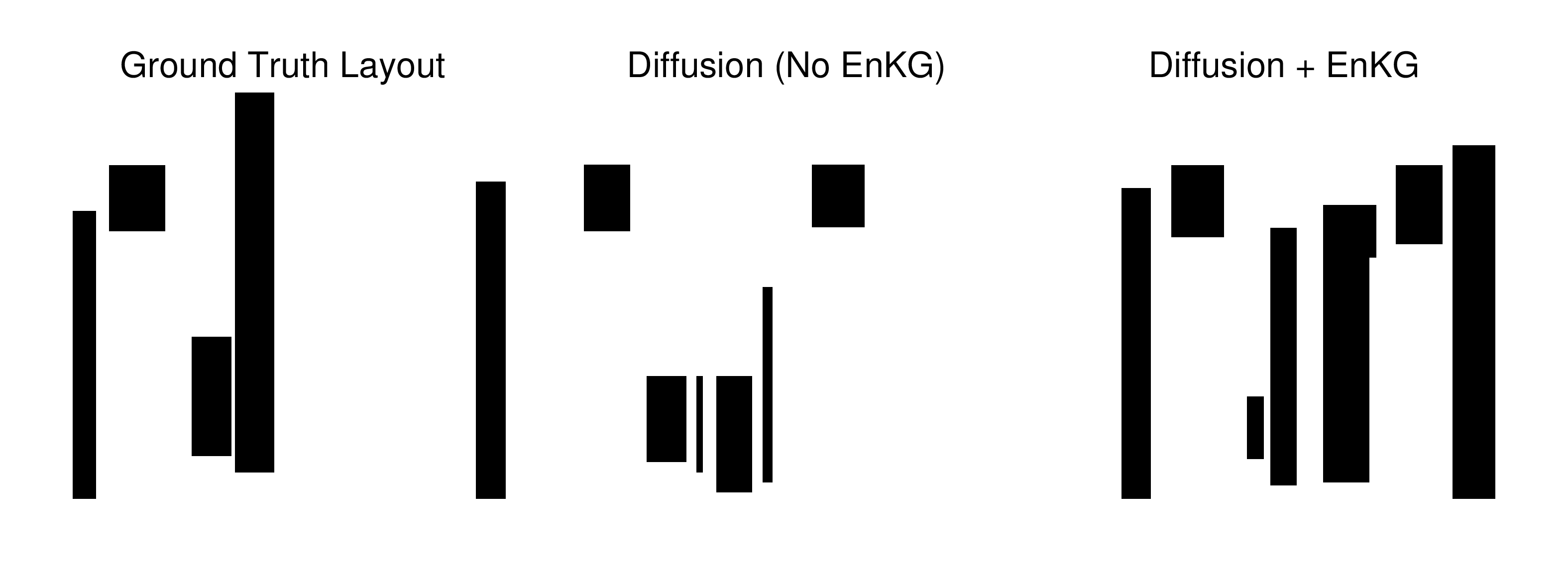}%
    \includegraphics[height=.17\linewidth]
    {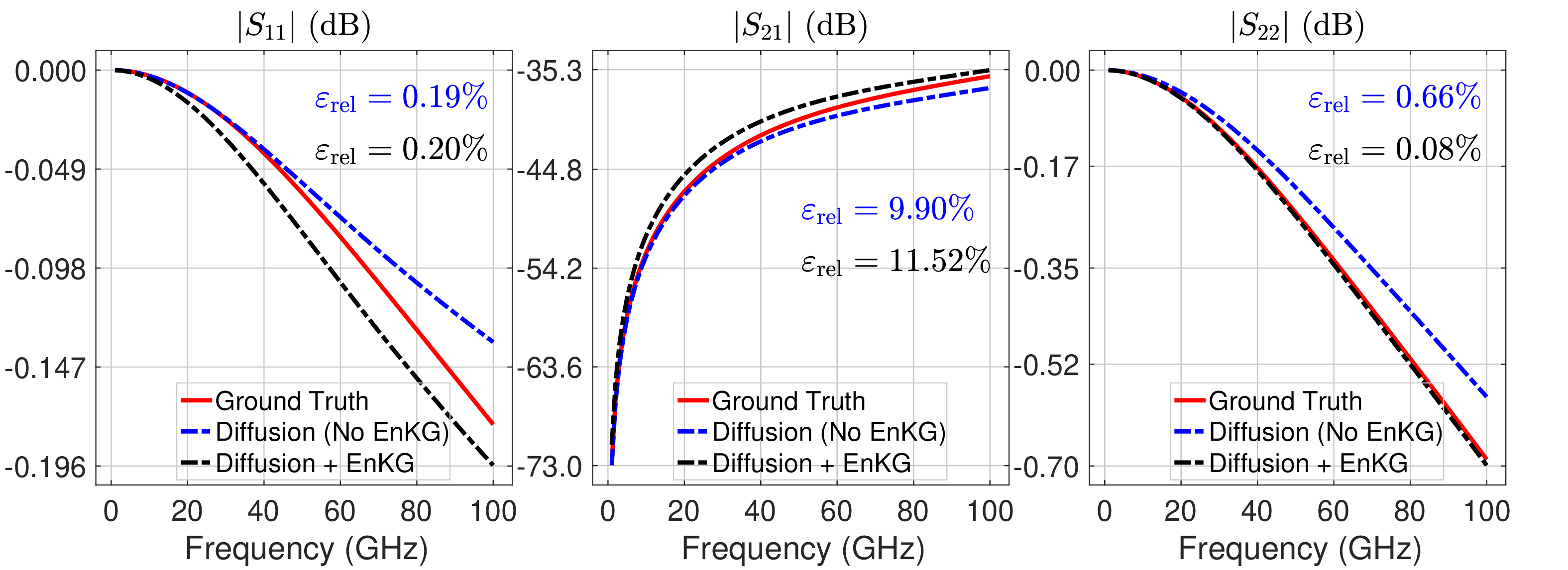}

            \includegraphics[
        height=.17\linewidth,
        trim={0.25cm 0.10cm 0.15cm 0.10cm},
        clip
    ]{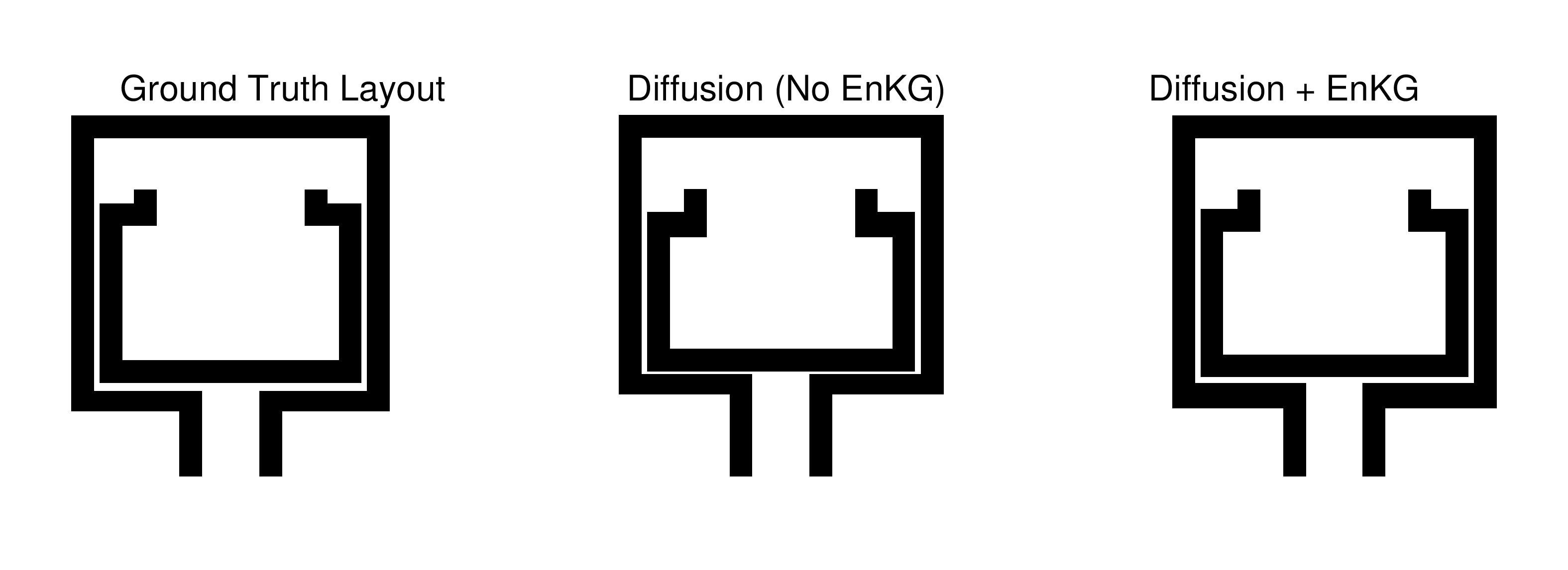}%
    \includegraphics[height=.17\linewidth]
    {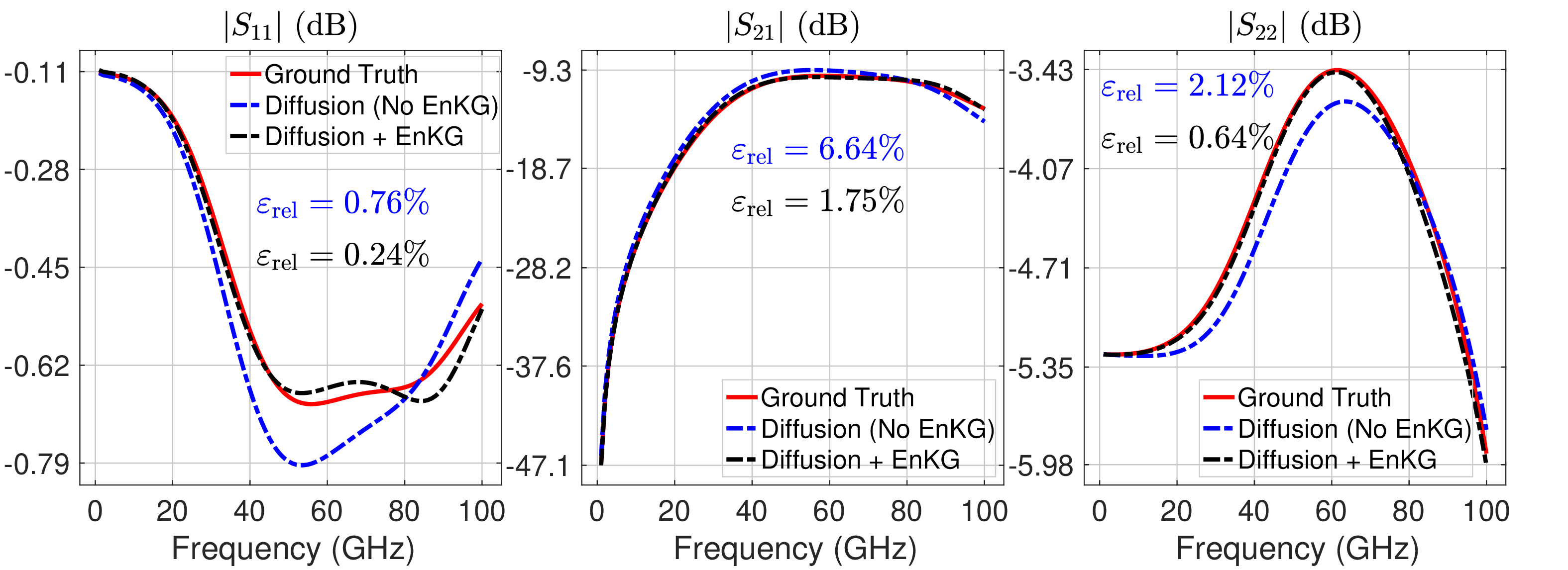}

        \includegraphics[
        height=.17\linewidth,
        trim={0.25cm 0.10cm 0.15cm 0.10cm},
        clip
    ]{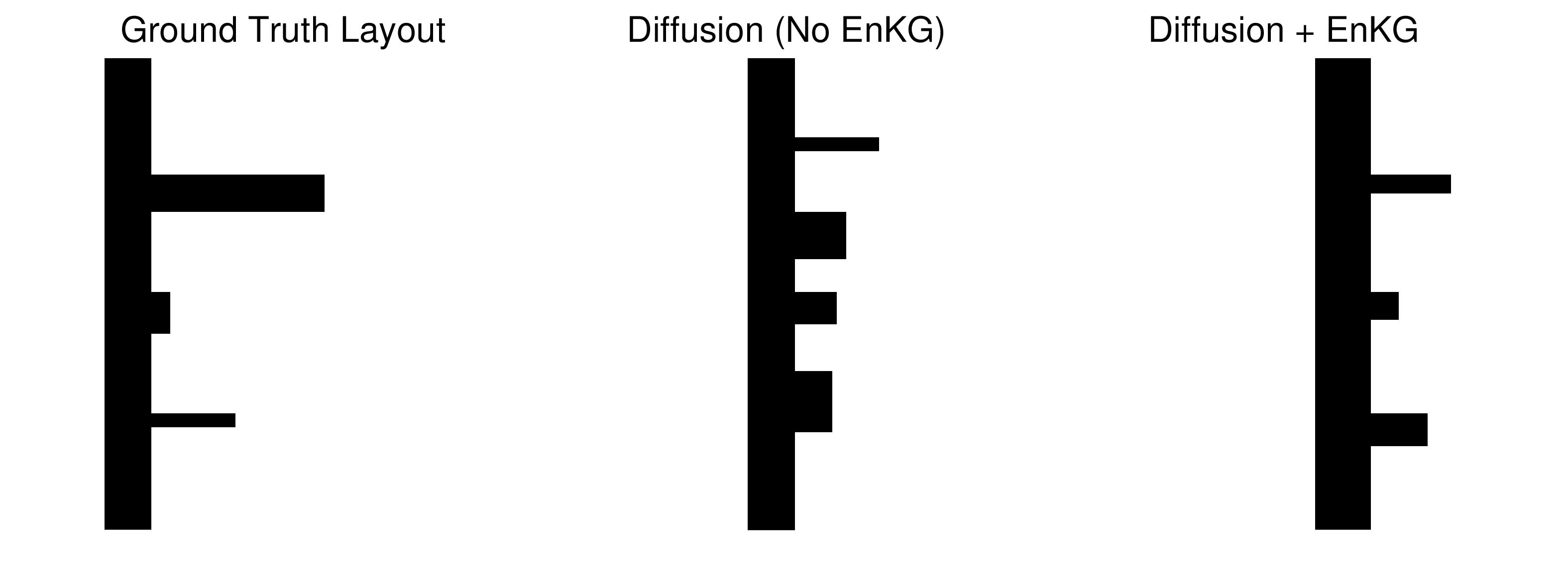}%
    \includegraphics[height=.17\linewidth]
    {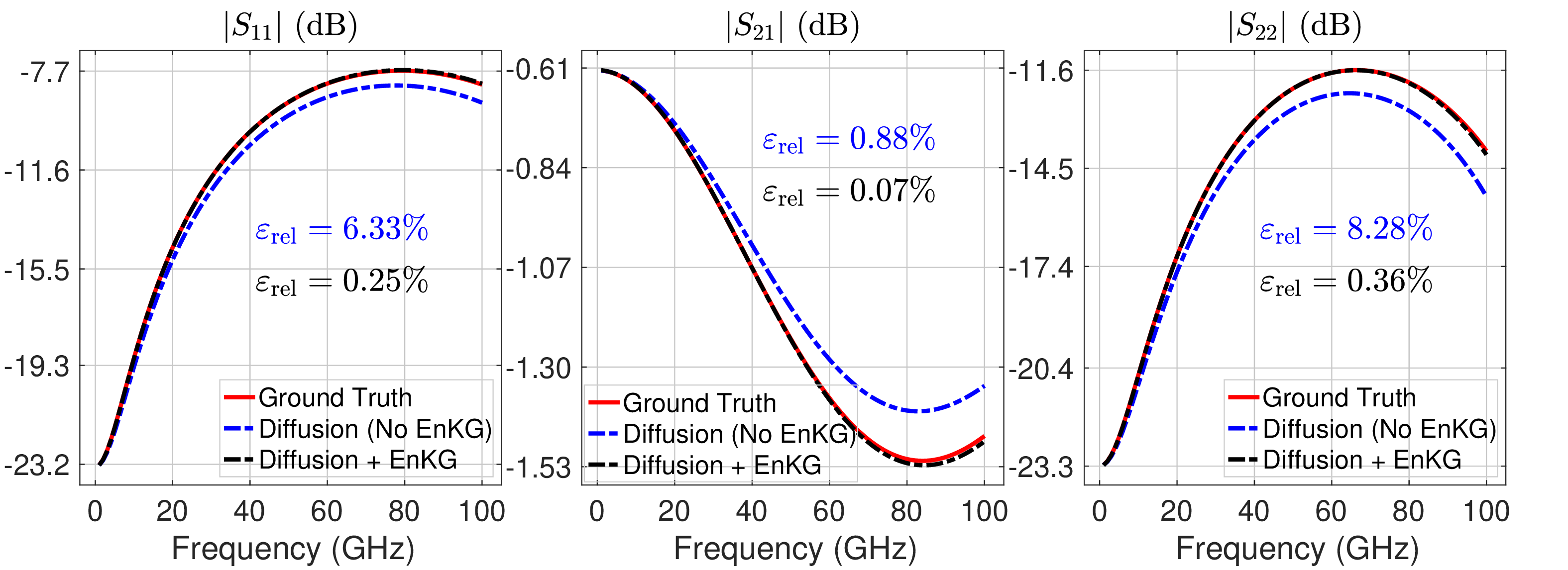}\par\vspace{0.15cm}

            \includegraphics[
        height=.17\linewidth,
        trim={0.25cm 0.10cm 0.15cm 0.10cm},
        clip
    ]{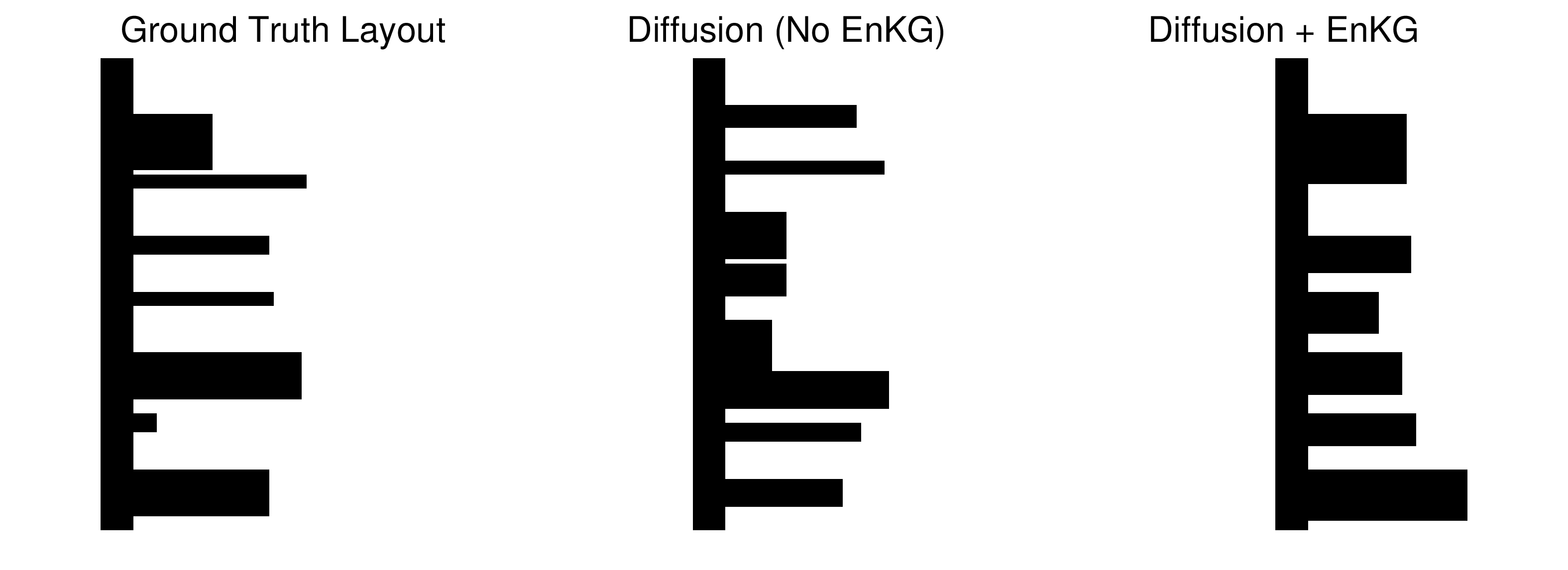}%
    \includegraphics[height=.17\linewidth]
    {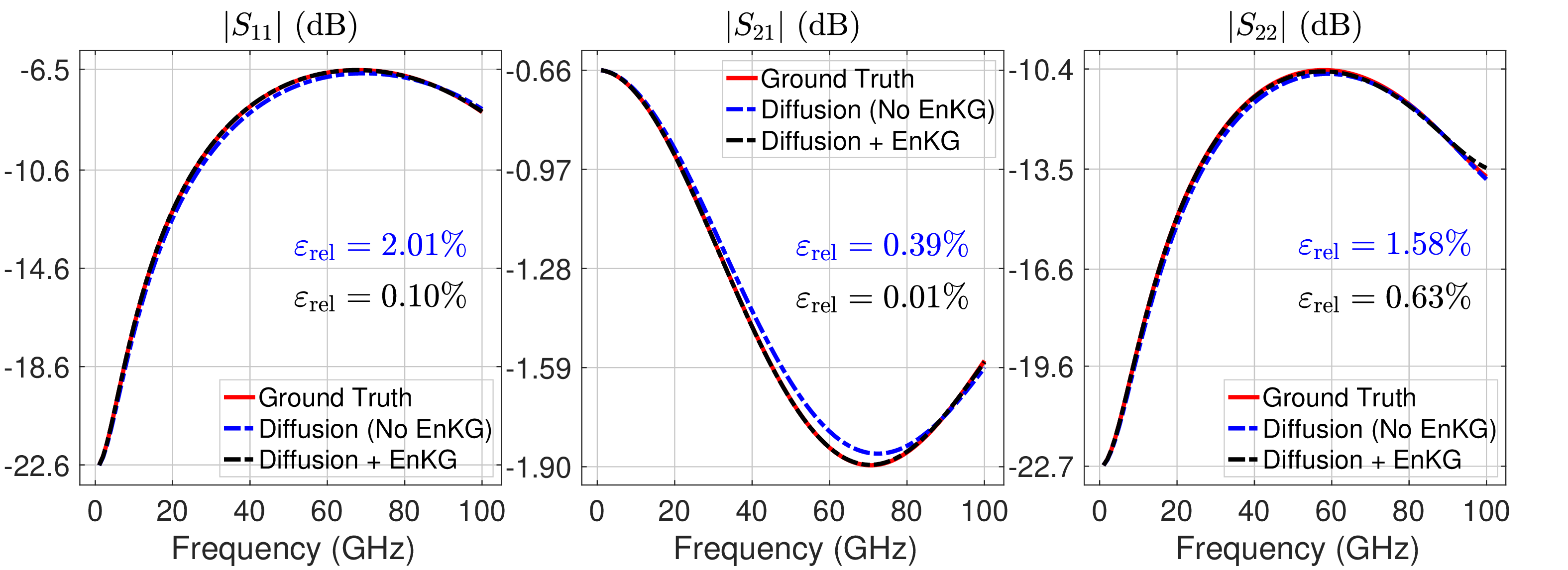}

    \caption{Representative simulator-guided synthesis results where $y$ consists of only the magnitude of the S-parameter values. For each experiment, the layout comparison shows the ground-truth design, the unguided diffusion sample, and the diffusion sample corrected using ensemble Kalman guidance. The corresponding S-parameter responses show that simulator feedback substantially improves agreement with the target response relative to unguided diffusion.}
    \label{fig:mag-exp}
\end{figure*}

In the experiments in Fig.~\ref{fig:full-exp}, the ground-truth layout is simulated to generate a full-set of complex S-parameter targets (at 100 uniformly spaced frequencies from 1 to 100 GHz) that are then used in the diffusion only and diffusion with Kalman guidance synthesis. Notably, in several practical design scenarios, only S-parameter magnitudes are relevant. Therefore, in the experiments in Fig.~\ref{fig:mag-exp}, only the magnitudes of S-parameters of the ground-truth layouts are provided as inputs to the diffusion models.

The results in Figs.~\ref{fig:full-exp} and~\ref{fig:mag-exp} show that the conditional diffusion model alone can generate visually plausible layouts, some of which closely resemble the ground-truth layouts, while others differ substantially from them. 
Despite their visual plausibility, however, the diffusion-only designs can exhibit substantial S-parameter errors, often in cases involving locally irregular or disconnected metal regions (e.g., the middle panels of the first three rows of Fig.~\ref{fig:full-exp} and the first two rows of Fig.~\ref{fig:mag-exp}). The K-TRAIL correction substantially reduces this mismatch, producing layouts whose simulated S-parameter responses more closely match the prescribed targets. For each generated design, the corresponding S-parameter plots report the relative error, defined as the Euclidean norm of the difference between the simulated and reference S-parameter responses, normalized by the norm of the reference response.

The nonuniform meandered-line examples in the first rows of Figs.~\ref{fig:full-exp} and~\ref{fig:mag-exp} illustrate this behavior particularly clearly. In these cases, the conditional-diffusion-only approach produces disconnected layouts with S-parameter relative errors on the order of 100\% or higher, whereas K-TRAIL avoids such structurally inconsistent designs and produces layouts with substantially smaller errors. K-TRAIL also exhibits an interesting ability to identify alternative designs that satisfy the prescribed electromagnetic response rather than simply reproducing the reference geometry. For example, in the third row of Fig.~\ref{fig:full-exp}, although the reference layout is a uniform meandered line, K-TRAIL generates a qualitatively different spiral layout whose S-parameters closely follow those of the reference design. This result highlights the inherently nonunique nature of the inverse-design problem and the ability of K-TRAIL to identify alternative physically meaningful solutions. While these results demonstrate the substantial benefits of K-TRAIL for passive-block synthesis, in the next section, we extend the approach to amplifier-level design, where multiple passive blocks are co-generated to satisfy higher-level circuit constraints.

\section{Inverse Design Based on RF Constraints Using K-TRAIL}
\label{sec:sparameter_conditions}

\begin{figure}[t]
\begin{overpic}[
            width=.485\textwidth,
            percent,
            tics=5
        ]{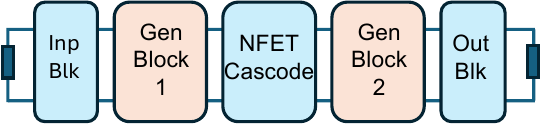}
        \end{overpic}\\
\begin{overpic}[
            width=.485\textwidth,
            percent,
            tics=5
        ]{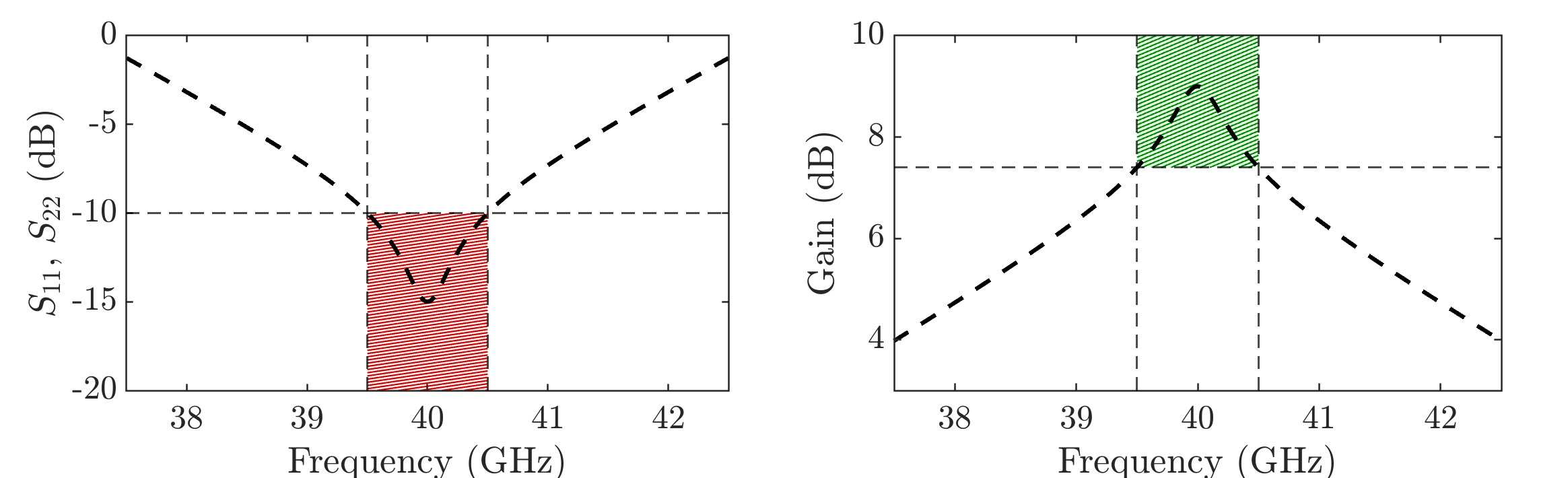}
        \end{overpic}%
  \caption{Block-level inverse design in which impedance transformation Blocks 1 and 2 are generated using the K-TRAIL approach to satisfy the corresponding block-level constraints. The constraints imposed on the cascaded system are defined in \eqref{eqn:constraint} and illustrated by the shaded regions in the bottom panel.}

    \label{fig:block_design}
\end{figure}

In many circuit-design problems, the desired behavior is not specified by a single S-parameter trace associated with a passive block. Instead, it is expressed through higher-level requirements such as passband insertion loss, stopband rejection, input matching, isolation, gain, or bandwidth. Fig.~\ref{fig:block_design} shows an example design problem where the input and output matching passive networks have been designed for a specific transistor, source, and load impedances in order to meet constraints such as gain and input/output match. These requirements define a set of acceptable responses rather than one target vector. To represent this set, we introduce a nonnegative violation map
$V_{\mathcal{C}}:\mathbb{C}^{m}\rightarrow
\mathbb{R}^{m_{\mathcal{C}}},$ where \(\mathcal{C}\) denotes the collection of RF specifications. Each component of \(V_{\mathcal{C}}\) is zero when its corresponding requirement is
satisfied and positive when it is violated. Composing this map with the EM simulator gives
\[
\mathcal{G}_{\mathcal{C}}(x)
=
V_{\mathcal{C}}\bigl(\mathcal{E}(x)\bigr).
\]
The constraint-driven synthesis problem can therefore be written in an inverse problem format as
\[
\min_x~\Psi_{\mathcal{C}}(x)
:=
\frac{1}{2}\left\|0-\mathcal{G}_{\mathcal{C}}(x)\right\|_2^2
=
\frac{1}{2}\left\|V_{\mathcal{C}}\bigl(\mathcal{E}(x)\bigr)\right\|_2^2 .
\]
Thus, the $y$-measurement or target in this formulation is the zero vector, corresponding to zero violation of all RF constraints. A feasible design is obtained when \(V_{\mathcal{C}}(\mathcal{E}(x))=0\), or when the violation norm is below a prescribed tolerance.

It is important to note that since in this case the specifications do not define a unique response trace, there is no
response vector \(y\) with which to condition the diffusion model. We instead
use an \emph{unconditional} diffusion prior \(p_\theta(x)\). In practical terms, this
model begins from noise and generates layouts resembling the feasible
structures in its training set, without being supplied target S-parameters.
It provides the structural prior, while the electrical requirements enter
only through simulator feedback. The prior may be trained directly from
layouts \(\{x_0^{(j)}\}_{j=1}^{n}\), or obtained from the unconditional branch
of a conditional diffusion model.

This formulation may be interpreted as sampling from
\[
p(x\mid\mathcal{C})
\propto
p_\theta(x)
\exp\left[
-\frac{1}{2\sigma_{\mathcal{C}}^2}
\left\|
V_{\mathcal{C}}\bigl(\mathcal{E}(x)\bigr)
\right\|_2^2
\right],
\]
where \(p_\theta(x)\) favors plausible layouts and the exponential term
penalizes specification violations. Equivalently, the acceptable response
set is $\mathcal{Y}_{\mathcal{C}}
=
\left\{
y:\;V_{\mathcal{C}}(y)=0
\right\}.$
No particular member of \(\mathcal{Y}_{\mathcal{C}}\) is selected as a
conditioning input; the simulator guides the generated layouts toward the
set as a whole.

\begin{figure}[t]
    \centering
    \setlength{\fboxsep}{0pt}%
    \colorbox{white}{%
        \begin{overpic}[
            width=.485\textwidth,
            percent,
            tics=5
        ]{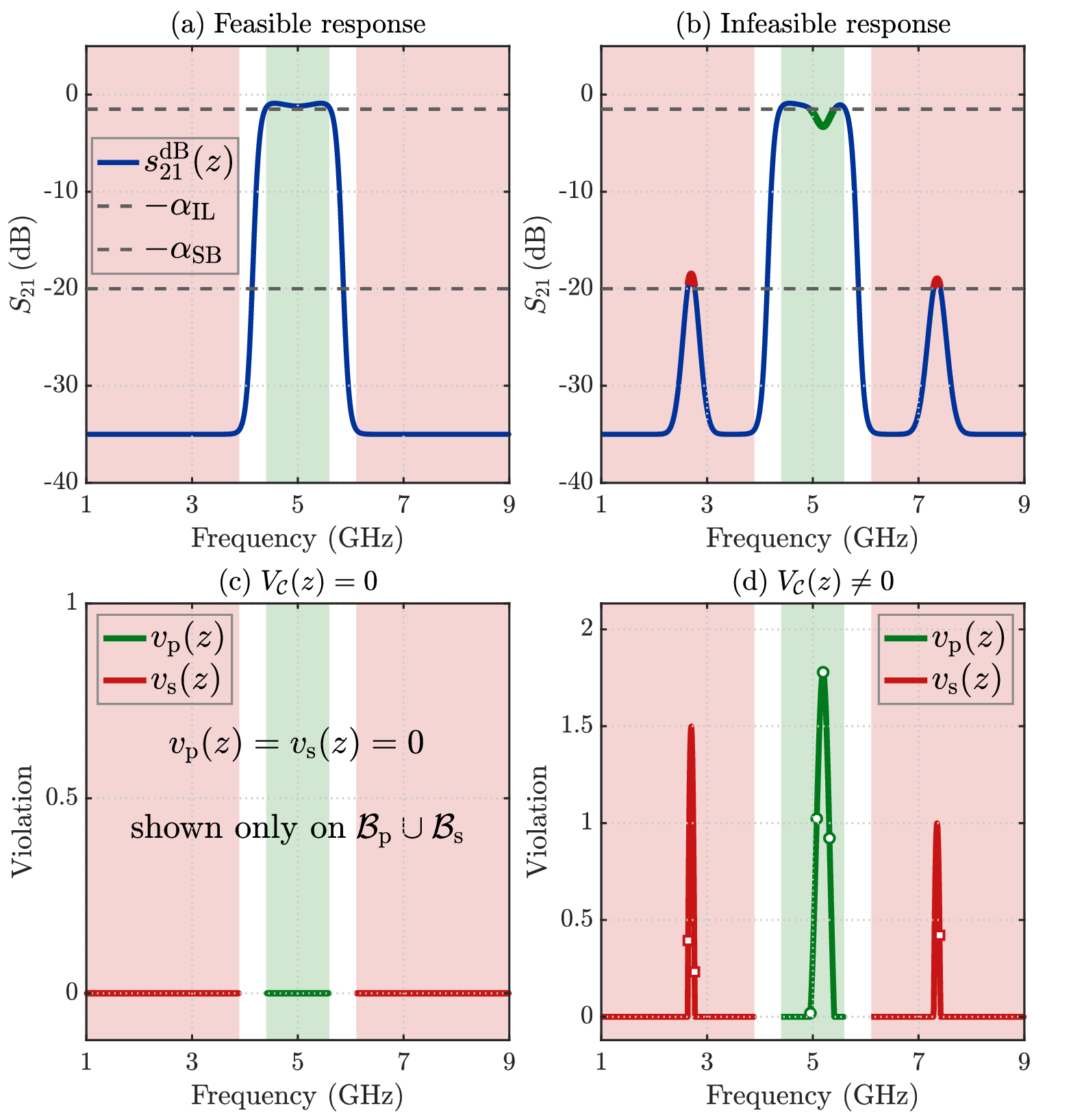}
        \end{overpic}%
    }
    \caption{Frequency-domain constraint violations for feasible and
    infeasible responses. A feasible response produces a zero violation
    vector, whereas a passband dip or stopband leakage produces nonzero
    components over the affected frequency range.}
    \label{fig:violation_plot}
\end{figure}

As an example, visualized in Figure \ref{fig:violation_plot}, consider a two-port filter with passband index set
\(\mathcal{B}_{\mathrm{p}}\) and stopband index set
\(\mathcal{B}_{\mathrm{s}}\). For a simulated response
\(z=\mathcal{E}(x)\), define
$s_{21,k}^{\mathrm{dB}}(z)
=
20\log_{10}\left|S_{21}(z;f_k)\right|.$
Suppose the design requires
\[
s_{21,k}^{\mathrm{dB}}(z)
\geq -\alpha_{\mathrm{IL}},
\qquad k\in\mathcal{B}_{\mathrm{p}},
\]
and
\[
s_{21,k}^{\mathrm{dB}}(z)
\leq -\alpha_{\mathrm{SB}},
\qquad k\in\mathcal{B}_{\mathrm{s}}.
\]
The corresponding violation components are
\[
v_{\mathrm{p},k}(z)
=
\left[
-\alpha_{\mathrm{IL}}
-s_{21,k}^{\mathrm{dB}}(z)
\right]_{+},
\qquad k\in\mathcal{B}_{\mathrm{p}},
\]
and
\[
v_{\mathrm{s},k}(z)
=
\left[
s_{21,k}^{\mathrm{dB}}(z)
+\alpha_{\mathrm{SB}}
\right]_{+},
\qquad k\in\mathcal{B}_{\mathrm{s}},
\]
where \([a]_{+}:=\max\{a,0\}\). Hence, the complete residual is
\[
V_{\mathcal{C}}(z)
=
\begin{bmatrix}
v_{\mathrm{p}}(z)\\
v_{\mathrm{s}}(z)
\end{bmatrix}.
\]
Consequently, \(V_{\mathcal{C}}(z)=0\) exactly when both the passband and
stopband requirements are satisfied, and $V_{\mathcal{C}}(z)\neq 0$ otherwise. Return-loss, gain, isolation,
bandwidth, and other circuit requirements can be incorporated by appending
their corresponding violation components.

The simulator-guided sampling procedure follows
Section~\ref{sec:exact_sparameter_synthesis}, with two modifications. First,
the conditional denoiser \(D_\theta(x_t,t,y)\) is replaced by the
unconditional denoiser \(D_\theta(x_t,t)\). Second, the response mismatch
\(y-z^{(i)}\) is replaced by a residual whose desired value is zero. For each
of the \(N\) trajectories,
\[
\hat{x}_0^{(i)}(t)
=
D_\theta(x_t^{(i)},t),
\qquad
z^{(i)}(t)
=
\mathcal{E}\!\left(\hat{x}_0^{(i)}(t)\right),
\]
and
\[
r^{(i)}(t)
=
V_{\mathcal{C}}\!\left(z^{(i)}(t)\right).
\]
Thus, all circuit specifications enter through \(r^{(i)}(t)\), rather than
through an artificial target S-parameter trace.

At selected correction steps
\(t\in\mathcal{T}_{\mathrm{corr}}\), the ensemble pairs
\(\bigl(\hat{x}_0^{(i)}(t),r^{(i)}(t)\bigr)\) are used to form the
layout--residual covariance \(C_{xr}(t)\) and residual--residual covariance
\(C_{rr}(t)\), analogously to \(C_{xz}(t)\) and \(C_{zz}(t)\) in
Section~\ref{sec:exact_sparameter_synthesis}. The corresponding Kalman gain is
$K_t^{\mathcal{C}}
=
C_{xr}(t)
\left(C_{rr}(t)+\sigma_{\mathcal{C}}^2 I\right)^{-1},$
where \(\sigma_{\mathcal{C}}^2\) regularizes the update and represents the
assumed residual tolerance. The predicted layouts are then corrected
toward zero violation via
\[
\tilde{x}_0^{(i)}(t)
=
\hat{x}_0^{(i)}(t)
-
\rho_t K_t^{\mathcal{C}}r^{(i)}(t),
\]
where \(\rho_t\) controls the correction strength. The corrected estimate
replaces the original clean-layout prediction in the next denoising step:
\[
x_{t-1}^{(i)}
=
\mathrm{SamplerStep}
\left(
x_t^{(i)},\tilde{x}_0^{(i)}(t),t
\right).
\]
Across the
ensemble, \(C_{xr}(t)\) provides a local estimate of which layout variations
are associated with changes in the circuit-level violations.

As in the exact-response case, the update is derivative-free and treats the
EM simulator as a black-box evaluator. It does not require differentiation
through the simulator, the S-parameter post-processing operations, or the
violation map. Because the unconditional prior spans a broader set of
layouts than a response-conditioned model, constraint-driven synthesis may
require a larger ensemble or more correction steps. After sampling, the
final candidates are simulated once more, and any design satisfying
\[
V_{\mathcal{C}}\bigl(\mathcal{E}(x_0^{(i)})\bigr)=0
\]
may be selected; if no candidate is exactly feasible, the design with the
smallest violation norm is retained.

\section{Block-level Inverse Design Using K-TRAIL Approach with Violation Maps}\label{sec:Block-level Inverse Design}
In the following, we consider the design problem shown in Fig.~\ref{fig:block_design}, where the transistor block is based on a simulated cascode transistor in 65\,nm CMOS design technology. The design targets are amplifier gain and input-output match at 40\,GHz. Notably, the amplifier-gain objective is set close to the network's maximum available gain, providing a challenging goal for the inverse design approach. The input and output matching-network layout uses the same templates trained in Fig.~\ref{fig:templates}. 

The experiments follow this approach to create a well-defined comparison point for evaluating performance: in each experiment, $k$, we generate a pair of reference blocks, $REF_{k,1}$ and $REF_{k,2}$. Next, we generate port interface blocks, $Inp_k$ and $Out_k$, to provide a conjugate input and output match for the cascade of $REF_{k,1}$, the transistor block, and $REF_{k,2}$. In the subsequent inverse-design approach, $Inp_k$ and $Out_k$ \emph{remain fixed}, and we evaluate and compare the generated layouts and performance to that achieved with the reference blocks. 

The circuit-level design requirements are set based on the maximum available gain from the transistor:
\begin{equation}\label{eqn:constraint}
S_{11}^{\mathrm{dB}}(f)\leq -10~\mathrm{dB}, ~~ S_{22}^{\mathrm{dB}}(f)\leq -10~\mathrm{dB},
~~
G^{\mathrm{dB}}(f)\geq 7.4~\mathrm{dB},
\end{equation}
for $39.5\leq f\leq40.5~\mathrm{GHz}$. 
As described in Section~\ref{sec:sparameter_conditions}, the sampled \(S_{11}\), \(S_{22}\) and \(G_{\max}\) responses are converted into nonnegative, frequency-dependent violation components and used as the residual in the K-TRAIL correction.

\begin{figure*}[!htbp]
    \centering

    \includegraphics[
        width=.495\linewidth
    ]{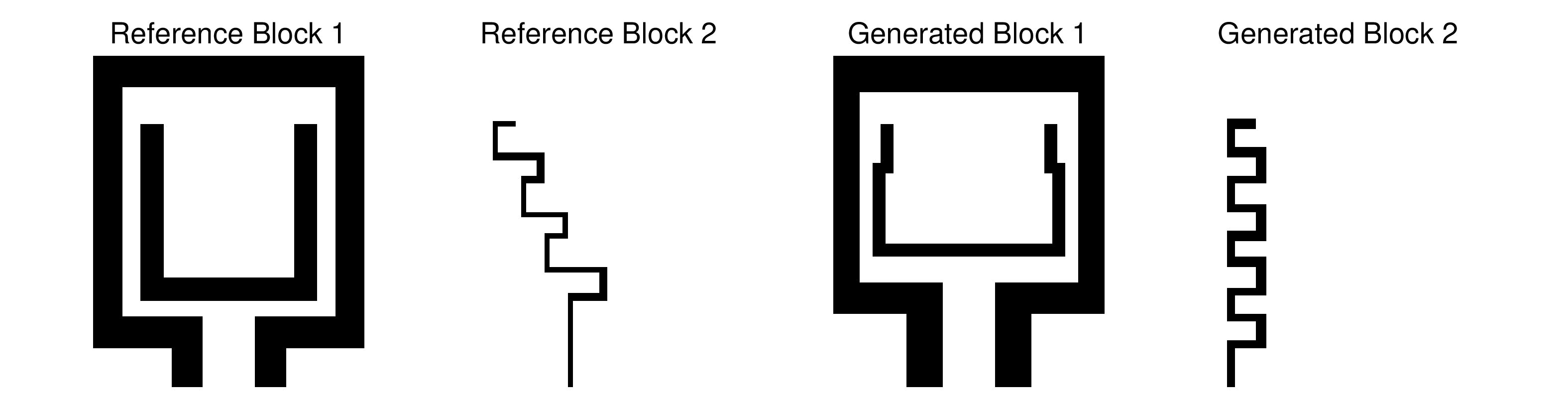}%
    \hfill%
    \includegraphics[
        width=.495\linewidth
    ]{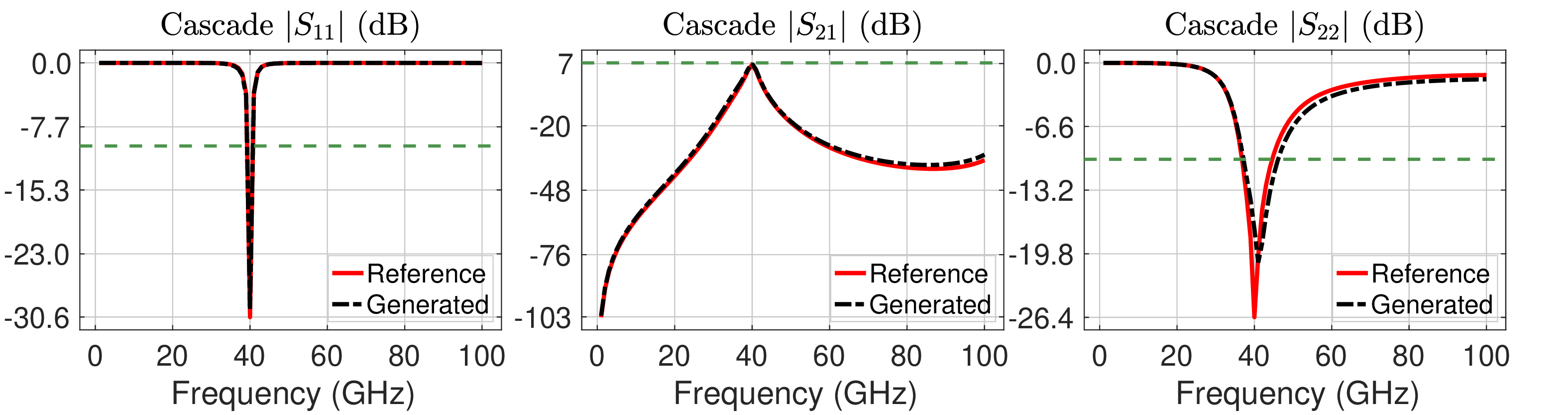}

    \vspace{0.1cm}

    \includegraphics[
        width=.495\linewidth
    ]{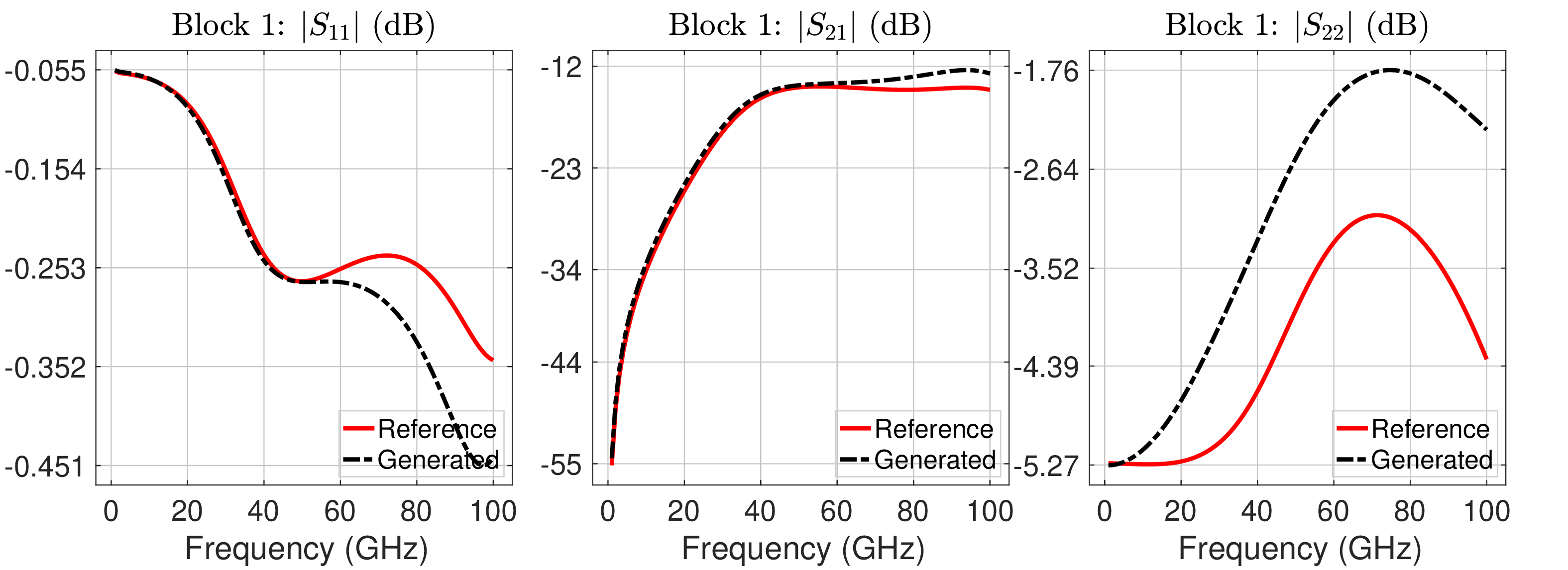}%
    \hfill%
    \begin{overpic}[
        width=.495\linewidth
    ]{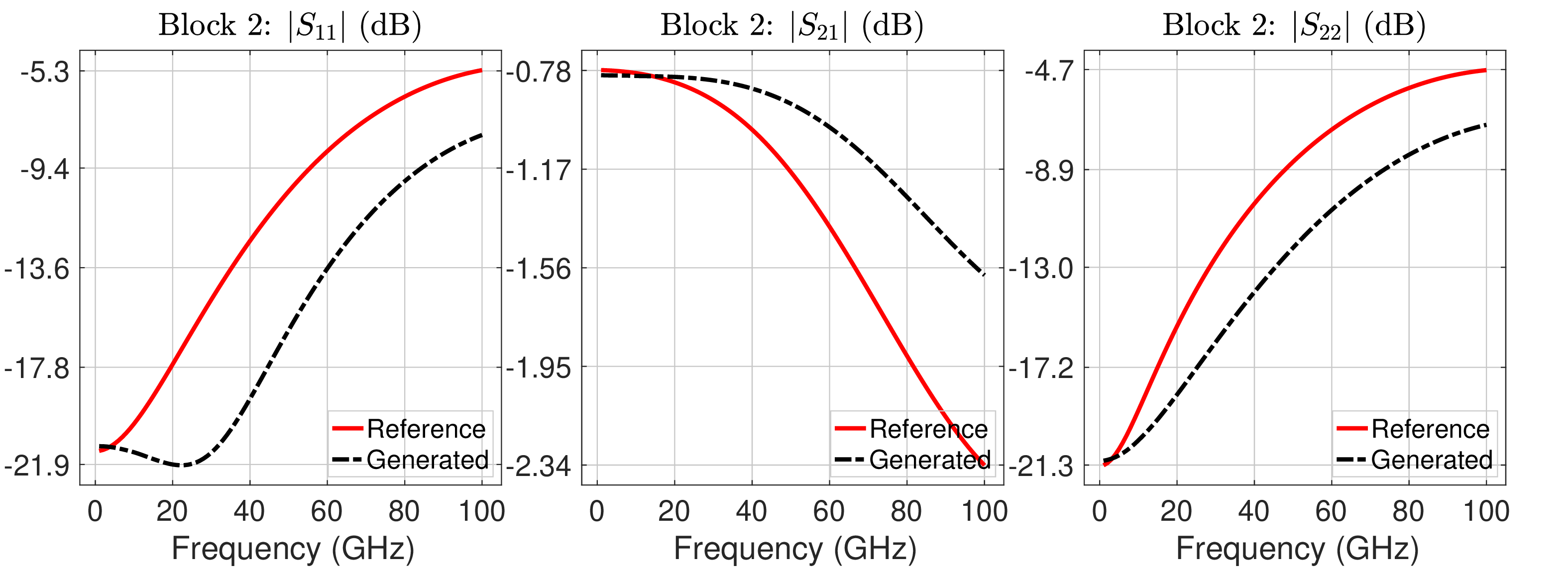}
        \put(0,-4){\makebox(0,0){\textnormal{(a)}}}
    \end{overpic}

    \vspace{0.6cm}

    \includegraphics[
        width=.495\linewidth
    ]{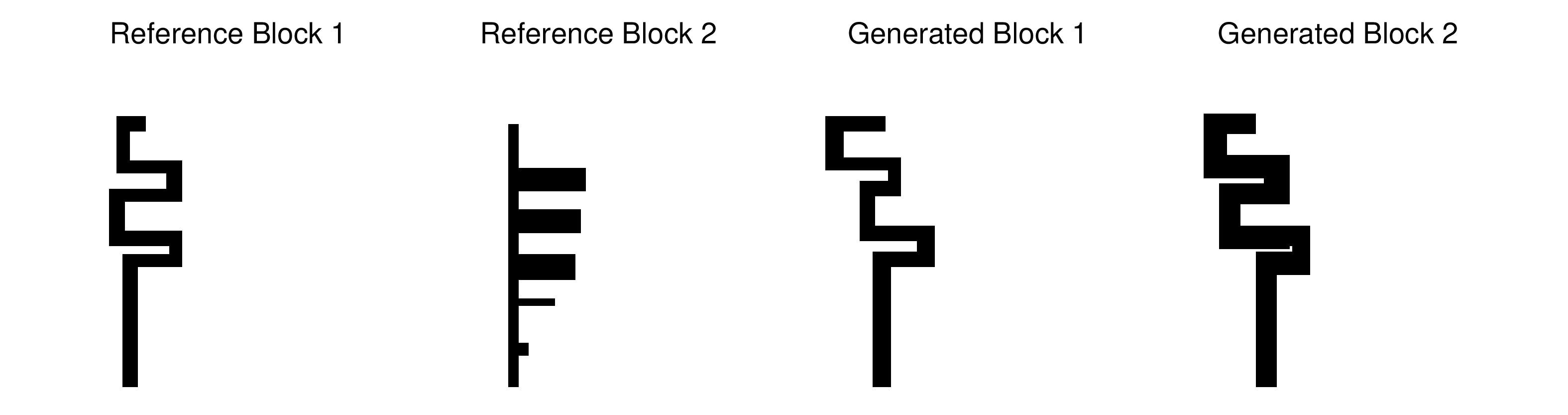}%
    \hfill%
    \includegraphics[
        width=.495\linewidth
    ]{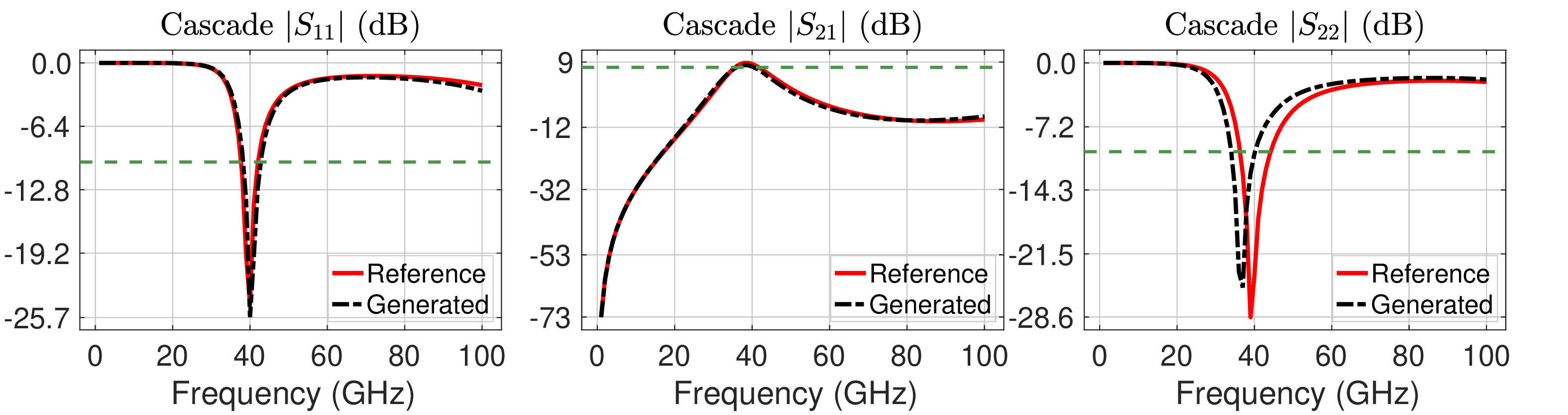}

    \vspace{0.1cm}

    \includegraphics[
        width=.495\linewidth
    ]{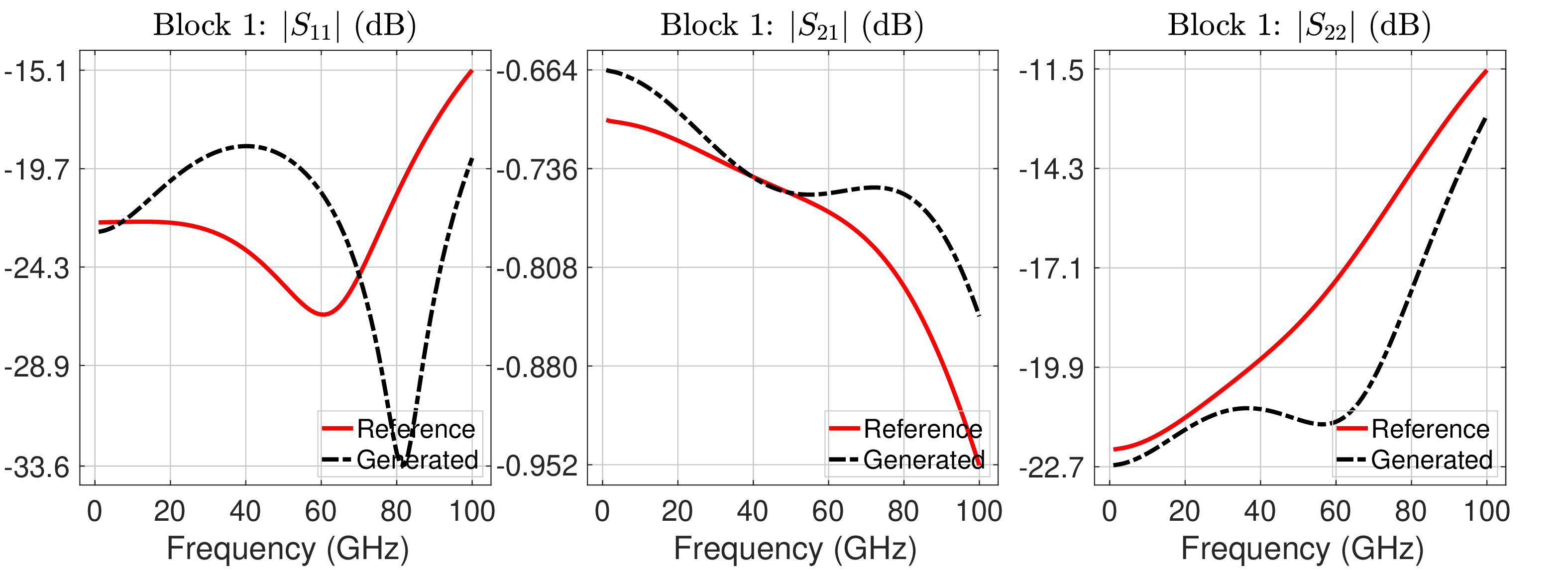}%
    \hfill%
    \begin{overpic}[
        width=.495\linewidth
    ]{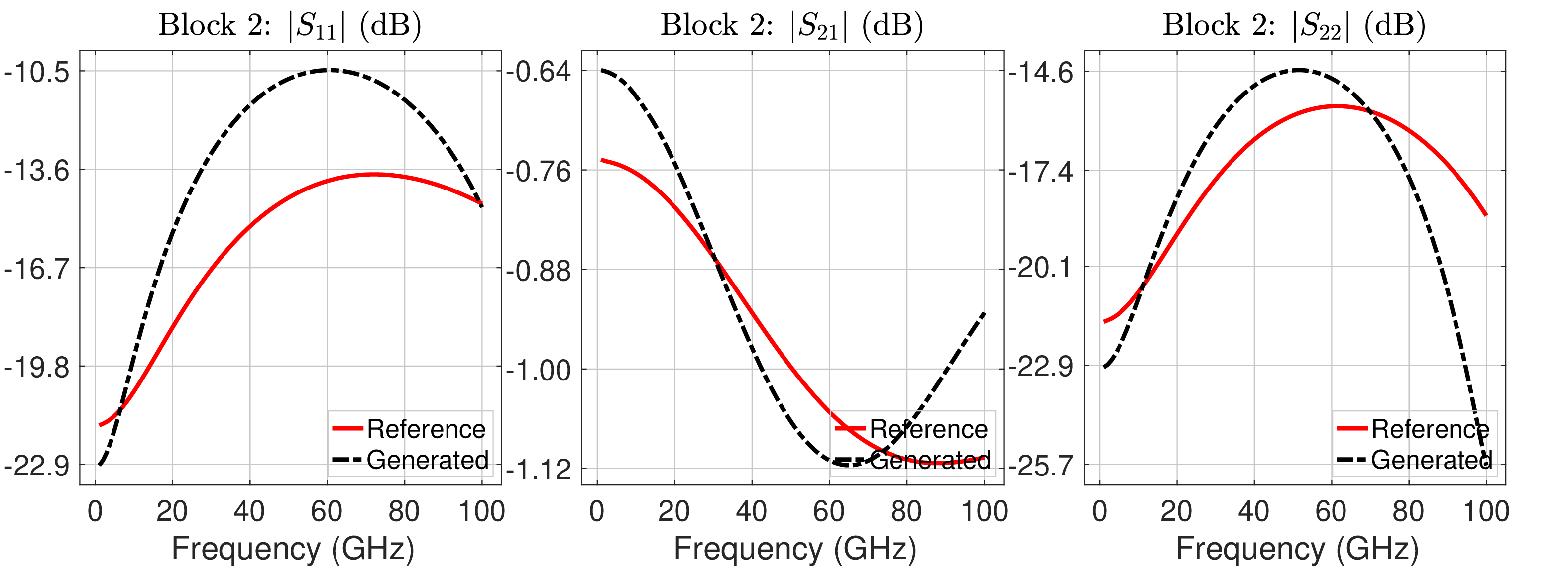}
        \put(0,-4){\makebox(0,0){\textnormal{(b)}}}
    \end{overpic}

    \vspace{0.6cm}

    \includegraphics[
        width=.495\linewidth
    ]{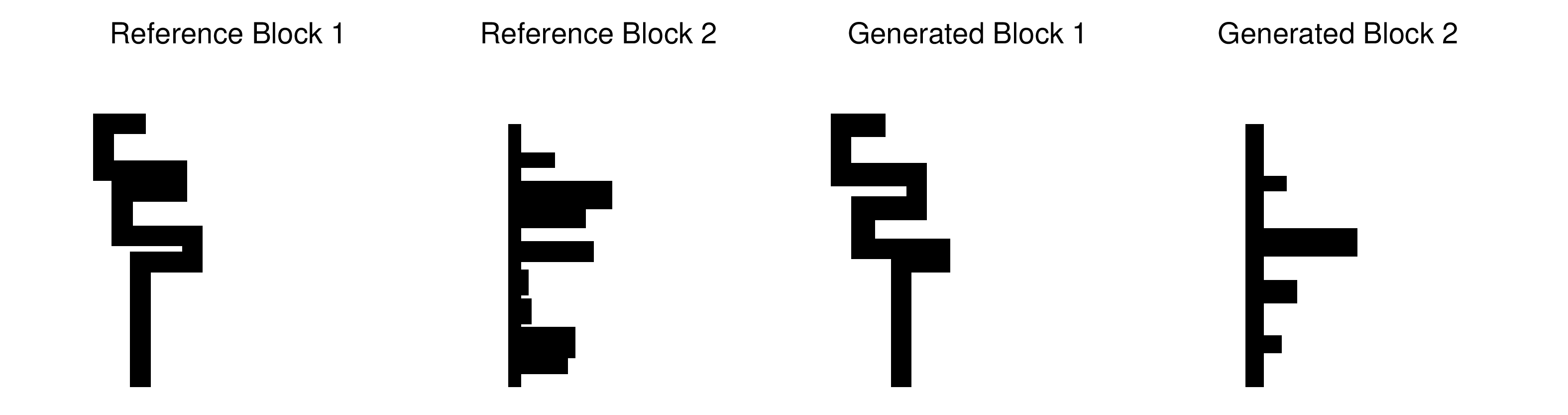}%
    \hfill%
    \includegraphics[
        width=.495\linewidth
    ]{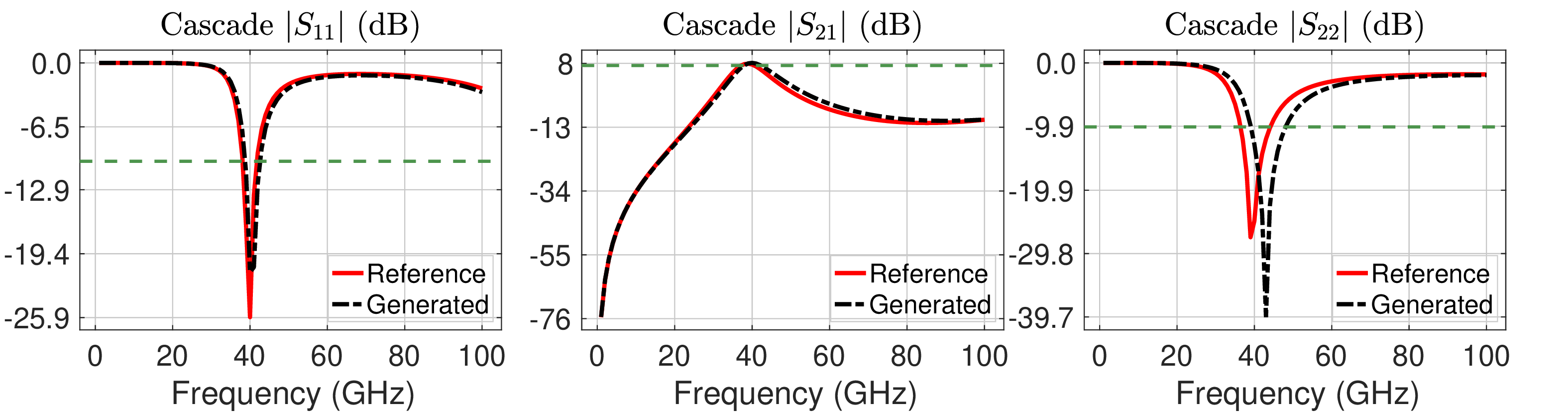}

    \vspace{0.1cm}

    \includegraphics[
        width=.495\linewidth
    ]{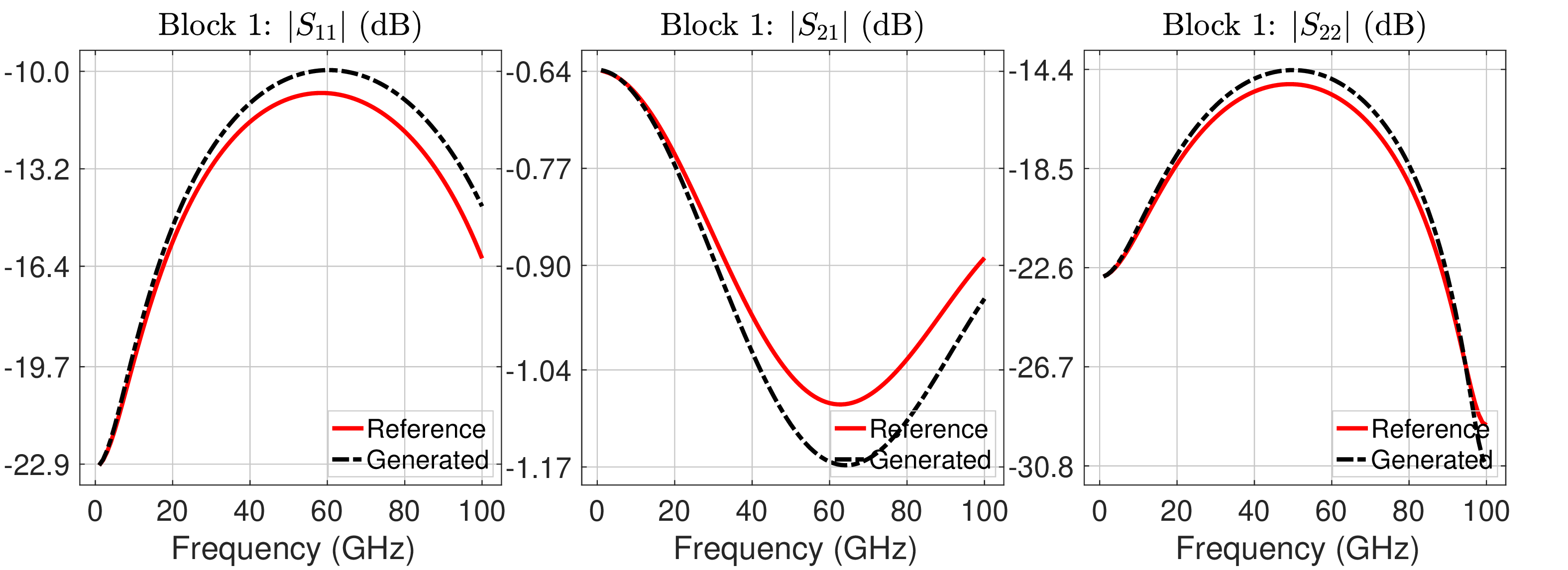}%
    \hfill%
    \begin{overpic}[
        width=.495\linewidth
    ]{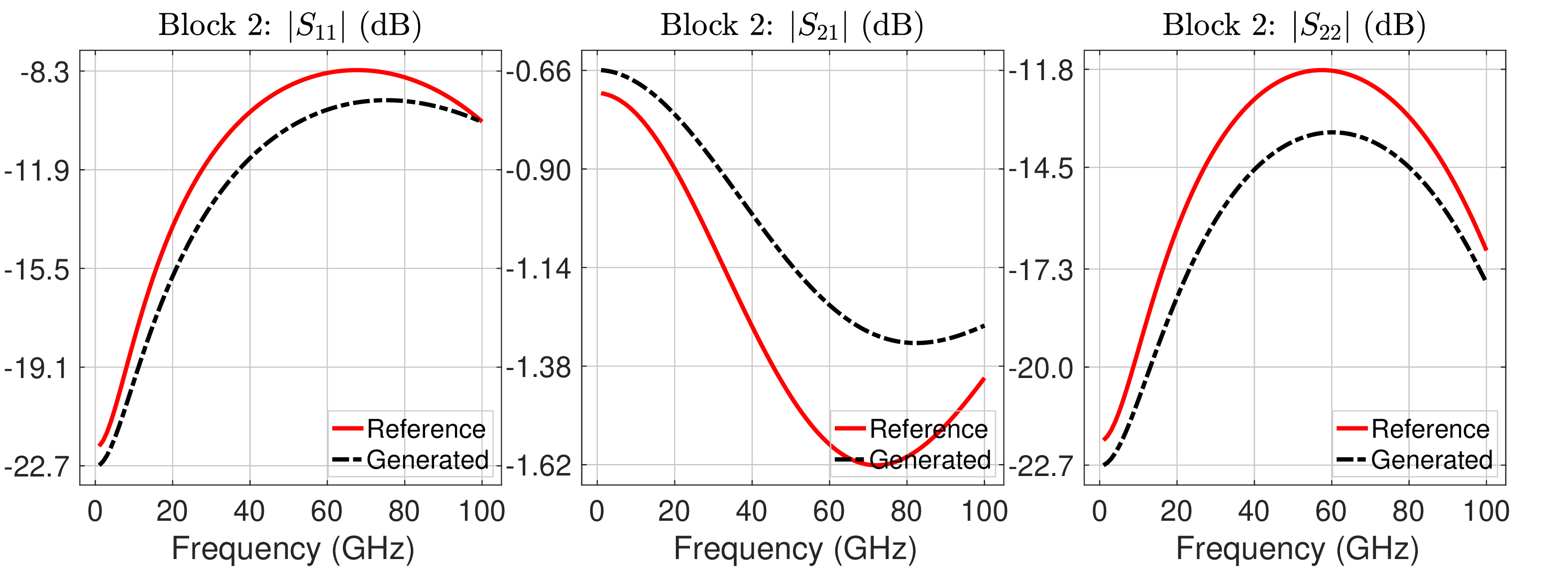}
        \put(0,-4){\makebox(0,0){\textnormal{(c)}}}
    \end{overpic}

    \vspace{0.3cm}

    \caption{Representative two-block design for three experiments.
    For each experiment, the first row shows the reference and generated
    block layouts together with the corresponding cascaded S-parameter
    responses. The constraint margins are shown with dashed green line. The second row compares the reference and generated
    S-parameter responses for Blocks~1 and~2 individually:
    (a) Experiment~1, (b) Experiment~2, and (c) Experiment~3.}

    \label{fig:two-block-experiments}

\end{figure*}

Two experimental configurations are considered. For the interleaved/meandered line-layout configuration, guidance begins at \(t=30\), the correction weight is \(0.75\), and corrections are applied every two reverse-diffusion steps. For the meandered line/stub tuner configurations, guidance begins at \(t=50\), the correction weight is \(0.4\), and corrections are again applied every two steps. All configurations use \(N=64\) trajectories. For each configuration, we perform one 64-trajectory K-TRAIL trial using a fixed random seed. All final particles are then evaluated using EMX and the complete cascade, and the reported candidate is selected as the particle that minimizes \(\|V_{\mathcal C}\|_2\).

After generation, we evaluate the final layouts using the complete circuit cascade. The simulated \(S_{11}\), \(G_{\max}\), and aggregate constraint-violation metric are computed over the \(35\)--\(45~\mathrm{GHz}\) band. The resulting layout and circuit-response comparisons are presented in Fig.~\ref{fig:two-block-experiments}. In each of these experiments, the reference block 1 and block 2 are used to generate the input and output blocks. Subsequently, the block-level inverse design approach is used to generate layouts that satisfy the constraints in (\ref{eqn:constraint}). As can be seen in Fig.~\ref{fig:two-block-experiments}, the generated blocks, while different in template or geometry from the reference blocks, achieve targeted performance and avoid unexplainable, irregular or non-manufacturable structures, demonstrating the capabilities of the proposed approach. 

\section{Conclusion and Future Work}

We present K-TRAIL, a simulator-guided generative framework for inverse design of EM/RF circuits that combines the expressive design-space exploration of diffusion models with derivative-free ensemble Kalman guidance from a black-box EM simulator. By incorporating physical verification directly into the generative trajectory, K-TRAIL substantially improves agreement with prescribed S-parameter responses relative to unguided generation while requiring neither adjoint sensitivities nor a differentiable surrogate model. The framework was demonstrated for synthesis from both complex and magnitude-only S-parameter targets and was further extended to specification-driven design through violation maps that directly encode circuit-level RF constraints. Importantly, K-TRAIL is not restricted to reproducing a reference geometry, since the generated designs can differ substantially in topology while realizing the desired electrical behavior, as shown in experiments at the individual-block and system levels. These results suggest a path toward specification-to-layout RF design in which learned generative priors provide physically meaningful candidate structures and high-fidelity simulation supplies the final measure of correctness. Future work will expand this approach to RF design, including small-signal and large-signal constraints that depend on co-design of active and passive circuits. Ultimately, integration with complete RFIC design flows and experimental validation of fabricated generated structures will be important steps toward practical, verification-aware generative RF design automation.

\appendix{EM and K-TRAIL Experimental Setup}
\label{sec:experimentalDetails}

In the following, we describe the layout dataset, EM simulation flow, diffusion-model configuration, and K-TRAIL parameters used in the experiments. 

\subsection{Layout Dataset and Representation}

\begin{figure}[t]
 \hspace{-.2cm}
\begin{overpic}[
    width=.48\textwidth,
    percent,
    tics=5
]{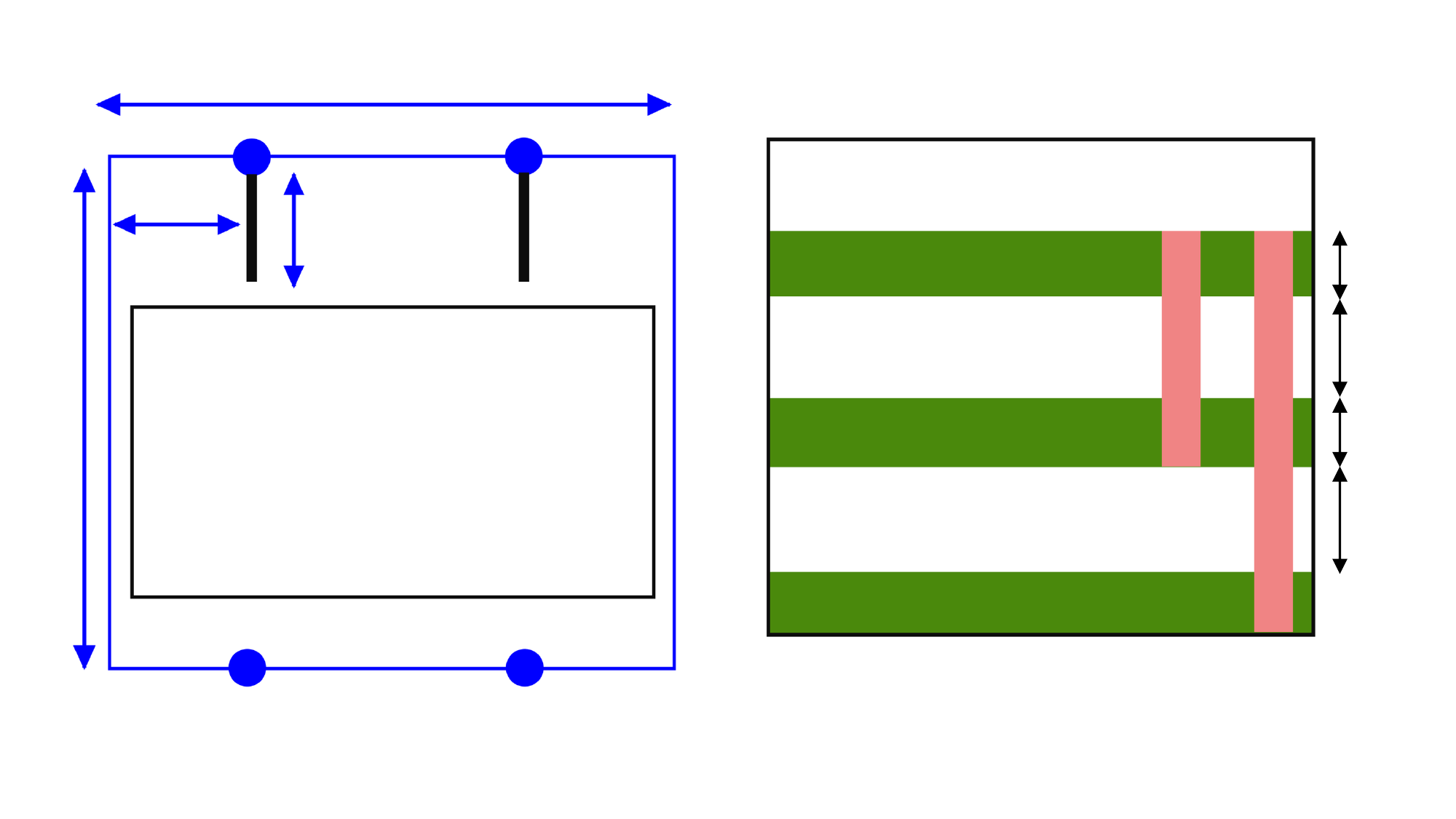}

\put(2,27){\makebox(0,0)[c]{\scalebox{.9}{\rotatebox{90}{$500~\mu m$}}}}
\put(27,25){\makebox(0,0)[c]{\scalebox{.9}{\rotatebox{0}{Generated Layout}}}}
\put(25,5){\makebox(0,0)[c]{\scalebox{.9}{\rotatebox{0}{(a)}}}}
\put(72,5){\makebox(0,0)[c]{\scalebox{.9}{\rotatebox{0}{(b)}}}}
\put(60,15){\makebox(0,0)[c]{\scalebox{.9}{\rotatebox{0}{GND}}}}
\put(60,27){\makebox(0,0)[c]{\scalebox{.9}{\rotatebox{0}{M1}}}}
\put(60,38.3){\makebox(0,0)[c]{\scalebox{.9}{\rotatebox{0}{M2}}}}
\put(70,20){\makebox(0,0)[c]{\scalebox{.9}{\rotatebox{0}{$\varepsilon_r=4$}}}}
\put(70,32.5){\makebox(0,0)[c]{\scalebox{.9}{\rotatebox{0}{$\varepsilon_r=4$}}}}
\put(97,38){\makebox(0,0)[c]{\scalebox{.8}{\rotatebox{0}{$4~\mu m$}}}}
\put(97,32){\makebox(0,0)[c]{\scalebox{.8}{\rotatebox{0}{$4~\mu m$}}}}
\put(97,26.7){\makebox(0,0)[c]{\scalebox{.8}{\rotatebox{0}{$2~\mu m$}}}}
\put(97,20){\makebox(0,0)[c]{\scalebox{.8}{\rotatebox{0}{$3~\mu m$}}}}
\put(23,41){\makebox(0,0)[c]{\scalebox{.7}{\rotatebox{90}{$110~\mu m$}}}}
\put(12,43){\makebox(0,0)[c]{\scalebox{.7}{\rotatebox{0}{$120~\mu m$}}}}
\put(25,52){\makebox(0,0)[c]{\scalebox{.9}{\rotatebox{0}{$500~\mu m$}}}}
\put(70,44){\makebox(0,0)[c]{\scalebox{.9}{\rotatebox{0}{Layers}}}}
\put(17,13){\makebox(0,0)[c]{\scalebox{.7}{\rotatebox{0}{P1}}}}
\put(36,13){\makebox(0,0)[c]{\scalebox{.7}{\rotatebox{0}{P3}}}}
\put(17,47.8){\makebox(0,0)[c]{\scalebox{.7}{\rotatebox{0}{P2}}}}
\put(36,47.8){\makebox(0,0)[c]{\scalebox{.7}{\rotatebox{0}{P4}}}}
\end{overpic}\vspace{-.5cm}
\caption{Fixed port and lower-layer geometry included in each generated GDS layout. }
\label{fig:static-layout/stackup}
\end{figure}

The diffusion models are trained using \(128\times128\)-pixel images derived from three-layer RFIC layouts. The images supplied to the model contain only the variable M2 geometry (see Fig. \ref{fig:static-layout/stackup}). The fixed ground, M1, port, and routing structures are restored when each generated image is converted to GDS for EM simulation. Before training, each image is mapped to a \(4\times32\times32\)-dimensional latent representation using a variational autoencoder (VAE) \cite{kingma2013auto} trained on the circuit layouts.

The training library contains six RF layout topologies, as shown and listed in Fig.~\ref{fig:templates}. For each topology, bounded geometric degrees of freedom were randomly varied to generate approximately 15,000 distinct layouts. The resulting 89,810-sample dataset was divided into 85,319 training, 2,245 validation, and 2,246 test layouts, corresponding approximately to a \(95\%/2.5\%/2.5\%\) split. The degrees of freedom varied for each topology are summarized in Table~\ref{tab:dof}.

\begin{table}[t]
\centering
\caption{Degrees of freedom used to generate the training layouts.}
\label{tab:dof}

\footnotesize
\setlength{\tabcolsep}{1.5pt}
\renewcommand{\arraystretch}{1.15}

\begin{tabularx}{\columnwidth}{|
>{\centering\arraybackslash}p{1.65cm}|
*{6}{>{\centering\arraybackslash}X|}}
\hline

DoF &
\rotatebox[origin=c]{90}{\makecell{Uniform\\Meander}} &
\rotatebox[origin=c]{90}{\makecell{Nonuniform\\Meander}} &
\rotatebox[origin=c]{90}{\makecell{Stub\\Tuner}} &
\rotatebox[origin=c]{90}{\makecell{Interleaved\\Transformer}} &
\rotatebox[origin=c]{90}{\makecell{Spiral\\Inductor}} &
\rotatebox[origin=c]{90}{\makecell{Coupled\\Line}} \\
\hline

Line Width 
& 1 & 1 & Multiple & 1 & 1 & 1 \\
\hline
Vertical Meander Spacing 
& 1 & Multiple & N/A & N/A & N/A & N/A \\
\hline
Horizontal Meander 
& 1 & Multiple & N/A & N/A & N/A & N/A \\
\hline
Number of Segments/Stubs 
& 1 & 1 & 1 & N/A & N/A & N/A \\
\hline
Length 
& N/A & N/A & Multiple & 2 & 1 & Multiple \\
\hline
Trace--Trace Spacing 
& N/A & N/A & Multiple & 1 & 1 & Multiple \\
\hline
Spacing Between Feeds 
& N/A & N/A & N/A & 1 & N/A & N/A \\
\hline

\end{tabularx}
\end{table}

A common \(500\times500~\mu\mathrm{m}^2\) design region and fixed port configuration are used for all topologies, as shown in Fig.~\ref{fig:static-layout/stackup}. Four simulation ports are placed at the corners of the design region. Ports~2 and~4 are located on M1 and accessed from M2 through vias positioned \(110~\mu\mathrm{m}\) from the upper boundary and \(120~\mu\mathrm{m}\) from the corresponding side boundaries. Each via is connected to its port through a \(7~\mu\mathrm{m}\)-wide M1 trace. Keeping the stackup, ports, and lower-layer routing fixed enables automatic port assignment and batch simulation of the generated layouts.

Topology-specific connectivity rules are imposed during data generation. The uniform and nonuniform meandered lines and the stub tuner connect Ports~1 and~2. The coupled-line topology maintains a DC connection between Port~1 and the via associated with Port~2. When coupled segments are present between Ports~3 and~4, the corresponding DC connection is also enforced. The spiral inductor connects Ports~1 and~2 and terminates on the right-hand side to permit automatic routing without intersecting the fixed geometry. For the interleaved transformer, the inner transmission line is constrained to intersect the vias associated with Ports~2 and~4.

The layouts are generated using a topology-independent Python-based pipeline built on \texttt{gdspy}. The framework constructs each topology from parameterized transmission-line elements while automatically maintaining geometric connectivity, eliminating the need for topology-specific coordinate definitions. Additional scripts provide bidirectional conversion between the GDS layouts and the \(128\times128\)-pixel image representation used by the diffusion model.

\subsection{EM Simulation Setup}

The S-parameter labels used for training are generated using Cadence EMX V6.1 over 1 to 100 GHz with a step size of 1GHz for a total of 100 frequency points. The three-layer stackup is shown in Fig.~\ref{fig:static-layout/stackup}. M1 is \(2~\mu\mathrm{m}\) thick and separated from the ground plane by a \(3~\mu\mathrm{m}\) dielectric layer. M2 is \(4~\mu\mathrm{m}\) thick and separated from M1 by a \(4~\mu\mathrm{m}\) dielectric layer. Both dielectric layers have relative permittivity \(\epsilon_r=4\). The conductor was taken to have conductivity of 5.96$\times 10^7$ $\mathrm{S}/\mathrm{m}$. The conductivity of each via is taken to be 2.5$\times 10^5$ $\mathrm{S}/\mathrm{m}$ with a cross sectional area of $1\mu\mathrm{m}^2$. The simulator used an edge width of $5\mu\mathrm{m}$ to reduce simulation time.

Generated layouts are converted to GDS and evaluated using the same stackup, port configuration, and simulation settings used to generate the training labels. The outcomes of inverse synthesis are evaluated using the same thresholding, geometry projection, and meshing operations in this automated evaluation pipeline.

\subsection{Diffusion-Model and K-TRAIL Configuration}

The conditional and unconditional diffusion models use a latent diffusion transformer operating on the \(4\times32\times32\) VAE representation. The denoising backbone uses a patch size of 2, a hidden dimension of 512, and a four-scale U-shaped transformer architecture with timestep-conditioned adaptive normalization. In the conditional model, the S-parameter vector is linearly projected to a \(4\times32\times32\) tensor and concatenated with the layout latent, whereas the unconditional model receives only the layout latent. Both models are trained using Adam with the default momentum coefficients, a constant learning rate of \(10^{-5}\), a global batch size of 16, and 301 epochs, corresponding to approximately \(1.6\times10^6\) optimization steps. Training uses a 1,000-step linear forward-noise schedule and an \(x_0\)-prediction mean-squared-error objective. The models share the same VAE latent representation, transformer backbone, diffusion objective, and training schedule. However, at the input the conditional model includes an additional response projection and operates on eight channels after concatenation, whereas the unconditional model operates directly on the four-channel layout latent.

All experiments use \(T=100\) reverse-diffusion steps. The ensemble size was selected by evaluating \(N\in\{16,32,64,96,128\}\) using the S-parameter response-space error for exact-response synthesis and the violation norm \(\|V_{\mathcal C}\|_2\) for constraint-driven synthesis. Performance gains diminished beyond \(N=64\), therefore, \(N=64\) is used in all reported experiments as a compromise between correction quality and simulation cost. At each selected K-TRAIL correction step, the predicted clean layouts are converted to simulation-ready geometries and evaluated by the EM simulator. The resulting ensemble covariances are then used to correct the parallel diffusion trajectories.

\subsection{Exact S-Parameter Target Experiments}
\label{subsec:exact_response_experiments}

The exact-response experiments use target S-parameter responses obtained from 2,246 held-out layouts excluded from diffusion-model training. Two conditioning representations are considered. In the \emph{full-response} experiments, the target contains both magnitude and phase information. In the \emph{magnitude-only} experiments, only the sampled S-parameter magnitudes are supplied. Before being used by the conditional diffusion model and the K-TRAIL correction, the response vectors are normalized using the coordinate-wise mean and standard deviation
  computed from the training data, via
  \(\widetilde{y}_j={y_j-\mu_j}/{(\sigma_j+10^{-8}}).\)
  The K-TRAIL correction itself uses the corresponding unnormalized
  linear S-parameter representation when constructing the
  simulator-based residual.

For each target, layouts are generated using both unguided conditional diffusion and K-TRAIL-guided conditional diffusion.  The K-TRAIL correction weight is \(0.5\) (unless otherwise stated). Guidance begins at reverse-diffusion step \(t=50\), and the correction is applied every five steps. One unguided sample and one guided ensemble of 64 trajectories are generated for each target. Following final EM evaluation of the guided ensemble, the candidate with the smallest unnormalized S-parameter residual is selected for reporting.

After generation, each selected layout is converted to GDS and simulated to obtain \(S_{11}\), \(S_{21}\), and \(S_{22}\). Agreement with the prescribed response is quantified using an unnormalized Euclidean error over all four S-parameters and all 100
  frequency points. For full-response conditioning, the difference error is
\[ E_{\mathrm{RI}}=
  \left\|
  \begin{bmatrix}
  \operatorname{Re}\widehat{ S}\\
  \operatorname{Im}\widehat{ S}
  \end{bmatrix}
  -
  \begin{bmatrix}
  \operatorname{Re}S^\star\\
  \operatorname{Im} S^\star
  \end{bmatrix}
  \right\|_2 ,\]
  whereas for magnitude-only conditioning it is $E_{\mathrm{mag}}=
  \|
  |\widehat{ S}|-| S^\star|
  \|_2$. Here, the arrays are flattened over frequency and port indices.

\bibliographystyle{IEEEtran}

\vspace{-4em}
\begin{IEEEbiography}
[{\includegraphics[width=1in,height=1.25in,clip,keepaspectratio]{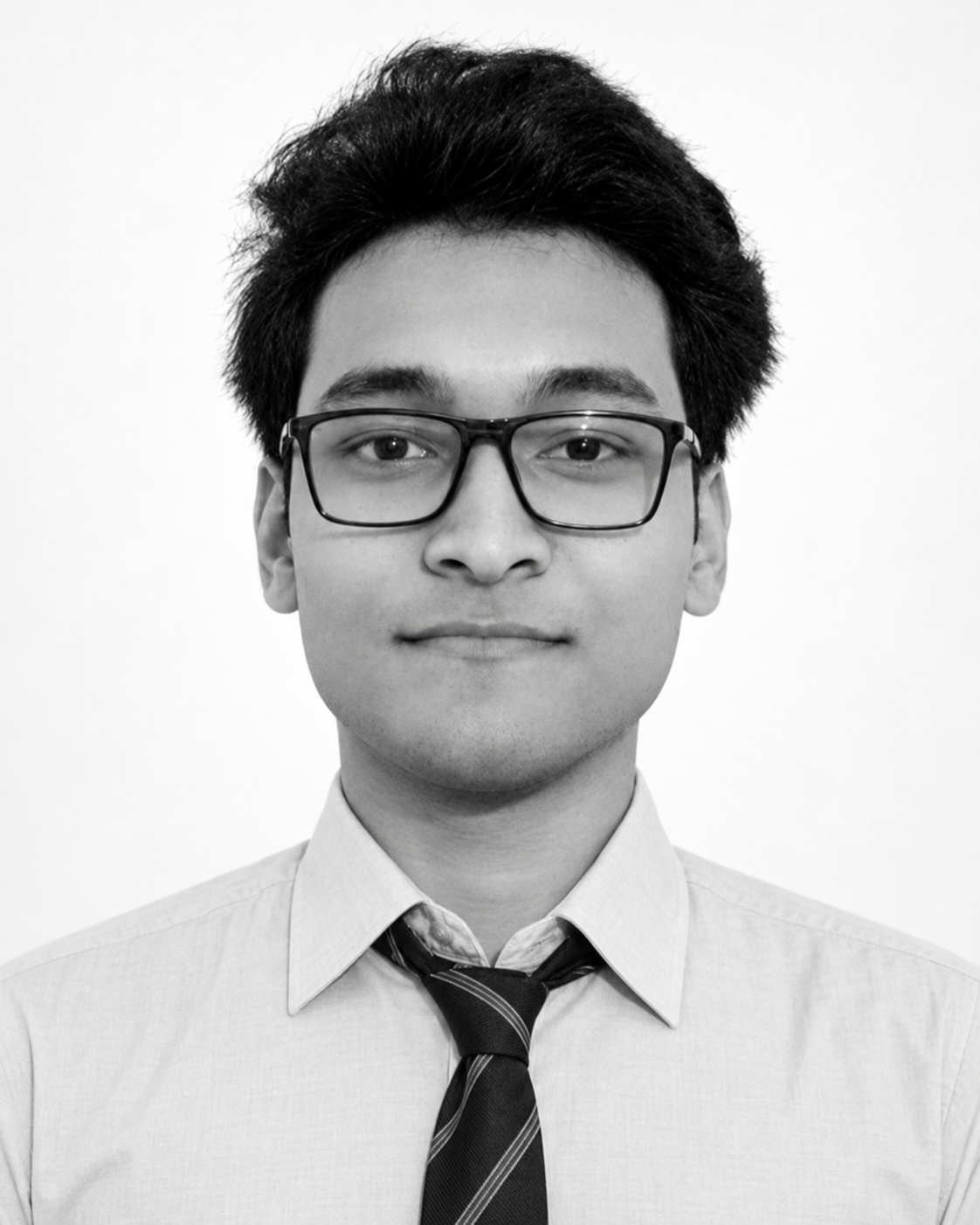}}]{Piyush Saha}
is currently a MS+PhD student with the School of Electrical Engineering and Computer Science. Previously, he held an AI Engineer position at IBM. He received his BTech in Electrical Engineering from Maulana Abul Kalam Azad University of Technology, India. His research interests include generative models, optimization theory, reinforcement learning, artificial intelligence, and signal processing.
\end{IEEEbiography}
\vspace{-2em}
\begin{IEEEbiography}[{\includegraphics[width=1in,height=1.25in,clip,keepaspectratio]{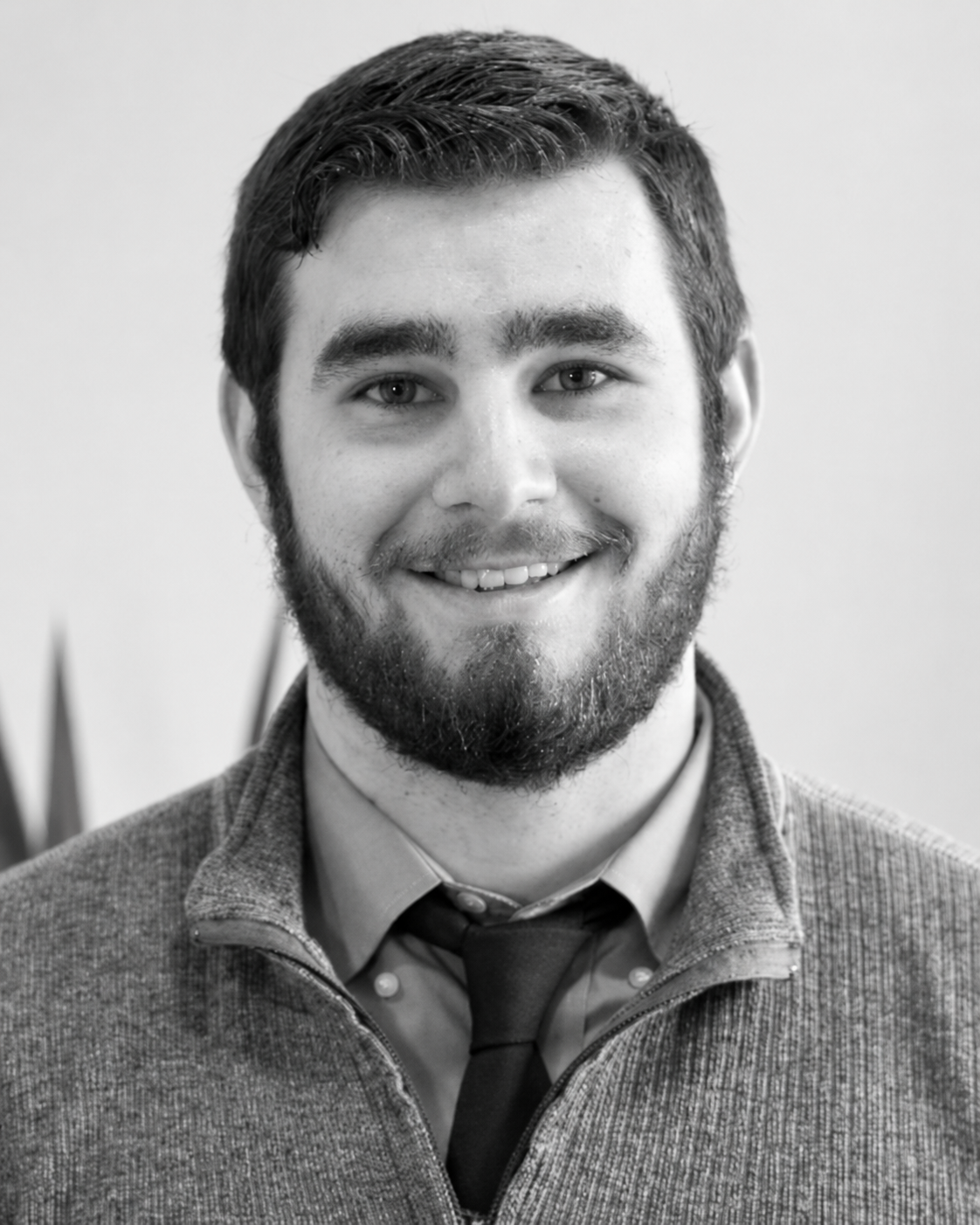}}]{Evan Newell}
received the B.S. and M.S. degrees in Electrical and Computer Engineering from Oregon State University, Corvallis, OR, USA, in 2024 and 2026 respectively, where he also earned a minor in Computer Science. He is currently an RF Engineer with Ether Form Inc. His research interests include computational electromagnetics and holographic MIMO (HMIMO) systems with a focus on near-field and coupling-aware beamforming.
\end{IEEEbiography}

\begin{IEEEbiography}[{\includegraphics[width=1in,height=1.25in,clip,keepaspectratio]{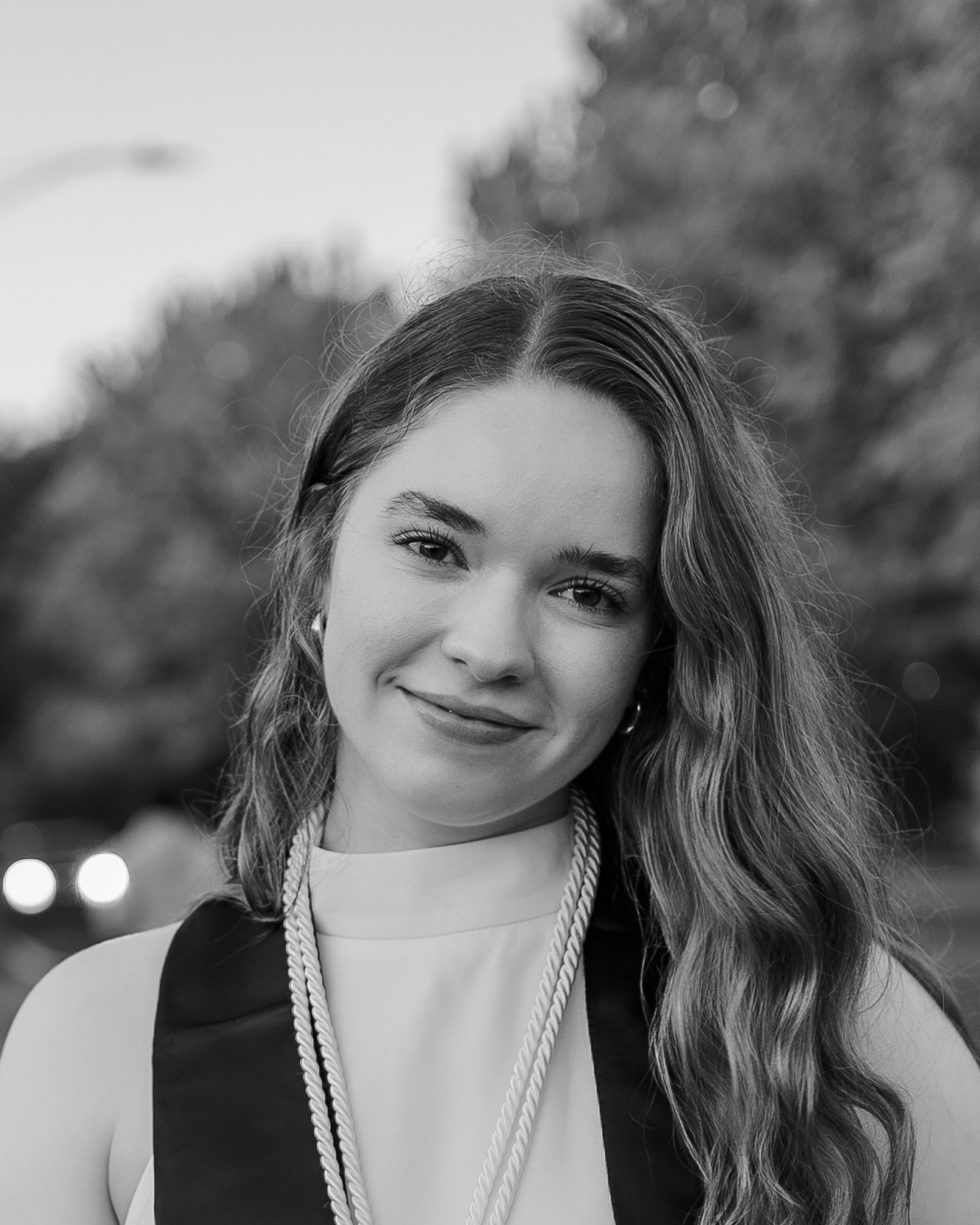}}]{Hanna O'Leary}
received her B.S. and M.S. degrees in Electrical and Computer Engineering from Oregon State University at Corvallis, Oregon, in 2024 and 2026, respectively, where she also completed a minor in Computer Science. She has gained industry and research experience through internships at Intel and the Johns Hopkins Applied Physics Laboratory. Her research interests span machine learning and artificial intelligence, generative models for inverse problems, signal processing, and algorithm development for analog and digital communication systems.
\end{IEEEbiography}

\vspace{-13em}
\begin{IEEEbiography}[{\includegraphics[width=1in,height=1.25in,clip,keepaspectratio]{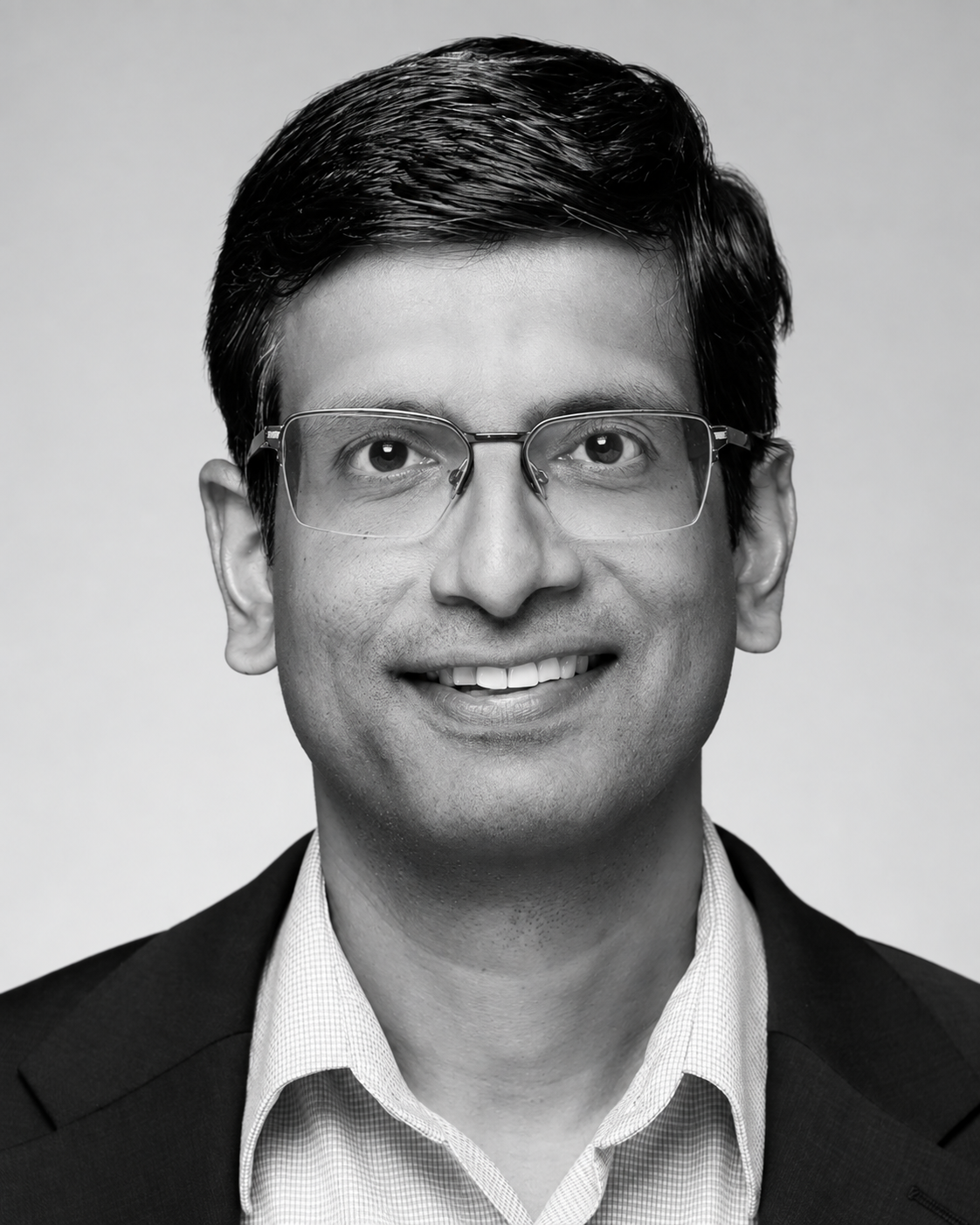}}]{Arun Natarajan}
(Senior Member, IEEE) received the B.Tech. degree in electrical engineering from IIT Madras, Chennai, India, in 2001, and the M.S. and Ph.D. degrees in electrical engineering from the California Institute of Technology, Pasadena, CA, USA, in 2003 and 2007, respectively.

He is currently a Professor of Electrical and Computer Engineering at Yale University, New Haven, CT, USA, where he directs the Analog/RF Circuits and Systems Lab. Prior to joining Yale, he was a Professor in the School of Electrical Engineering and Computer Science at Oregon State University, Corvallis, OR, USA. From 2007 to 2012, he was a Research Staff Member at the IBM T. J. Watson Research Center, Yorktown Heights, NY, USA, where he worked on millimeter-wave phased arrays for multi-Gb/s data links and airborne radar.

He was also part of the leadership team at Mixcomm Inc., Chatham, NJ, USA, a start-up focused on mm-wave beamformers for 5G and satellite communications, which Sivers Semiconductors acquired in 2022. His research interests include RF and mm-wave integrated circuits and systems for high-speed wireless communication and imaging, as well as low-power circuits for pervasive sensing and communication.

Prof. Natarajan was a recipient of the National Talent Search Scholarship from the Government of India from 1995 to 2000, the Caltech Atwood Fellowship in 2001, the IBM Research Fellowship in 2005, the 2011 Pat Goldberg Memorial Award for the Best Paper in Computer Science, Electrical Engineering, and Mathematics published by IBM Research, the CDADIC Best Faculty Project Award in 2014 and 2016, the NSF CAREER Award in 2016, the Oregon State University Engelbrecht Young Faculty Award in 2016, and the DARPA Young Faculty Award in 2017.

He has served on the Technical Program Committee of the IEEE International Solid-State Circuits Conference (ISSCC), the IEEE Radio Frequency Integrated Circuits Conference (RFIC), and the IEEE Compound Semiconductor Integrated Circuits Symposium (CSICS). He has served as an Associate Editor for IEEE Transactions on Very Large Scale Integration (VLSI) Systems, IEEE Transactions on Microwave Theory and Techniques, and IEEE Journal of Solid-State Circuits.
\end{IEEEbiography}
\vspace{-13em}
\begin{IEEEbiography}[{\includegraphics[width=1in,height=1.25in,clip,keepaspectratio]{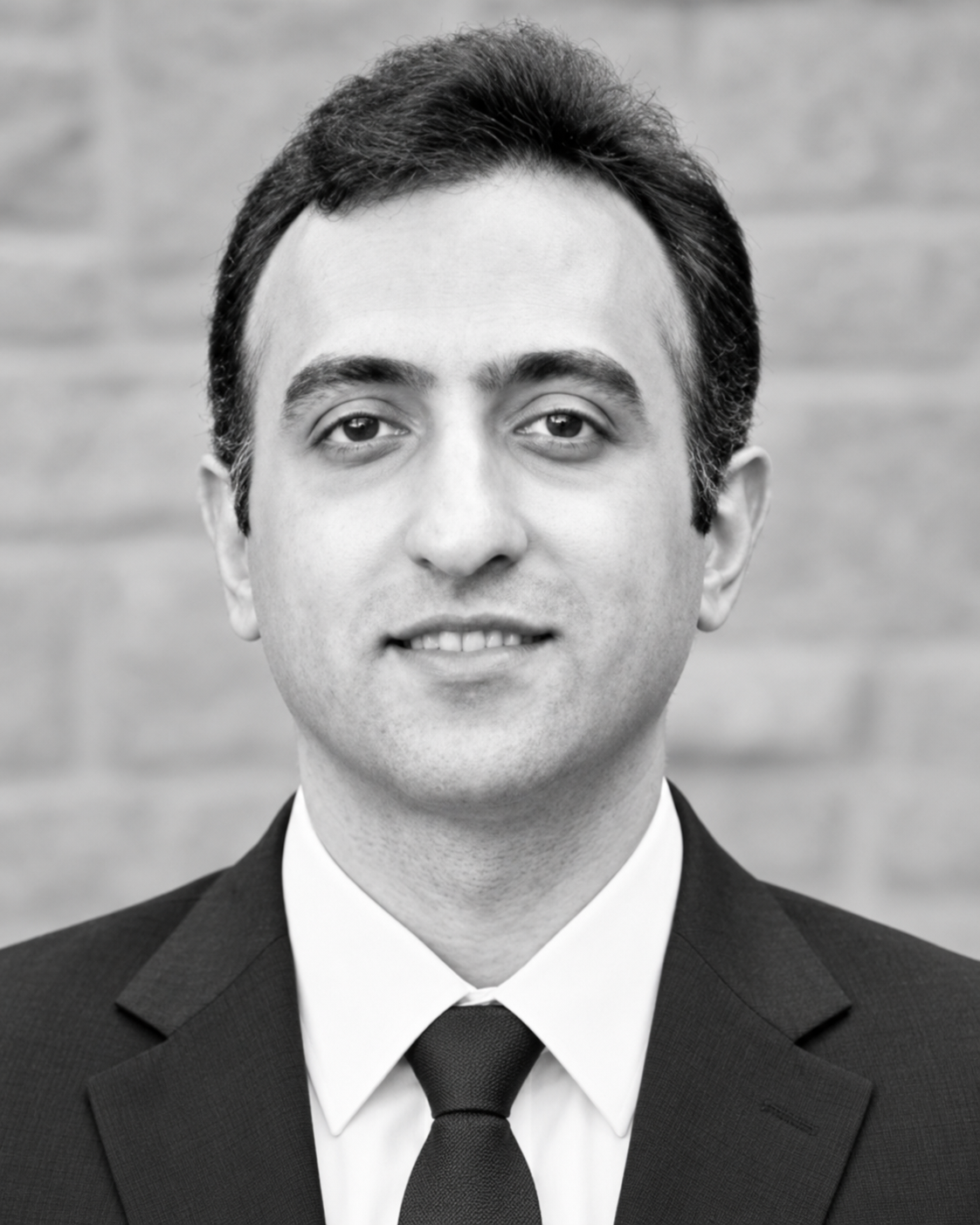}}]{Alireza Aghasi}
is currently an Associate Professor with the School of Electrical Engineering and Computer Science at Oregon State University. Previously, he held research positions at Georgia Tech, MIT, and IBM T. J. Watson Research Center, and was an Assistant Professor at Georgia State University. He received the Ph.D. degree in electrical and computer engineering from Tufts University, in 2012. His research interests include optimization theory, statistics, artificial intelligence, signal processing, high-dimensional probability, and physics-based inverse problems.
\end{IEEEbiography}

\end{document}